\documentclass{article}

\AtBeginDocument{%
  }

\usepackage{arxiv}
\usepackage{amsfonts}
\usepackage{url}
\usepackage{booktabs}
\usepackage{subfig}
\usepackage{placeins}
\usepackage{amsmath}
\usepackage[capitalise,noabbrev]{cleveref}
\usepackage{graphicx}
\usepackage{multicol,multirow}
\usepackage{enumitem}
\usepackage{xcolor}
\usepackage[normalem]{ulem}
\usepackage{siunitx}
\usepackage{xspace}
\makeatletter
\DeclareRobustCommand\onedot{\futurelet\@let@token\@onedot}
\def\@onedot{\ifx\@let@token.\else.\null\fi\xspace}
\DeclareRobustCommand\nodot{\futurelet\@let@token\@nodot}
\def\@nodot{\ifx\@let@token.\else~\null\fi\xspace}

\newcommand{\comm}[2]{{\textcolor{#1}{#2}}}

\newcommand{\sg}[1]{\comm{black}{#1}}

\newcommand{\gm}[1]{\comm{black}{#1}}

\def\eg{\emph{e.g}\onedot} 

\def\ie{\emph{i.e}\onedot}

\def\wrt{w.r.t\onedot}

\def\etal{\emph{et al}\onedot}
\makeatother

\begin{document}

\title{Sketch\&Patch++: Efficient Structure-Aware 3D Gaussian Representation}

\author{
  \textbf{Yuang Shi}$^{1}$,
  \textbf{Géraldine Morin}$^{2}$,
  \textbf{Simone Gasparini}$^{2}$,
  \textbf{Wei Tsang Ooi}$^{1}$ \vspace{1mm} \\
  $^1$National University of Singapore\quad
  $^2$IRIT - Université de Toulouse \quad   \\
  \small{\texttt{\{yuangshi, ooiwt\}@comp.nus.edu.sg}} \\
  \small{\texttt{\{simone.gasparini, geraldine.morin\}@toulouse-inp.fr}} 
  \vspace{-5mm}
}





\maketitle

\begin{figure*}[ht]
    \centering
    \includegraphics[width=0.99\textwidth]{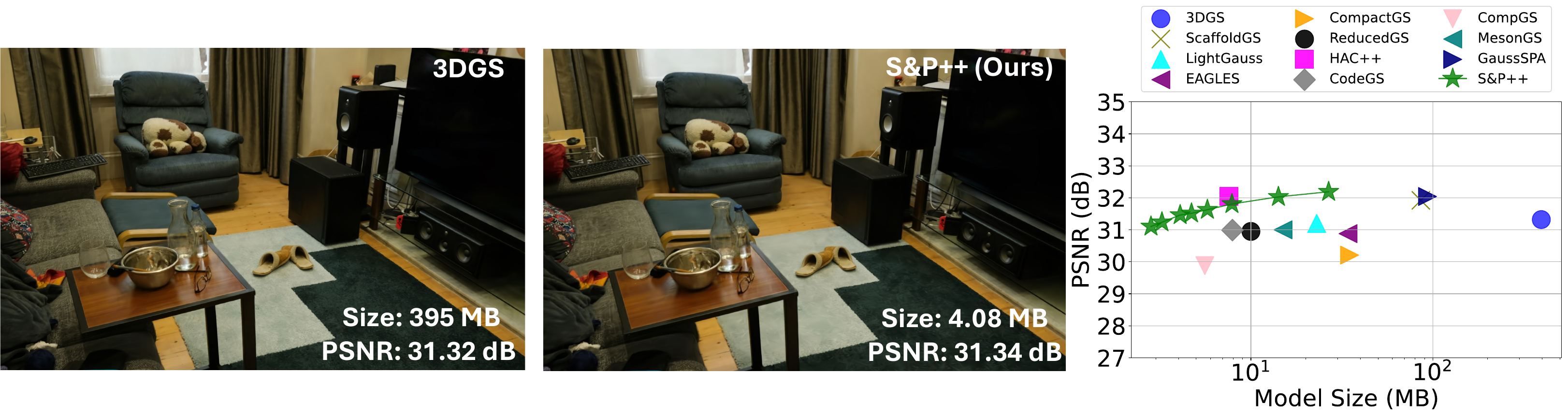}
    \caption{We propose Sketch\&Patch++, a hybrid Gaussian representation with 3D structure prior, significantly reducing the storage of 3DGS model by an order of magnitude while maintaining the visual quality.}
\end{figure*}

\begin{abstract}
    3D Gaussian Splatting (3DGS) has emerged as a promising representation for photorealistic rendering of 3D scenes. 
    However, its high storage requirements pose significant challenges for practical applications such as remote visualization, collaborative 3D content sharing, and bandwidth-constrained streaming in extended reality systems. 
    We observe that Gaussians exhibit distinct roles and characteristics analogous to traditional artistic techniques --- like how artists first sketch outlines before filling in broader areas with color, some Gaussians capture high-frequency features such as edges and contours, while others represent broader, smoother regions analogous to brush strokes that add volume and depth.
    
    Based on this observation, we propose a hybrid representation that categorizes Gaussians into (i) \textit{Sketch Gaussians}, which represent high-frequency, boundary-defining features, and (ii) \textit{Patch Gaussians}, which cover low-frequency, smooth regions. This semantic separation naturally enables layered progressive streaming, where the compact Sketch Gaussians establish the structural skeleton before Patch Gaussians incrementally refine volumetric detail.
    In our previous work, we demonstrated the effectiveness of this categorization for man-made scenes by identifying Sketch Gaussians using 3D line segments extracted via Line3D++. However, this line-based approach was fundamentally limited to scenes with linear features. 
    
    In this work, we extend our method to arbitrary 3D scenes by proposing a novel hierarchical adaptive categorization framework that operates directly on the 3DGS representation. Our approach employs multi-criteria density-based clustering, combined with adaptive quality-driven refinement. This method eliminates dependency on external 3D line primitives while ensuring optimal parametric encoding effectiveness.
    After categorization, Sketch Gaussians are efficiently encoded using parametric models leveraging their coherence, while Patch Gaussians undergo optimized pruning, retraining, and quantization. 
    Our comprehensive evaluation across diverse scenes, including both man-made and natural environments, demonstrates that our method achieves up to \SI{1.74}{\decibel} improvement in PSNR, \SI{6.7}{\percent} in SSIM, and \SI{41.4}{\percent} in LPIPS at equivalent model sizes compared to uniform pruning baselines. For indoor scenes, our method can maintain visual quality with only \SI{0.5}{\percent} of the original model size, representing a {175$\times$} compression ratio. 
    This structure-aware representation enables efficient storage, adaptive streaming, and rendering of high-fidelity 3D content across bandwidth-constrained networks and resource-limited devices.
\end{abstract}


\section{Introduction}
\label{sec:intro}

\subsection{Background and Motivation} \label{subsec:intro-brackground}
The increasing demand for immersive experiences in extended reality (XR) applications, such as virtual reality (VR), augmented reality (AR), and cloud gaming, has driven significant progress in 3D scene representation technologies.
These applications rely on accurately modeling complex 3D environments to deliver high-quality user experiences. 
Classical 3D representations such as meshes and point clouds face challenges in balancing storage efficiency with rendering quality, particularly for large-scale environments where point clouds can introduce visual artifacts and holes during rendering~\cite{kerbl20233D,shi2024volumetric}.

Recently, learning-based methods have given rise to alternative representations that seamlessly model geometry and appearance. 
3D Gaussian Splatting (3DGS) \cite{kerbl20233D} has emerged as a leading approach for learning more explicit and efficient point-based 3D representations while keeping the mix between geometry and appearance, leading to a very satisfying visual quality.
Compared to point clouds, 3DGS represents 3D scenes with a collection of 3D ellipsoidal Gaussian splats (or Gaussians for short), which can mitigate the presence of holes with a much sparser distribution. 
By leveraging the properties of Gaussians and taking advantage of modern GPUs, 3DGS achieves comparable photorealistic quality to NeRF~\cite{mildenhall2021nerf} but with significantly improved rendering speed,
showing great promise as a foundational representation for future immersive multimedia systems and applications. 
While 3DGS achieves impressive visual fidelity, it still generates vast numbers of Gaussians to capture fine geometric and appearance details, with five attributes stored independently. 
For instance, Kerbl~\etal~\cite{kerbl20233D} employ one to five million Gaussians to model static scenes, demanding up to \SI{1}{\giga\byte} of storage per scene. 
Such a heavy representation, while suitable for local rendering on powerful hardware, poses significant challenges for remote rendering, sharing, and real-time transmission of 3DGS models across networks.

The substantial storage requirements of 3DGS models have motivated continuous efforts in compression.
Numerous recent works (see \cref{subsec:related-compact3dgs}) have proposed various compression strategies, achieving significant size reductions while maintaining visual quality.
However, compression alone is insufficient for practical deployment, as efficient delivery of 3DGS content requires adapting compressed representations to streaming pipelines that enable progressive transmission and adaptive quality based on network conditions and viewing requirements~\cite{sun2025lts,krishna2025netsplat}.
Common point-based streaming strategies, such as 3D downsampling or spatial tiling, are widely used to generate scalable and multi-resolution representations for 3D streaming~\cite{han2020vivo,viola2023volumetric,shi2023enabling,shi2024qv4,rudolph2023rabbit}. 
Given that 3DGS is also a point-based representation (\ie the centers of the Gaussians), it is natural to consider adapting these techniques for Gaussians.
However, these streaming techniques are difficult to apply to 3DGS due to the unique properties of its Gaussian-based representation and optimization. 
Unlike conventional point clouds, 3DGS employs heterogeneous Gaussians parameterized by ellipsoid shape and view-dependent color, with attributes optimized iteratively to minimize discrepancies between rendered and ground-truth images, leading to a non-uniform and highly interdependent distribution. 
This tightly coupled optimization means that spatial partitioning or downsampling techniques often result in substantial visual quality degradation~\cite{shi2025lapisgs,shi20253d,sun2025lts,tsai2025l3gs,kim2025vega}, as the interdependent Gaussians cannot be meaningfully decomposed without retraining.
Furthermore, most existing compression methods treat 3DGS as a monolithic representation without semantic structure, making it challenging for progressive/scalable streaming.


\subsection{Key Observations} \label{subsec:intro-observation}

\begin{figure*}[t!]
    \centering
    \subfloat[3DGS (Relaxed, $\tau=0.01$).]{%
        \begin{minipage}[b]{0.3\textwidth}
            \centering
            \includegraphics[width=\textwidth]{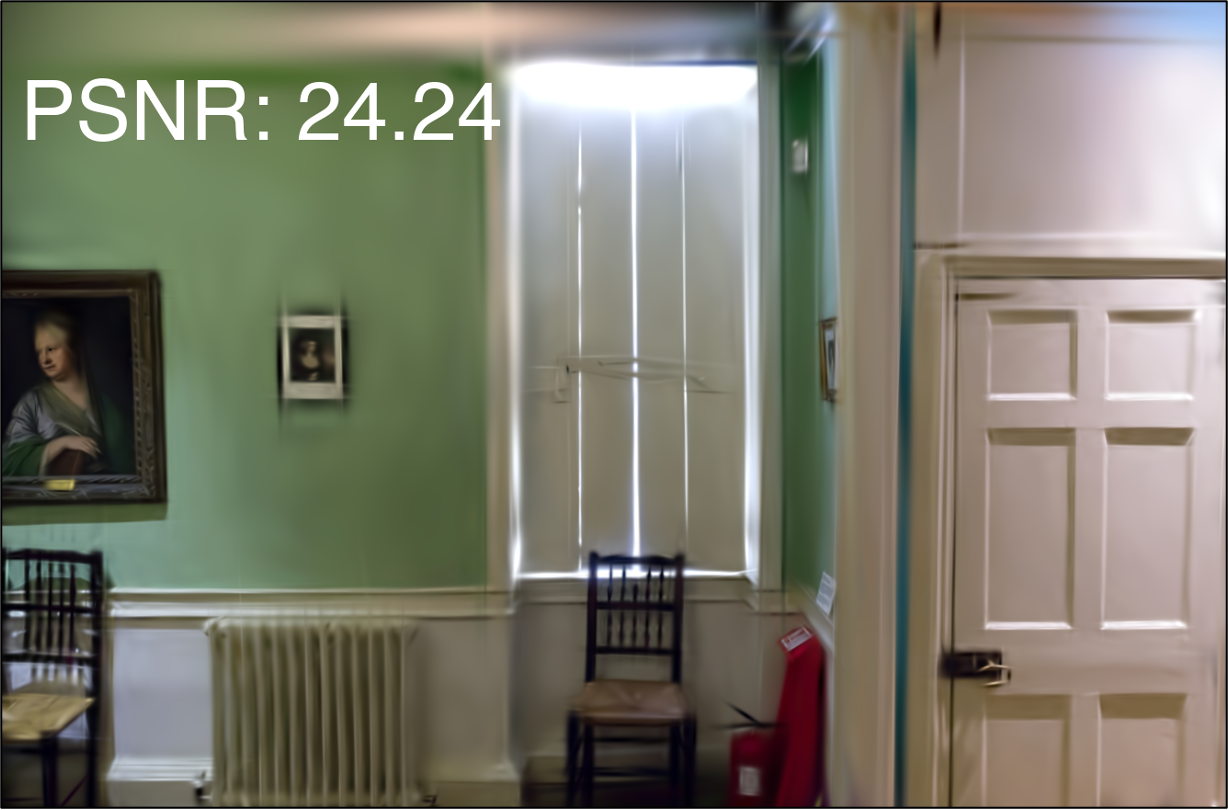}\\
            \includegraphics[width=\textwidth]{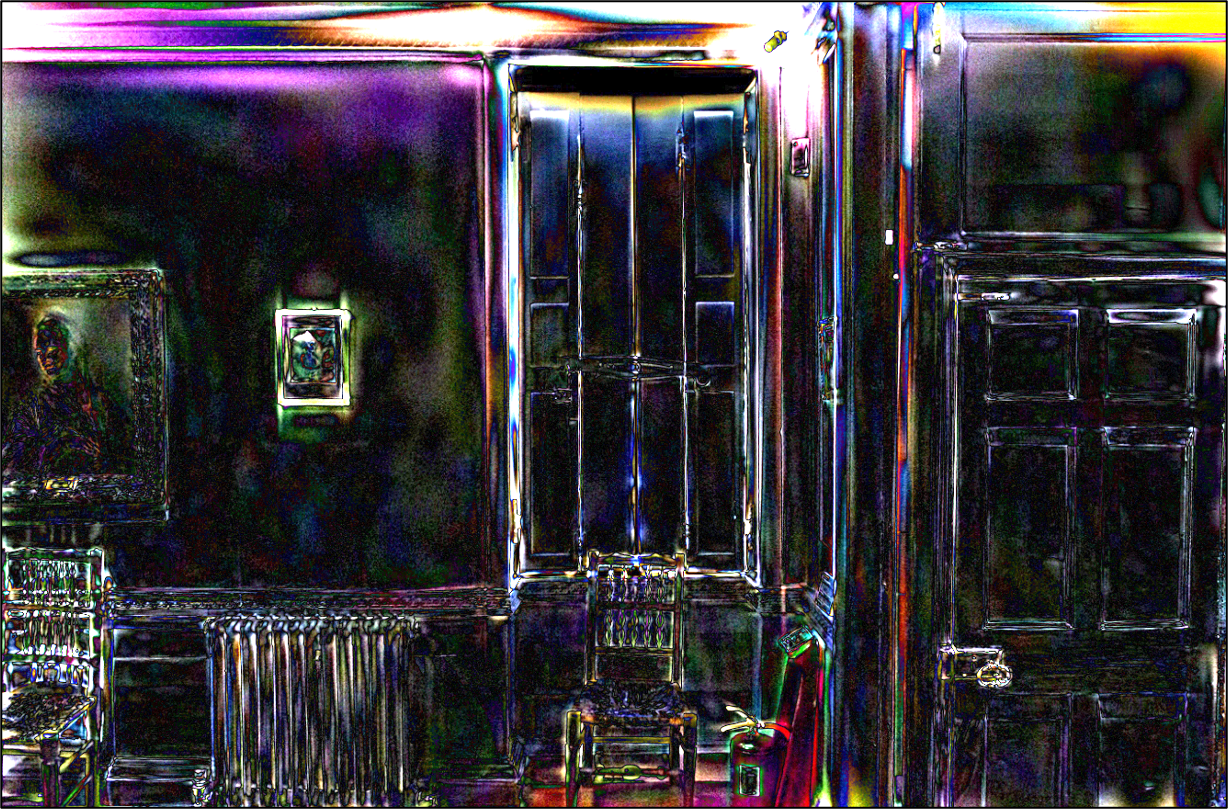}\\
            \includegraphics[width=\textwidth]{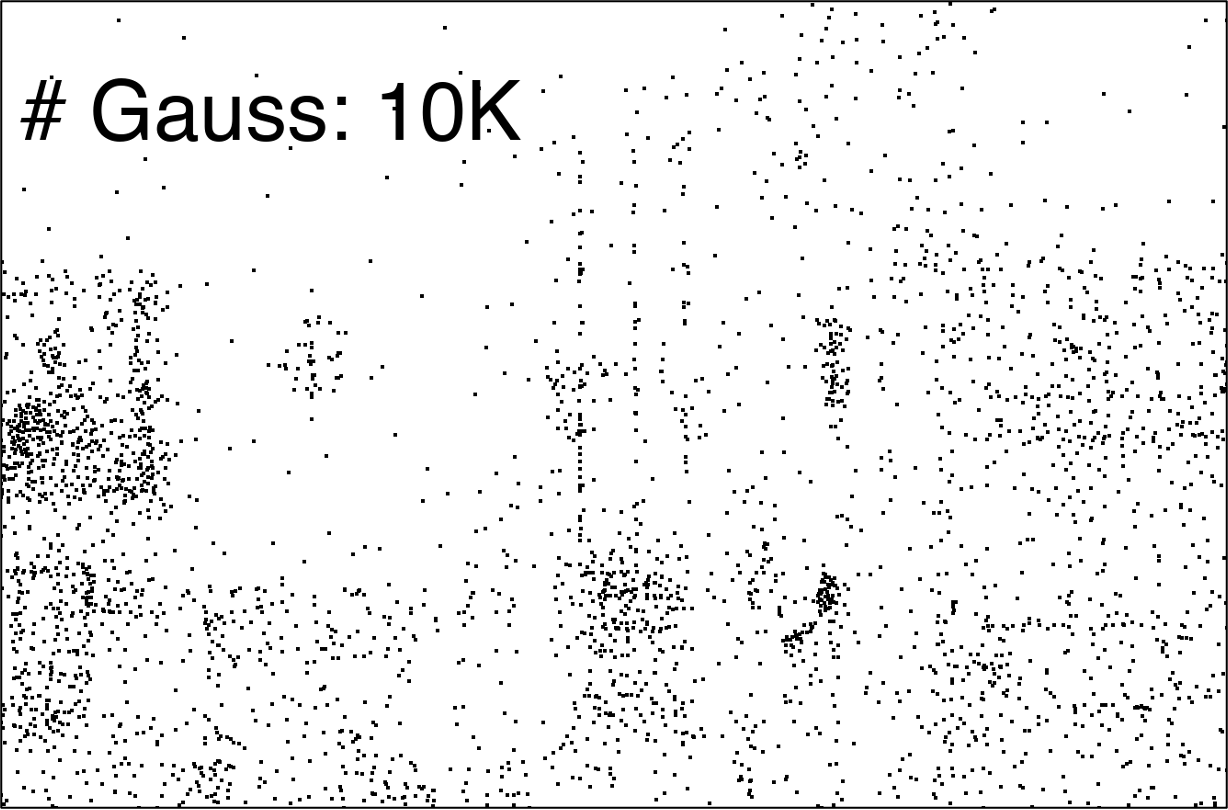}
        \end{minipage}
        \label{subfig:densify_relax}
    }
    \hfill
    \subfloat[3DGS (Medium, $\tau=0.002$).]{%
        \begin{minipage}[b]{0.3\textwidth}
            \centering
            \includegraphics[width=\textwidth]{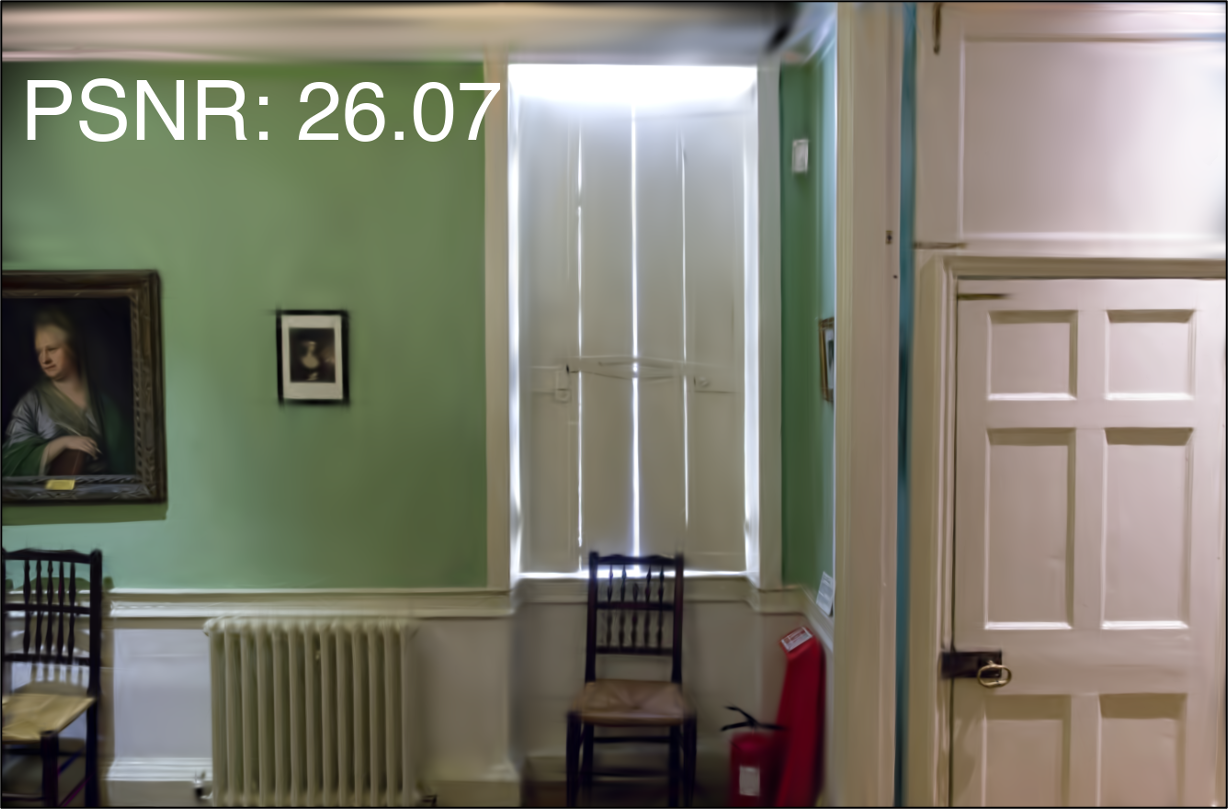}\\
            \includegraphics[width=\textwidth]{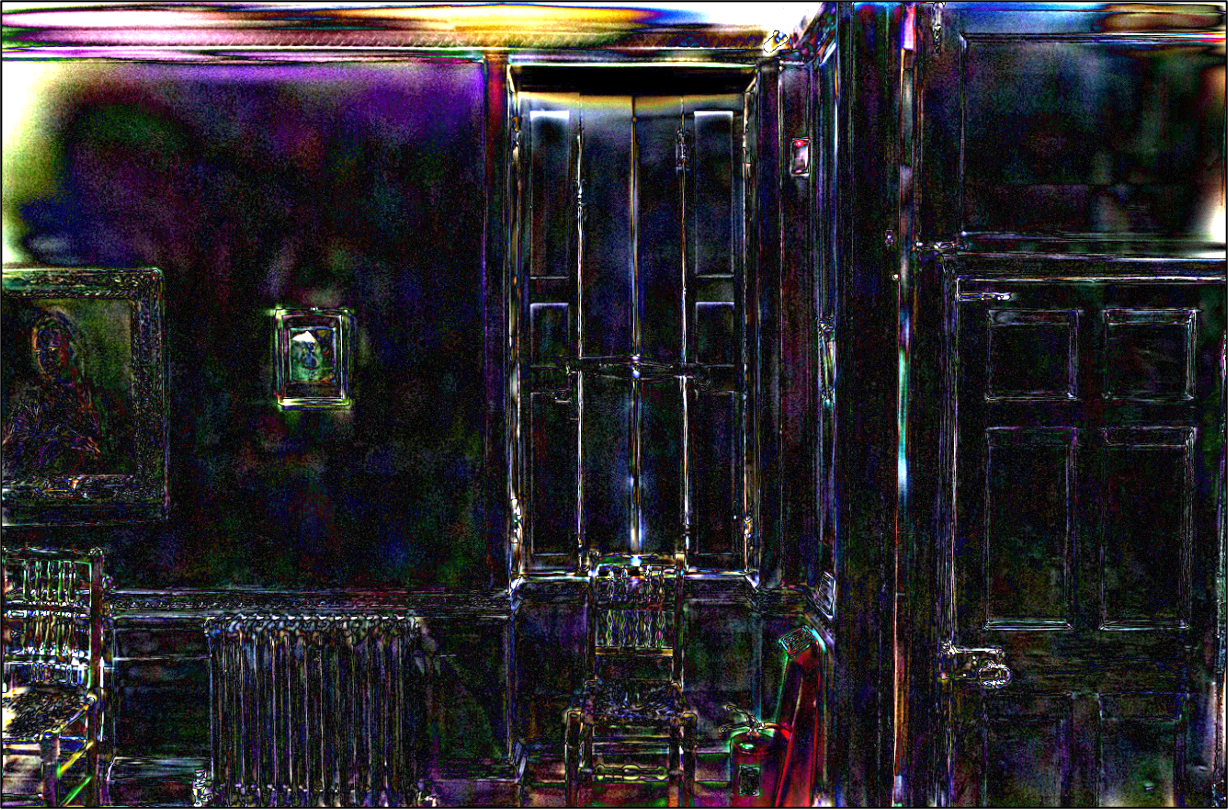}\\
            \includegraphics[width=\textwidth]{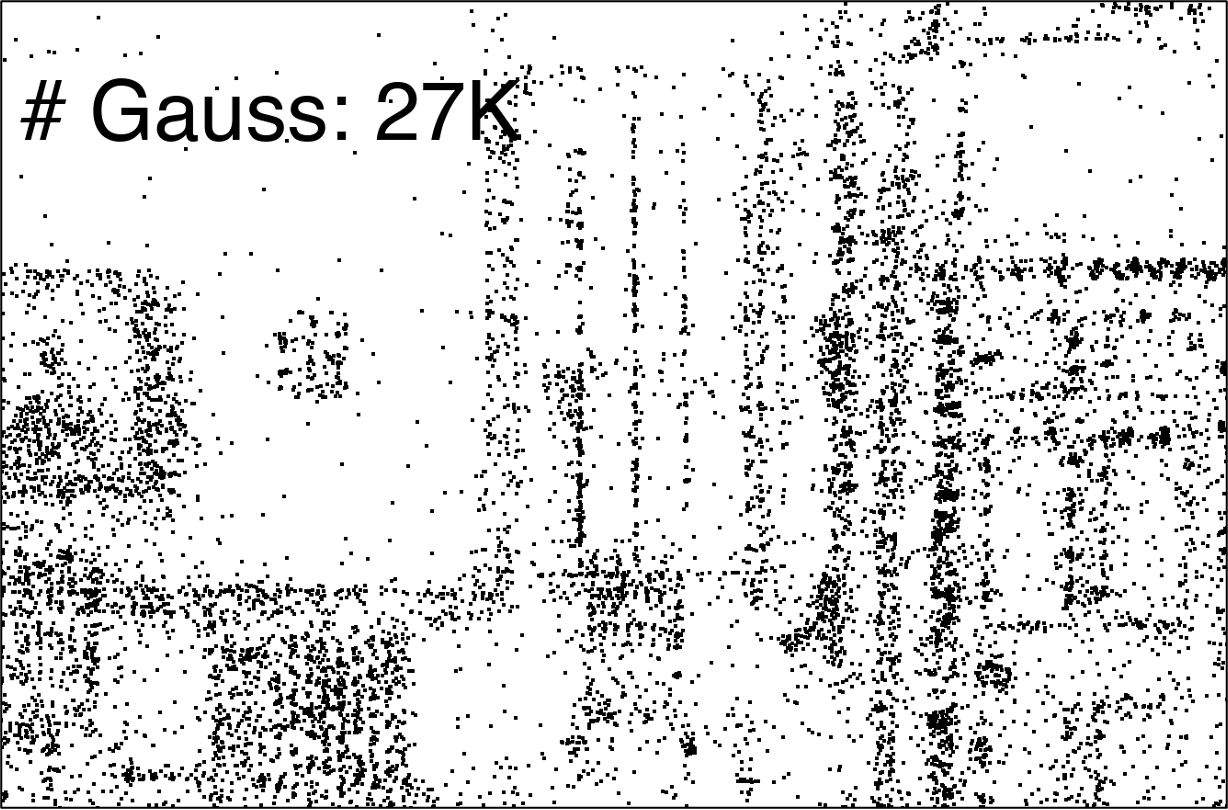}
        \end{minipage}
        \label{subfig:densify_medium}
    }
    \hfill
    \subfloat[3DGS (Strict, $\tau=0.0002$).]{%
        \begin{minipage}[b]{0.3\textwidth}
            \centering
            \includegraphics[width=\textwidth]{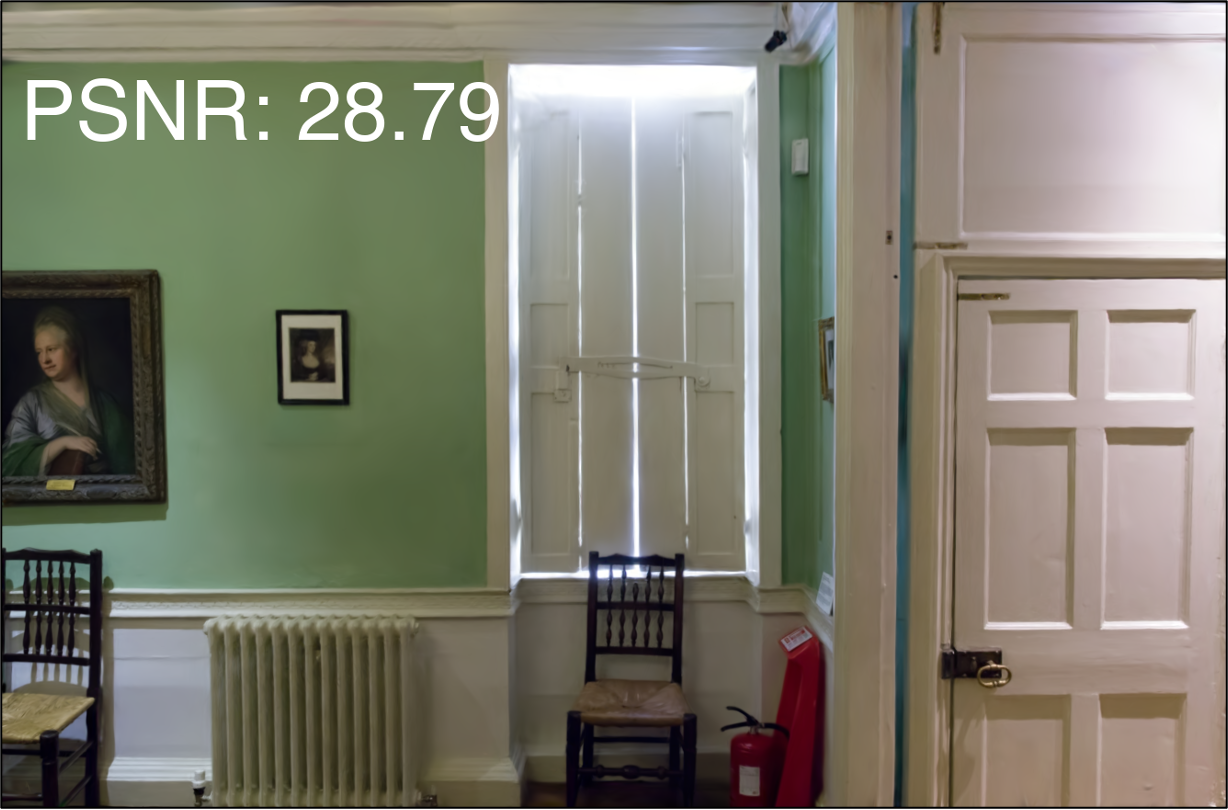}\\
            \includegraphics[width=\textwidth]{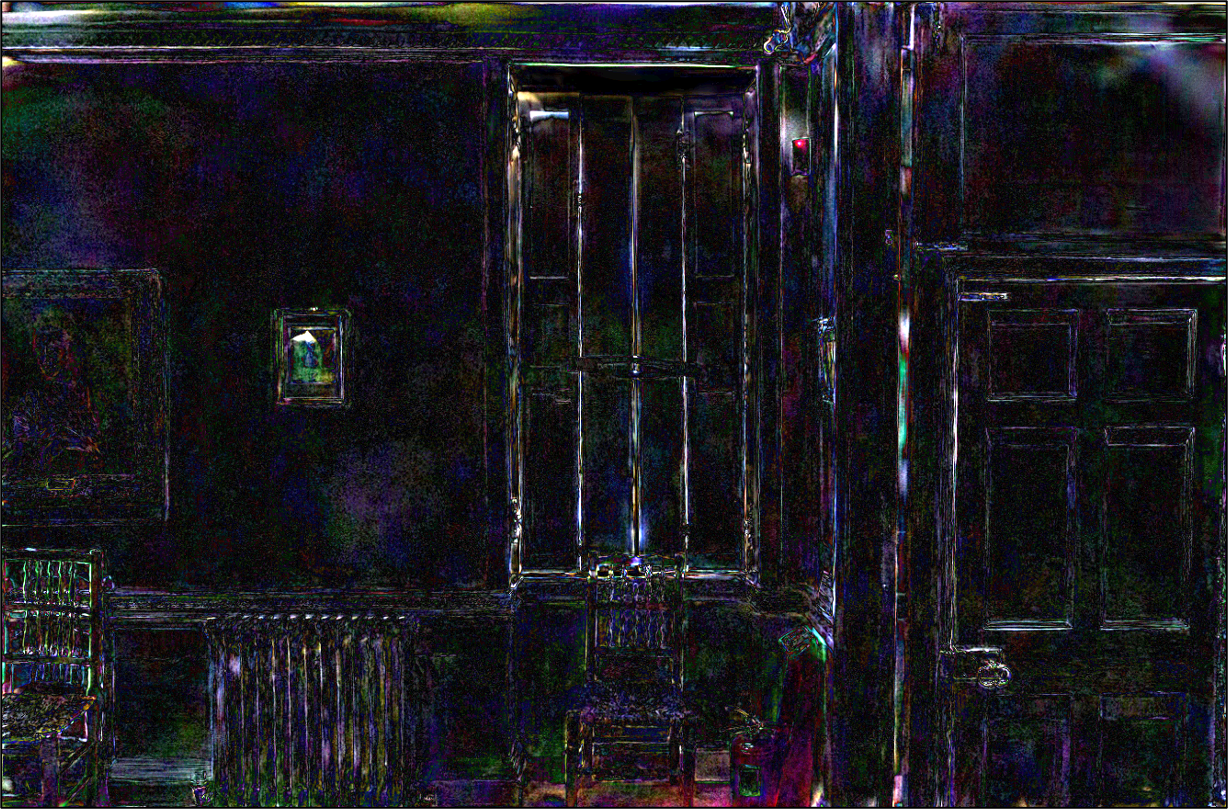}\\
            \includegraphics[width=\textwidth]{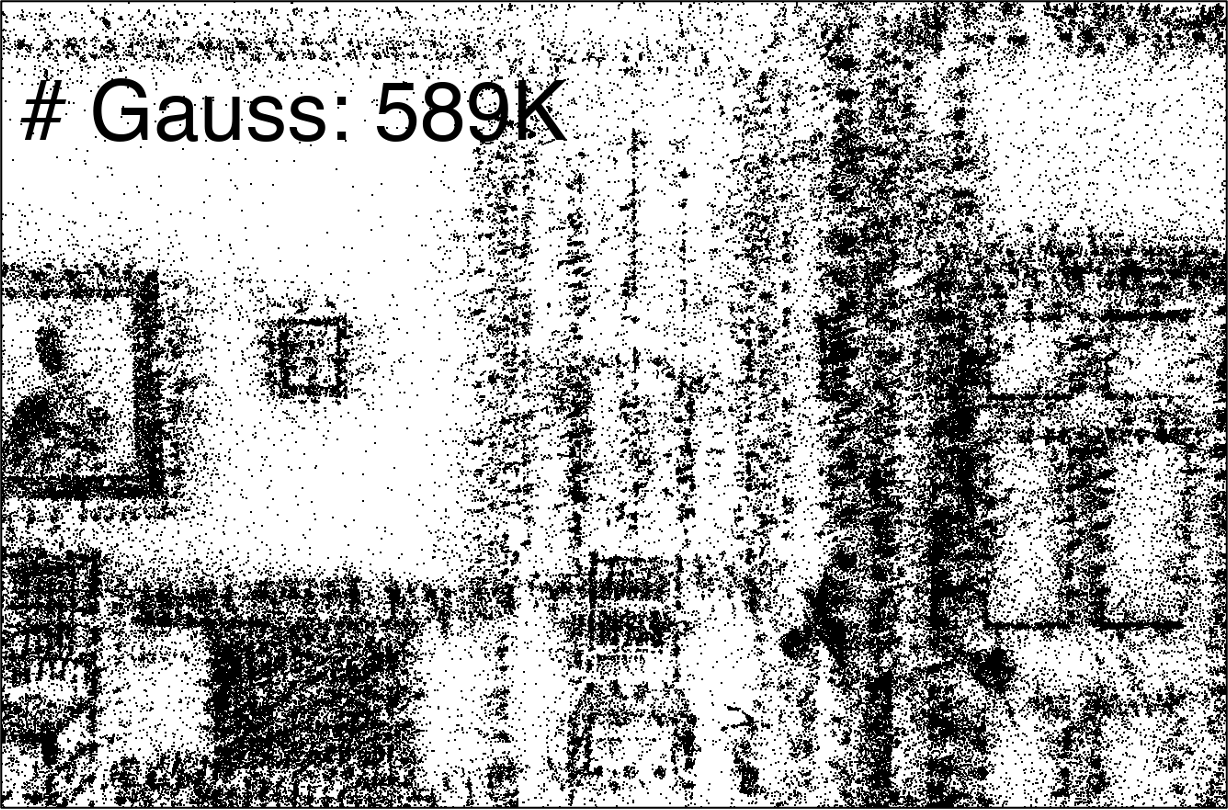}
        \end{minipage}
        \label{subfig:densify_strict}
    }
    \caption{Impact of densification on 3DGS models. (a) Relaxed ($\tau=0.01$), (b) medium ($\tau=0.002$), and (c) strict ($\tau=0.0002$) densification thresholds produce 10K, 27K, and 589K Gaussians respectively. 
    Top row: rendered images with PSNR values. 
    Middle row: per-pixel reconstruction error maps. Bright regions indicate higher reconstruction error, while darker areas correspond to well-reconstructed regions. 
    Bottom row: Gaussian distributions. Note the correlation between the density of Gaussians and the sharp/smooth areas. 
    }
    \label{fig:densification_illu}
\end{figure*}

3DGS's adaptive density control mechanism introduces storage inefficiencies.
The adaptive density control aims to enhance visual fidelity by cloning Gaussians in under-reconstructed regions and splitting those in over-reconstructed areas, based on 2D positional gradients. 
And the reliance on 2D gradients often highlight the need for densification in high-frequency and boundary-defining areas, such as edges and contours~\cite{shi2025sketch}. 
As a result, 3DGS naturally forms dense clusters of Gaussians along structural boundaries and detailed areas, while smooth regions such as walls remain sparsely populated.
This adaptive behavior is essential for capturing fine details and avoiding artifacts.
However, it also leads to a highly non-uniform distribution where many Gaussians convey redundant information in locally dense regions~\cite{kim2024color,li2025geogaussian,mallick2024taming3dgs}.

For better understanding, we visualize the impact of adaptive density control in \cref{fig:densification_illu}, which compares three 3DGS models trained with different densification thresholds $\tau$. 
Here, we vary this parameter to produce three levels of density: a relaxed configuration ($\tau=0.01$), a medium configuration ($\tau=0.002$), and a strict configuration ($\tau=0.0002$, identical to the original 3DGS implementation~\cite{kerbl20233D}), resulting in 10K, 27K, and 589K Gaussians, respectively.
As shown, the per-pixel reconstruction error maps (middle row) reveal that reconstruction errors are predominantly concentrated in high-frequency regions such as edges, contours, and fine details, while smooth areas like walls exhibit minimal error.
To reduce these errors and improve visual quality, 3DGS adaptively allocates more Gaussians to these challenging regions, as evidenced by the dense clustering along edges and structural boundaries in the bottom row.
This gradient-driven densification naturally produces dense clusters at edges and sparse coverage in smooth regions.

This observation motivates our representation approach, which exploits this structural difference by applying specialized encoding strategies to these two distinct types of regions.
Specifically, we propose categorizing Gaussians into two distinct roles: \textit{Sketch Gaussians} (SketchGS in short) and \textit{Patch Gaussians} (PatchGS in short), drawing inspiration from traditional artistic techniques.
Like how artists first sketch outlines before filling in broader areas with color, SketchGS captures boundary-defining features such as edges and contours, serving as the semantic scaffolding of the scene. 
In contrast, PatchGS, analogous to broader brush strokes that add volume and depth to a painting, provides volumetric coverage for smoother and broader regions.

\begin{figure*}[t!]
    \centering
    \begin{minipage}{\textwidth}
        \subfloat[SketchGS (1.16 million Gaussians).]{%
            \includegraphics[width=0.49\textwidth]{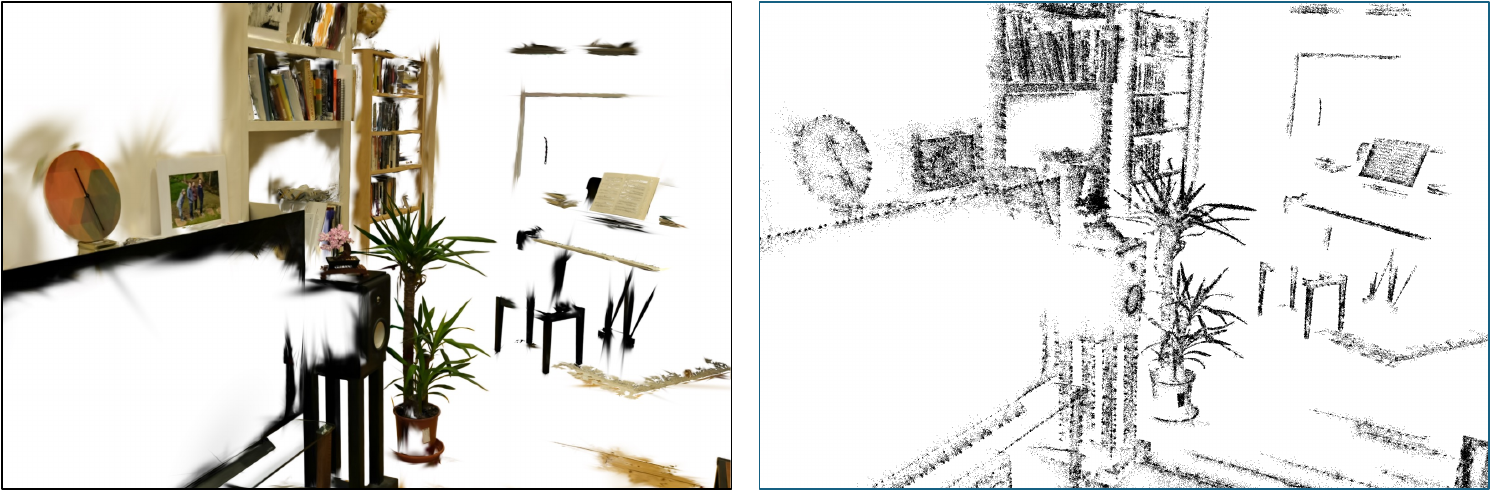}
            \label{subfig:sketchgs}
            }
        \subfloat[PatchGS (0.43 million Gaussians).]{%
            \includegraphics[width=0.49\textwidth]{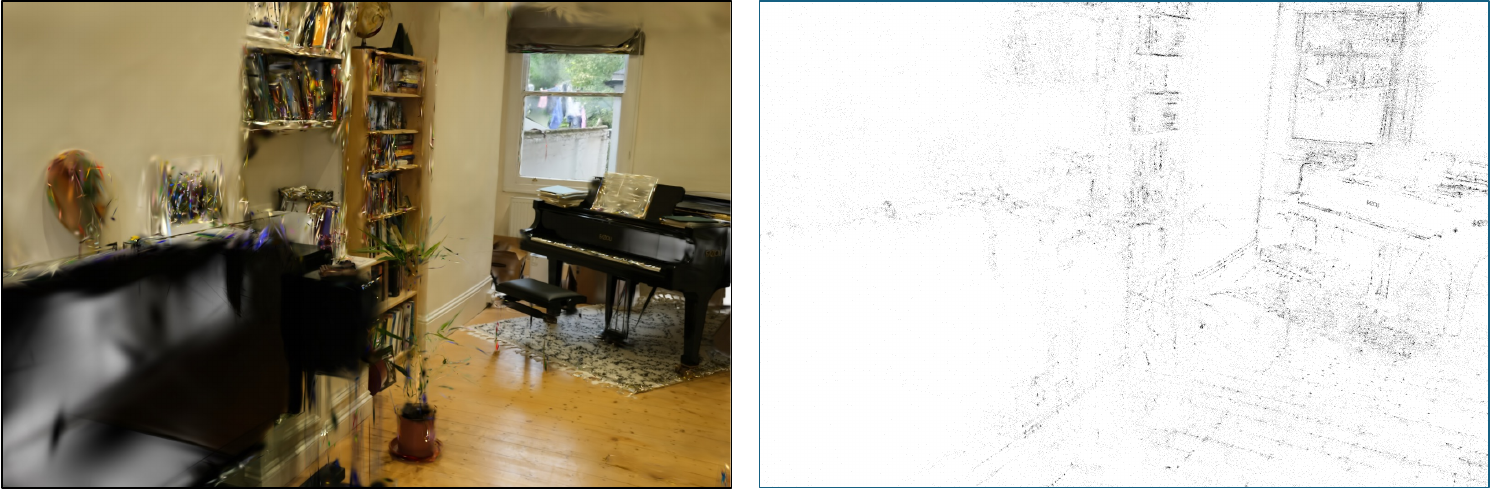}
            \label{subfig:patchgs}
            }
        \caption{Renderings of SketchGS and PatchGS in the \textit{Room} scene. (a) SketchGS (1.16 million Gaussians) exhibits dense edge-aligned structures. (b) PatchGS (0.43 million Gaussians) shows scattered surface coverage. For each category, we show the 3DGS rendering (left) and Gaussian center rendering (right).}
        \label{fig:sketch_patch_renders}
    \end{minipage}
    
    \begin{minipage}{\textwidth}
        \centering
        \subfloat[Number.]{%
            \includegraphics[width=0.24\textwidth]{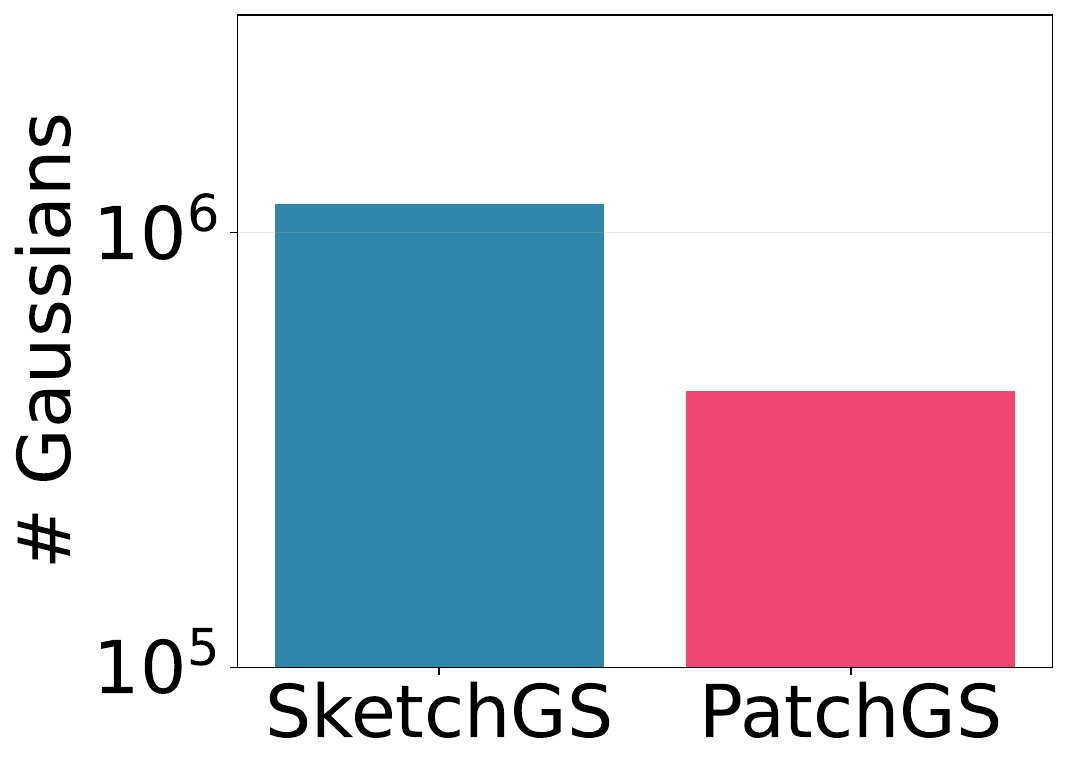}
            \label{subfig:number}
            }
        \subfloat[Density.]{%
            \includegraphics[width=0.24\textwidth]{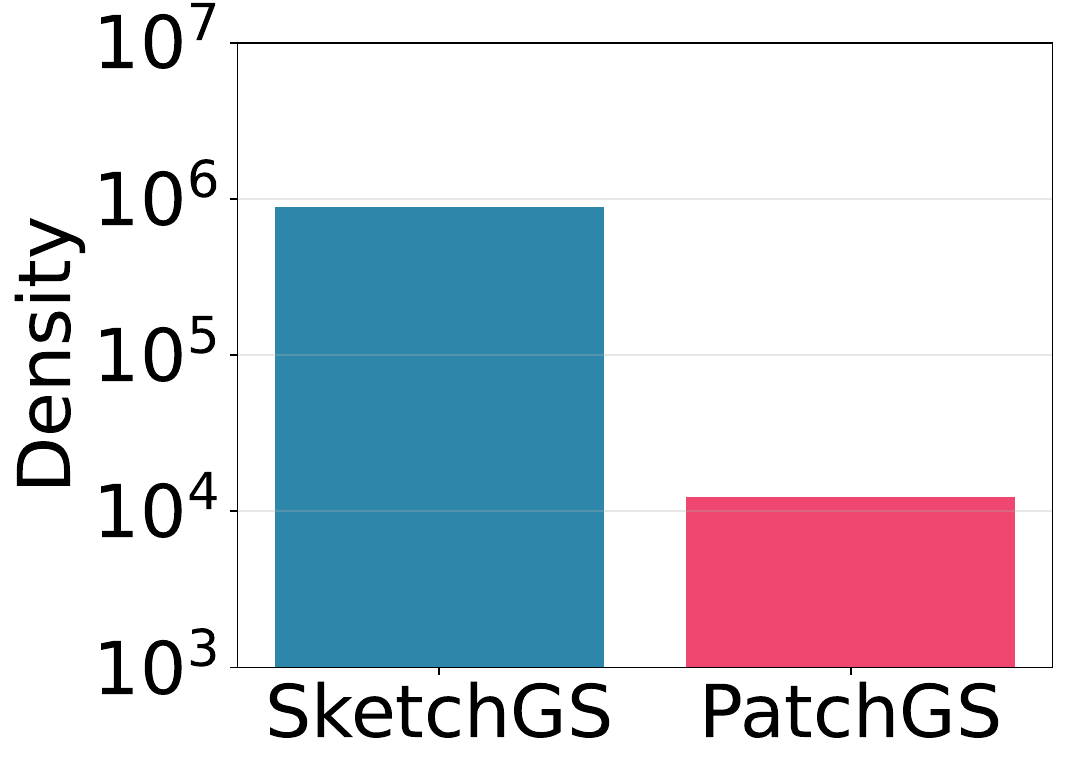}
            \label{subfig:density}
            }
        \subfloat[Elongation.]{%
            \includegraphics[width=0.24\textwidth]{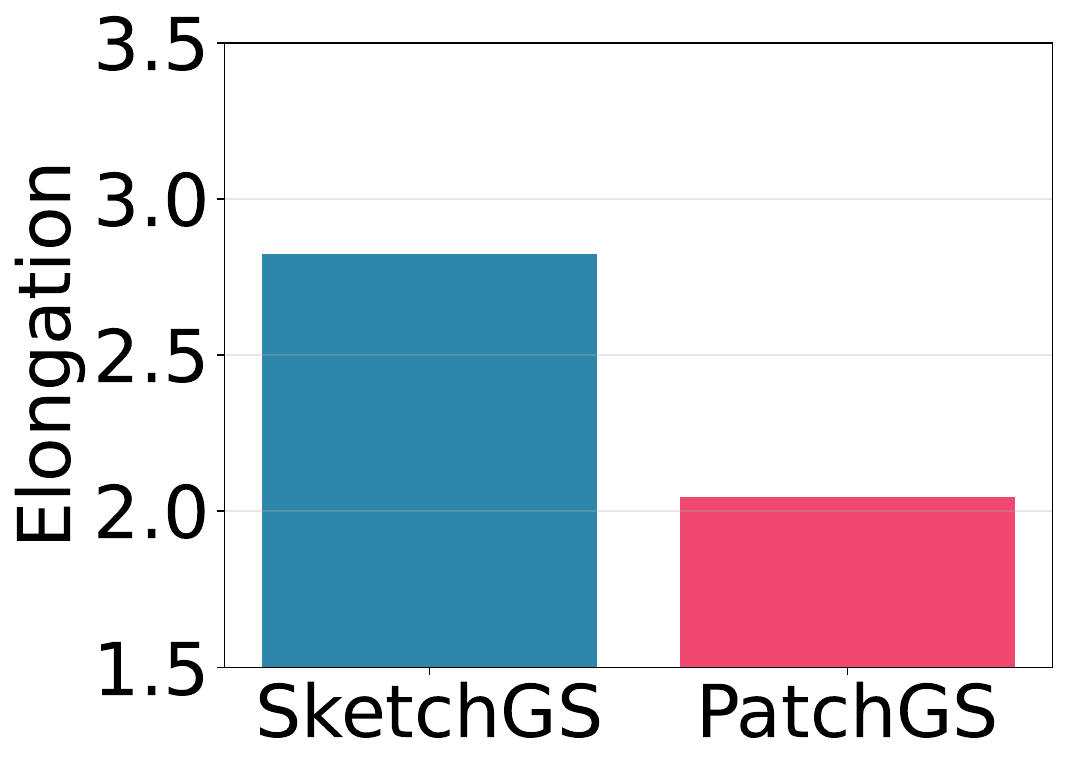}
            \label{subfig:elongation}
            }
        \subfloat[Spatial Volume.]{%
            \includegraphics[width=0.24\textwidth]{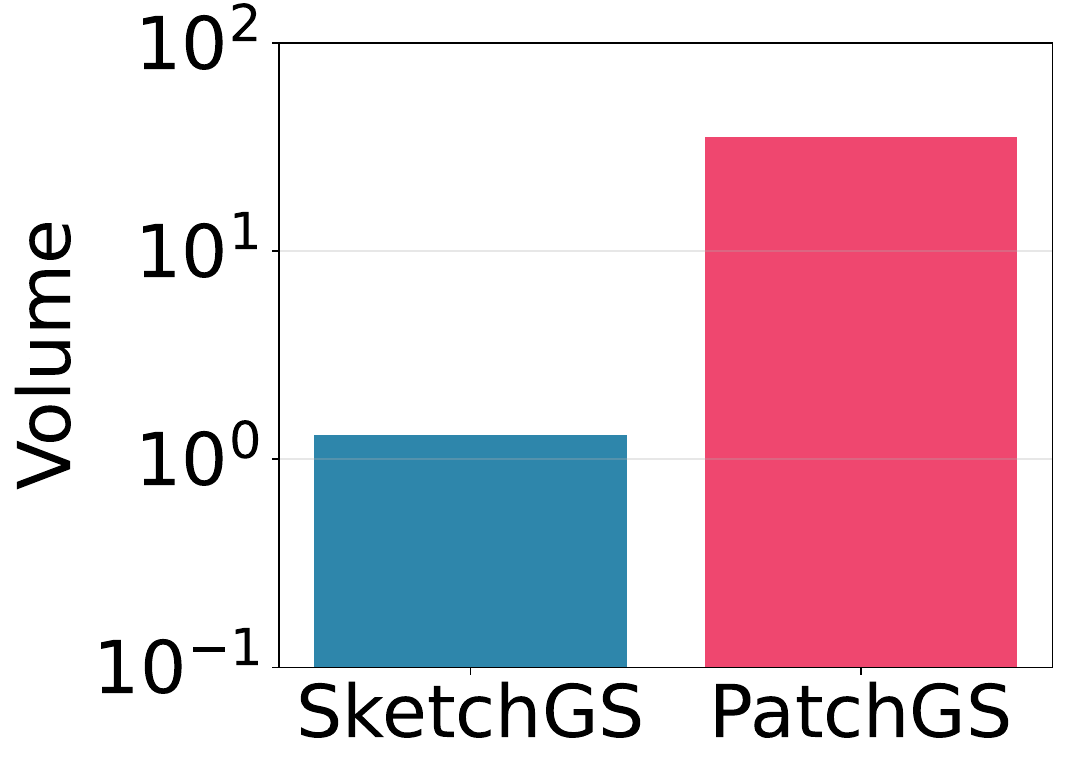}
            \label{subfig:volume}
            }
        \caption{Characteristic of SketchGS and PatchGS in the \textit{Room} scene: (a) Gaussian count, (b) density, (c) elongation, and (d) spatial volume. Compared with PatchGS (1.16 million Gaussians), SketchGS (0.43 million Gaussians) are $100\times$ denser, $1.5\times$ more elongated, and $27\times$ more compact, confirming their distinct roles as edge-aligned structures versus scattered surface coverage.}
        \label{fig:sketch_patch_stats}
    \end{minipage}
\end{figure*}

To unveil the characteristics of this categorization, we extract SketchGS and PatchGS in the \textit{Room} scene~\cite{barron2022mip} as an example (shown in \cref{fig:sketch_patch_renders}), and quantify their characteristics (shown in \cref{fig:sketch_patch_stats}). 
Specifically, the SketchGS comprises $1.16$ million Gaussians organized into coherent clusters, while PatchGS contains $0.43$ million unclustered Gaussians (\cref{subfig:number}). 
More critically, \cref{subfig:density} demonstrates that SketchGS exhibits significantly higher spatial density ($10^6$ Gaussians per unit volume) compared to PatchGS ($10^4$ Gaussians per unit volume), approximately $100\times$ denser, confirming their role in capturing fine-grained details within compact regions. \Cref{subfig:elongation} further shows that SketchGS clusters are notably more elongated ($2.82$ on average elongation) than PatchGS samples ($1.99$ on average elongation), where elongation is computed as the ratio of the first to second principal component variance from PCA, quantifying the degree of linear alignment along edge structures.
This observation is consistent with their alignment along linear or curvilinear edge structures. 
Furthermore, \cref{subfig:volume} reveals that SketchGS occupy minimal spatial volume ($1.30$ $\text{unit}^3$ on average), while PatchGS are distributed across substantially larger volumes ($35.11$ $\text{unit}^3$). 

\begin{figure*}[t]%
    \centering
    \includegraphics[width=\textwidth]{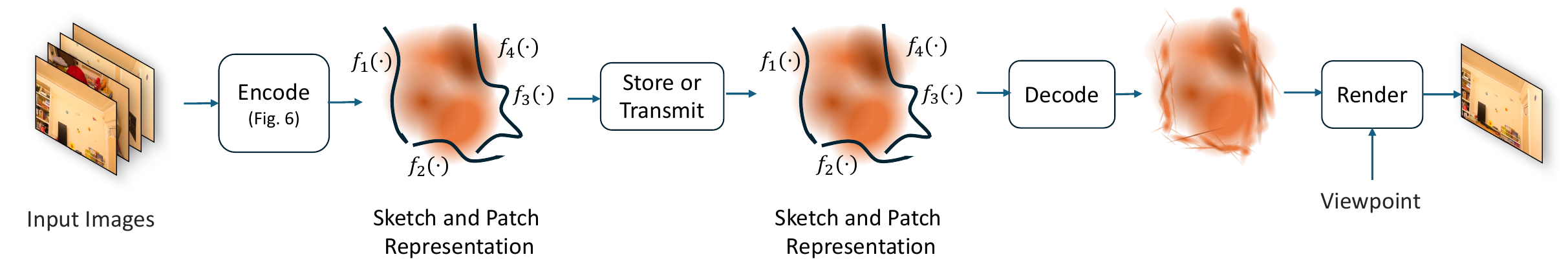}%
    \caption[]{The streaming pipeline. We propose to build an efficient representation by leveraging the 3D structure information, which can be essentially regarded as a ``codec'' in the context of the streaming system.}%
\label{fig:pipeline}
\end{figure*}

Grounded on these observations, we propose leveraging the unique structural properties of SketchGS, which typically represent high-frequency and boundary-defining features that are inherently linear or curvilinear. 
Specifically, instead of explicitly densely populating these areas with SketchGS, we propose encoding the coherent SketchGS into compact parametric models, to preserve both the visual quality and semantic information with data of significantly smaller size. 
For smoother and low-frequency regions, the PatchGS is stored with Gaussian parameters, given the naturally sparse distribution of Gaussians in these areas.
This dual-role categorization forms the basis of our hybrid Gaussian representation, and is designed to enhance scalability, structure awareness, and storage efficiency for 3DGS. Moreover, the semantic separation between boundary-defining SketchGS and volumetric PatchGS naturally enables layered progressive streaming, where the compact SketchGS layer can be transmitted first to establish the scene's structural skeleton, followed by incremental PatchGS layers that progressively refine visual detail according to available bandwidth (see \cref{fig:pipeline}).

\subsection{Approach Overview} \label{subsec:intro-overview}

In our previous work~\cite{shi2025sketch}, we introduced Sketch\&Patch (S\&P in short), a method that utilizes these distinct characteristics for efficient 3D Gaussian representation. 
We proposed identifying SketchGS using 3D line segments extracted via Line3D++~\cite{hofer2015line3D,hofer2017efficient}. By abstracting coherent SketchGS along these line segments into polynomial models and applying targeted optimization to PatchGS, our representation maintains the visual quality with only \SI{2.3}{\percent} of the original model size on man-made scenes. 

However, this approach had a fundamental limitation.
The reliance on Line3D++, which explicitly reconstructs 3D line segments from multi-view 2D line detections through matching and triangulation, restricted the applicability primarily to scenes with linear geometric features.
Line3D++ is specifically designed to abstract edges and contours into straight line segments, making it highly effective for man-made environments with well-defined edges and planar surfaces.
However, natural scenes, organic objects, and environments with curved boundaries or irregular structures cannot be adequately represented by piecewise linear approximations, limiting the generalizability of our previous approach beyond architectural and manufactured environments.
Furthermore, Line3D++ employs a multi-stage pipeline grounded in 2D image analysis to estimate 3D line segments. 
As reported in~\cite{yang2024linegs}, this cascading estimation process introduces multiple sources of uncertainty and error accumulation when performed on natural scenes with curved boundaries and irregular structures, resulting in incomplete scene coverage and degraded reconstruction quality. 

To overcome this limitation and extend the applicability of our hybrid Gaussian representation to arbitrary 3D scenes, we propose a fundamentally different approach, Sketch\&Patch++ (S\&P++ in short), that introspectively analyzes the 3DGS representation itself, rather than imposing external geometric assumptions. 
Our key insight is that the coherence patterns of boundary-defining Gaussians (\ie, their spatial clustering and attribute consistency) emerge directly from 3DGS optimization dynamics as shown in \cref{fig:sketch_patch_stats}, and are independent of whether underlying scene geometry follows straight edges or curves.
The same arguments and observations were also made in recent 3DGS works~\cite{chelani2025edgegaussians,gao2025curve,ying2025sketchsplat,yang2024linegs}.
Rather than estimating 3D structures from uncertain 2D observations, we directly \gm{analyze} the converged 3D Gaussian distribution, \gm{to understand the} scene structure \gm{in 3D}.

Our categorization method consists of two parts. First, we develop a hierarchical adaptive clustering framework that identifies SketchGS through multi-dimensional analysis.
Our specially designed multi-criteria density-based clustering simultaneously considers spatial proximity, directional alignment of Gaussian orientations, and color similarity. 
Second, we introduce an adaptive quality-driven refinement mechanism. For each identified cluster, the performance of the polynomial regression modeling is evaluated.
Clusters that exceed quality thresholds are automatically subdivided using a residual-based splitting strategy that groups Gaussians with similar modeling errors. 
The partitioning is iteratively refined until each cluster achieves satisfactory encoding quality. 
Finally, clusters containing fewer than a minimum threshold of Gaussians are filtered out, as the parametric model overhead would outweigh compression benefits for small clusters.
This creates a feedback loop where partitioning adapts not merely to spatial density, but to actual parametric encoding effectiveness.

This hierarchical framework offers fundamental advantages over line-segment-based methods. First, by operating on the optimized Gaussian distribution rather than 2D images, we eliminate the cascading error propagation inherent to multi-view 2D segment reconstruction pipelines. Second, whether boundaries follow straight lines, smooth curves, or irregular contours, the clustering tend to identify coherent Gaussian concentrations based on their emergent properties rather than presumed geometric primitives. This enables the applicability to both man-made and natural scenes. Third, our adaptive refinement ensures that the resulting clusters are optimally suited for parametric encoding, balancing cluster size against modeling accuracy to achieve efficient compression without sacrificing visual quality.
Once identified, SketchGS is encoded using parametric models that capture smooth attribute variations along coherent clusters, while PatchGS undergoes pruning and compression. The result is a generalized hybrid representation that achieves better trade-off between storage efficiency and visual quality across diverse scene types, while uniquely enabling adaptive streaming through its semantically meaningful layered structure --- a capability not offered by existing compression methods that lack such decomposition.

Beyond the generalized categorization, S\&P++ introduces additional compression techniques to further reduce storage requirements.
Specifically, we first apply tailored vector quantization~\cite{papantonakis2024reducing} to PatchGS attributes, exploiting the fact that low-frequency volumetric regions tolerate aggressive quantization without perceptual degradation.
Furthermore, we develop a cascaded bitstream compression and serialization pipeline, including half-precision conversion for polynomial coefficients, Draco geometry compression~\cite{draco} for SketchGS positions, and entropy coding, to systematically exploit redundancies at multiple levels.
These enhancements, combined with the generalized categorization framework, enable compression ratios of $116\times$ to $175\times$ compared to vanilla 3DGS.

Our contributions can be summarized as follows.
\begin{itemize}
    \item Sketch\&Patch++, a novel and generalized dual-role categorization method of 3DGS into Sketch and Patch Gaussians, reflecting their distinct functions in scene representation. It extends our hybrid representation approach beyond man-made scenes to arbitrary scenes.
    \item A hybrid and efficient Gaussian representation that \emph{i}) semantically encodes high-frequency boundary-defining features with linear/curvilinear parametric approximation, \emph{ii}) preserves volumetric detail for low-frequency regions using optimized Gaussians with selective pruning and quantization, and \emph{iii}) enables layered progressive streaming through the semantic separation of structural and volumetric content.
    \item Comprehensive experimental validation demonstrating that our generalized method achieves consistent improvements across diverse scene types, improving visual quality by up to \SI{1.74}{\decibel} in PSNR, \SI{6.7}{\percent} in SSIM, and \SI{41.4}{\percent} in LPIPS at equivalent model sizes. Correspondingly, our method compresses indoor scenes with up to \SI{0.5}{\percent} of the original model size (a 175$\times$ compression ratio) while maintaining comparable visual quality.
\end{itemize}


\section{Background and Related Work}
\label{sec:related}

\subsection{3D Gaussian Splatting}
\label{subsec:3D_gaussian_splatting}

3D Gaussian splatting (3DGS)~\cite{kerbl20233D} proposes a 3D reconstructed model capable of novel view synthesis, known for its high reconstruction quality, faster training speed, and real-time rendering through rapid rasterization. 
The 3DGS represents a scene using a collection of Gaussians optimized to fit a set of input images, compared with a splatting-based~\cite{zwicker2001ewa} rendering of the model.

Each Gaussian is parameterized by a center position $\mathbf{\mu} \in \mathbb{R}^{3\times3}$, a 3D covariance matrix $\mathbf{\Sigma}$, an opacity $\alpha$, and a color $c$ which is described by a set of Spherical Harmonic (SH) coefficients $\mathbf{K}$. 
A Gaussian centered at $\mathbf{\mu} $ is defined as: 
$
        G(x) = e^{-\frac{1}{2}x^T\Sigma^{-1}x}.
$

To construct a collection of 3D Gaussians that accurately captures the scene’s essence, 3DGS introduces an optimization method using differentiable rendering to estimate the parameters of the 3D Gaussians, fitting a set of calibrated input images of the given scene. 
During optimization, 3DGS iteratively renders 2D images from the training views and minimizes the loss between the rendered images and the ground-truth input images. 
The loss function is a combination of the $\mathcal{L}_1$ loss and the Structural Similarity loss (SSIM)~\cite{wang2004image}, weighted by 
$
        \mathcal{L} = \lambda \cdot \mathcal{L}_1 + (1 -\lambda) \cdot \mathcal{L}_{SSIM}.
$

During the 3DGS training process, Gaussian\gm{s} adapt to better fit their surroundings. 
Density is controlled by a positional gradient threshold, balancing between the number of parameters and the fitting accuracy. 
Starting from a sparse point set generated by Structure from Motion (SfM), the model iteratively removes the Gaussian splats whose opacities are below the pre-set threshold, and densifies the Gaussian splats in regions with high positional gradients. 
In areas lacking detail, low-density Gaussians are duplicated and shifted along the gradient to create new geometry. 
In areas of excessive overlap, large Gaussians are split into smaller ones to achieve finer granularity. 
This adaptive density control refines the scene representation while managing the number of Gaussians to balance model complexity and rendering quality, as illustrated in \cref{fig:densification_illu} which gives an example of how different densification thresholds affect such balance. 

\subsection{3D Scene Structure Representations} \label{subsec:3D_scene_abstract_repres}


Our work benefits from the correlation between scene structure and the 3DGS representation (see \cref{fig:densification_illu,fig:sketch_patch_renders,fig:sketch_patch_stats}). 
In the following, we review existing methods for extracting geometric structure from 3D scenes by identifying structural features at different dimensionalities.
Just as 2D images can be decomposed into key points, edges, and homogeneous regions, 3D scenes contain structural features at multiple scales: 2D surface regions (planes or coherent surfaces), 1D features (line segments, edges, or curves), and 0D features (corners and 3D key points).
Methods for extracting these structures vary depending on the input data type (2D images versus 3D data) and the target feature representation.

In earlier methods, extracting features from 2D inputs to abstract 3D line segments typically relies on Structure from Motion (SfM) algorithms~\cite{Hartley2000,schonberger2016structure, wu2013towards}. 
SfM reconstructs 3D structures from unordered image sets by simultaneously estimating intrinsic and extrinsic camera parameters and a sparse 3D point cloud. 
With SfM results, the mapping from 2D planes to 3D space becomes feasible. 
Some methods extract features from images to identify or abstract line descriptors — 2D line segments on the image plane — using classical techniques~\cite{ipol.2012.gjmr-lsd, huang2020tp, zhang2021elsd}, or neural network-based methods such as DeepLSD~\cite{pautrat2023deeplsd}. 
These 2D descriptors are then matched across multiple views to identify the same or similar line segments visible from different viewpoints. 
This process is influenced by the range of viewpoint coverage and occluded areas. 
Once matching is complete, geometric methods like triangulation~\cite{baillard1999automatic, zhang2014structure, ramalingam2015line, micusik2017structure} or other epipolar constraints~\cite{hofer2014improving, hofer2015line3D, hofer2017efficient} are used to back project the segments into 3D space. 
Some approaches further leverage deeper geometric relationships, such as planes~\cite{wei2022elsr}. 
Also, recent methods~\cite{ye2023nef, li20243d, xue2024neat} extract 3D data containing only edge regions from 2D images to reduce noise inherent to reconstructed 3D data. 
Despite significant progress in recent works, abstracting 2D data \gm{edges} into 3D \gm{edges/curves} still faces inherent limitations. 
These include robustness issues introduced by geometric computation errors.

Scene \gm{structure} can also be achieved directly from 3D point clouds or other 3D data structures. 
Such methods directly classify and identify edge regions within 3D point clouds, which are then abstracted or parameterized into line segments or curves. 
Processing directly on 3D data avoids occlusion issues caused by multi-view matching but faces challenges from the inherent noise in 3D data and the complexities of parameter tuning \gm{for feature detection}.
Some works~\cite{schnabel2007efficient, chen2017fast, wang2020pie} address these challenges through point filtering steps, robust loss functions, and parameterized estimation techniques that improve robustness to noise.
Methods~\cite{liu2021pc2wf, ma20223d} leverage deep neural networks trained on large annotated datasets to learn hidden edge features within point cloud distributions, ultimately predicting a 3D wireframe.

While these methods provide valuable structural representations for various applications, they typically require explicit feature detection pipelines, which either through geometric algorithms or learned models, and may not generalize well across diverse scene types.
In contrast, our approach directly analyzes the distribution of optimized 3D Gaussians to infer structural characteristics, leveraging the inherent clustering patterns that emerge from 3DGS optimization dynamics without requiring explicit geometric priors or feature detection.

\subsection{Compact 3DGS Representation} \label{subsec:related-compact3dgs}

While 3DGS~\cite{kerbl20233D} leverages efficient techniques such as anisotropic Gaussians~\cite{zwicker2001ewa}, tile-based sorting, and approximate $\alpha$-blending to achieve high-performance rendering, the presence of redundancy in the set of  Gaussian splats continues to impact computational efficiency.
Minimizing these redundancies is crucial for optimizing model performance across various applications.

The model size in 3DGS is primarily affected by two factors. 
First, complex and high-frequency areas require sophisticated parameters to represent.
In particular, advanced spherical harmonics coefficients are needed for accurate modeling. 
Second, as discussed in~\cref{subsec:3D_gaussian_splatting}, the model generation process introduces additional Gaussian splats based on density thresholds during scene fitting.
The combination of these factors
leads to significant storage overhead.

Recent research has approached these challenges from multiple angles.
To address parameter complexity, several methods have been proposed: region-based vector quantization~\cite{niedermayr2024compressed}, K-means codebooks~\cite{navaneet2023compact3d}, and view-direction exclusion~\cite{cai2025radiative}. Fan \etal~\cite{fan2023lightgaussian} employ knowledge distillation~\cite{hinton2015distilling} to compress spherical harmonics parameters, while Lee \etal~\cite{lee2024compact} utilize learned binary masks and grid-based neural networks as alternatives to spherical harmonics.
Based on the anchor-based representation constructed by Scaffold-GS~\cite{lu2024scaffold}, HAC~\cite{chen2024hac} and HAC++~\cite{Chen2025} leverages 3D coordinates to guide quantization and entropy coding, while ContextGS~\cite{globersons2024contextgs} encodes the anchors hierarchically.

Parallel efforts have focused on managing Gaussian density~\cite{ren2024octree,lu2024scaffold,kerbl2024hierarchical,shi2025lapisgs,mallick2024taming3dgs,zhang2025gaussianspa}. 
Some works utilize the technique of Level of Detail (LOD).
For instance, Ren \etal~\cite{ren2024octree} propose an octree-based organization where each level corresponds to anchor Gaussian splats defining different LODs. 
Their approach, enhanced by Scaffold-GS~\cite{lu2024scaffold}, combines anchor Gaussians with MLPs for anchor-level feature estimation, resulting in improved representation efficiency. 
We introduce LapisGS~\cite{shi2025lapisgs}, a layered representation scheme enabling progressive and continuous quality adaptation for bandwidth-aware streaming applications.
Mallick \etal applies selective densification based on pixel saliency and gradients to reduce the redundancy of 3DGS~\cite{mallick2024taming3dgs}.
Similarly, Zhang \etal propose GaussianSpa~\cite{zhang2025gaussianspa} which constraints 3DGS simplification during training to force Gaussians to distribute sparsely. 

These methods, however, primarily focus on either parameter compression or LOD structuring without considering the inherent roles of different Gaussian splats in scene representation. 
Our approach fundamentally differs by recognizing and leveraging the distinct characteristics of Gaussians, categorizing them into Sketch and Patch components based on their geometric significance. 
This structure-aware strategy enables more efficient representation by applying appropriate encoding techniques to each category: compact parametric models for boundary-defining features and optimized pruning for volumetric regions.
Meanwhile, our Sketch and Patch categorization is complementary to these existing compression and LOD techniques. 
Parameter compression methods like vector quantization~\cite{niedermayr2024compressed}, codebook learning~\cite{navaneet2023compact3d}, or SH compression~\cite{fan2023lightgaussian,lee2024compact} can be applied to SketchGS and PatchGS while maintaining their distinct roles. 
This compatibility ensures that our method can serve as a foundation for further optimizations while providing its unique benefits in storage efficiency and visual quality preservation.


\begin{figure*}[t!]%
    \centering
    \includegraphics[width=\textwidth]{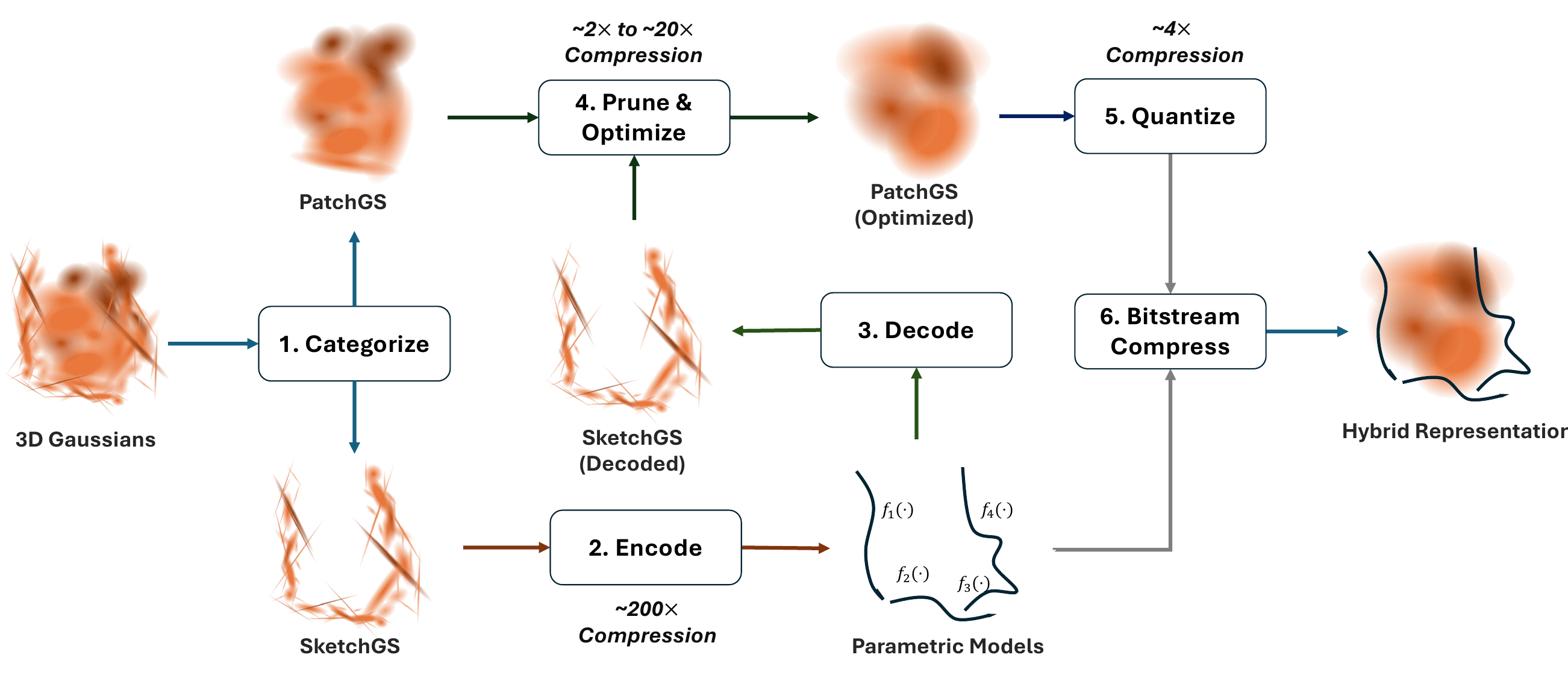}%
    \caption[]{
    Starting from the original 3DGS, our method first \textit{categorizes}~\sg{(1)} Gaussians into SketchGS and PatchGS (\cref{subsec:scene_abstract}). 
    SketchGS, which capture high-frequency boundary-defining features with spatial-attribute coherence, are \textit{encoded}~\sg{(2)} using parametric polynomial models, achieving up to $\sim200\times$ compression. 
    For smoother regions, Patch Gaussians are \textit{pruned and retrained}~\sg{(4)} together with the \textit{decoded}~\sg{(3)} Sketch Gaussians to ensure compactness ($\sim2\times$ to $\sim20\times$ compression) while preserving visual fidelity. \textit{Vector quantization}~\sg{(5)} is then applied to further reduce the storage of the optimized PatchGS with additional $\sim4\times$ compression.
    The resulting hybrid representation is finally packed into a unified \textit{bitstream}~\sg{(6)} with cascaded compression, enabling efficient storage \gm{coding}, progressive streaming, and practical deployment in networked rendering applications 
    (\cref{subsec:encode_optimize}).
    }%
\label{fig:methodology}
\end{figure*}

\section{Proposed Methodology}
\label{sec:method}

While 3DGS can represent complex 3D scenes, the globally estimated set of Gaussian splats is non-uniform due to heterogeneous attributes; also, the adaptive density control mechanism creates numerous Gaussian splats, which leads to storage inefficiencies and thus challenges the need for more scalable and efficient representations.

To address these challenges, we propose S\&P++, a hybrid Gaussian representation that categorizes Gaussians into SketchGS and PatchGS. 
SketchGS captures high-frequency, boundary-defining features and is compactly encoded using parametric models based on their spatial-attribute coherence, achieving up to $248\times$ compression ratio. 
In contrast, PatchGS represents low-frequency, smoother regions and is optimized for storage efficiency through retraining and quantization. 
This categorization allows us to leverage the structural properties of the scene and its correlation to 3DGS density and bias (\cref{fig:densification_illu}), achieving a representation that balances scalability, efficiency, and visual quality.

\sg{As illustrated in \cref{fig:methodology}}, S\&P++ consists of the following steps. 
First, we identify coherent clusters of Gaussian splats that exhibit structural regularity across spatial, geometric, and appearance dimensions using a multi-criteria density-based clustering approach (\cref{subsec:scene_abstract}). 
Unlike our previous work~\cite{shi2025sketch}, which relied on explicit 3D line segment extraction as a prior for identifying SketchGS, the new approach \gm{classifies} Gaussians directly from their intrinsic properties, enabling a more flexible and generalizable partitioning strategy.
The identified clusters are then refined through an adaptive splitting process guided by polynomial modeling performance, which distinguishes geometrically structured and boundary-defining clusters (SketchGS) from scattered PatchGS.
The Sketch Gaussians are then encoded using parametric models that compactly represent their attributes, significantly reducing storage requirements.
For smoother regions, represented by Patch Gaussians, \gm{we first apply pruning to sparsify their distribution and \sg{then} retraining to maintain visual fidelity. 
Since they represent coherent regions, we} perform vector quantization to further reduce the model size (\cref{subsec:encode_optimize}). 
Finally, all encoded components are serialized into a unified bitstream format with cascaded compression for each component type.

\subsection{Categorizing Sketch and Patch Gaussians} 
\label{subsec:scene_abstract} 
The first step of the pipeline analyzes the 3D scene to identify Gaussians \sg{with} coherent relationships and partitions them into SketchGS and PatchGS.
This partitioning is critical to achieving an effective hybrid representation, as it directly determines which Gaussians benefit from parametric encoding versus alternative compression strategies.

\textbf{Motivation for Clustering-Based Partitioning}.
In our previous work~\cite{shi2025sketch}, we employed a line-guided approach that first extracted 3D line segments from 2D input images using Line3D++~\cite{hofer2017efficient}, then identified SketchGS through radius search and RANSAC-based filtering around these pre-defined 3D lines. 
While this line-based method effectively captured Gaussians along \gm{linear} edges and contours, it imposed several limitations. 
First, the approach is inherently constrained by the quality and completeness of the line extraction algorithm. 
Line3D++ is specially designed for man-made scenes with sharp, clear edges and boundaries, which cannot well detect curved features, irregular boundaries, or regions where 2D line detection is unreliable (\eg, natural scenes or organic objects). 
Second, the dependency on 2D-to-3D projection can introduce errors in scenarios with sparse views or challenging camera configurations.

To overcome these limitations, we introduce a clustering-based partitioning strategy that directly analyzes the Gaussians themselves, discovering structured configurations without relying on external geometric priors. 
This data-driven approach is more robust to scene variations and naturally adapts to the local structure of the Gaussian distribution.
By identifying clusters that exhibit multi-dimensional coherence (spatial proximity, directional alignment, and color similarity), we can discover not only linear features but also more complex structured patterns that do not align perfectly with straight line segments.

\textbf{Multi-Criteria Density-Based Clustering}.
We propose an extension of the Density-Based Spatial Clustering of Applications with Noise (DBSCAN) algorithm~\cite{ester1996density} that simultaneously evaluates multiple complementary criteria to identify coherent Gaussian clusters. 
Unlike conventional DBSCAN, which considers only spatial proximity, our approach leverages three distinct distance metrics that capture different aspects of Gaussian splat coherence.
Specifically, for each Gaussian splat $\mathcal{G}_i$, we define three complementary distance measures:
\begin{itemize}
    \item \textit{Spatial distance}. The Euclidean distance between Gaussian centers in 3D space captures geometric proximity: $d_{\text{spatial}}(\mathcal{G}_i, \mathcal{G}_j) = \|\mathbf{\mu}_i - \mathbf{\mu}_j\|_2$, where $\mathbf{\mu}_i \in \mathbb{R}^3$ represents the position of splat $\mathcal{G}_i$.
    \item \textit{Directional distance}. To capture orientation coherence, which is particularly important for Gaussians concentrated along edges and boundaries, we compute a cosine similarity: $d_{\text{direction}}(\mathcal{G}_i, \mathcal{G}_j) = 1 - (\frac{ {\mathbf{n}_i} \cdot {\mathbf{n}_j} }{||{\mathbf{n}_i}|| ||{\mathbf{n}_j}||} + 1) / 2$, where ${\mathbf{n}_i}$ represents the normalized principal direction derived from the Gaussian's covariance structure. The resulting similarity ranges from \SI{0}, meaning exactly the same, to +\SI{1}, meaning exactly the opposite.
    \item \textit{Color distance}. We incorporate perceptual color similarity using RGB features: $d_{\text{color}}(\mathcal{G}_i, \mathcal{G}_j) = \|{\mathbf{c}}_i - {\mathbf{c}}_j\|_2$, where ${\mathbf{c}}_i$ represents the RGB color vector of Gaussian $\mathcal{G}_i$.
\end{itemize}

We define Gaussian $\mathcal{G}_j$ as a neighbor of $\mathcal{G}_i$ if and only if all three distance constraints are simultaneously satisfied:
\begin{equation}
    \mathcal{G}_j \in \mathcal{N}(\mathcal{G}_i) \Leftrightarrow 
    \left( d_{\text{spatial}}(\mathcal{G}_i, \mathcal{G}_j) \leq \epsilon_s \right) \land 
    \left( d_{\text{direction}}(\mathcal{G}_i, \mathcal{G}_j) \leq \epsilon_d \right) \land 
    \left( d_{\text{color}}(\mathcal{G}_i, \mathcal{G}_j) \leq \epsilon_c \right),
\end{equation}
where $\epsilon_s$, $\epsilon_d$, $\epsilon_c$ are the spatial threshold, directional threshold, and color threshold, respectively.
The use of separate epsilon thresholds ($\epsilon_s$, $\epsilon_d$, $\epsilon_c$) provides fine-grained control over each criterion's influence on the clustering outcome.
In contrast to the single radius parameter $r$ in our previous line-based method~\cite{shi2025sketch}, this multi-criteria formulation offers greater flexibility in capturing different types of structural coherence.

As a 3DGS scene usually consists of millions of Gaussians, to accelerate the neighborhood queries, we employ a two-stage filtering strategy. 
First, we \gm{partition Gaussian positions into a $k$-d tree to} retrieve all splats within the spatial epsilon threshold $\epsilon_s$, leveraging the efficient $\mathcal{O}(\log n)$ nearest neighbor search.
Second, for the spatially proximate candidates, we compute the directional and color distances and retain only those splats satisfying all three epsilon constraints.
This hierarchical filtering reduces computational complexity from $\mathcal{O}(n^2)$ to approximately $\mathcal{O}(n \log n + nk)$, where $k$ is the average number of spatial neighbors.

Following standard DBSCAN terminology, we classify Gaussians as core points (with at least \texttt{min\_samples} neighbors satisfying all three criteria), border points (non-core Gaussians within the neighborhood of a core point), or noise points (Gaussians not belonging to any cluster).
Clusters are formed by connecting core points and their neighborhoods through a breadth-first traversal, ensuring density-connected components are merged into \gm{coherent} groups.

\textbf{Adaptive Polynomial Regression-Based Refinement}.
While multi-criteria DBSCAN effectively identifies spatially and perceptually coherent groups, not all clusters exhibit the characteristic of SketchGS. 
To distinguish geometrically structured clusters from scattered point configurations, we introduce an adaptive refinement stage based on polynomial modeling performance.
This strategy is inspired by our previous work~\cite{shi2025sketch}, where we employed random sample consensus (RANSAC)~\cite{fischler1981random} to filter out Gaussians with large modeling error from SketchGS, to ensure modeling coherence.
However, rather than using RANSAC for outlier rejection within line segments, we now use polynomial modeling quality as a criterion for adaptive cluster splitting and refinement.

\textit{Polynomial Regression Modeling}.
For each cluster $\mathcal{C}$ obtained from multi-criteria DBSCAN, we model the relationship between Gaussian positions and their attributes using polynomial regression (PR). 
Similar to our previous approach~\cite{shi2025sketch}, we fit separate polynomial models for four key attributes: scaling, rotation (quaternion), opacity, and color.
Specifically, given the normalized 3D positions $\mathbf{X} \in \mathbb{R}^{n \times 3}$ of splats in cluster $\mathcal{C}$, we fit polynomial models, where the degree of polynomial is chosen by grid search in the range of $[1,10]$:
$
    \mathbf{a} = f_a(\mathbf{X}) + \boldsymbol{\epsilon}_a,
$
where $\mathbf{a}$ represents one of the four attributes and $f_a(\cdot)$ is the polynomial model trained using least-squares regression on polynomial features generated by:
$\Phi(\mathbf{X}) = [1, x_1, x_2, x_3, x_1^2, x_1x_2, \ldots, x_1^d, x_2^d, x_3^d].$
The number of coefficients for a degree-$d$ polynomial in 3D is $\frac{1}{6}(d+1)(d+2)(d+3)$, yielding at most $286$ coefficients per attribute model.

We evaluate the quality of polynomial regression models using Mean Squared Error (MSE):
$
    \text{MSE}_{\text{attr}} = \frac{1}{n} \sum_{i=1}^{n} \|\hat{\mathbf{a}}_i - \mathbf{a}_i\|_2^2,
$
where $\hat{\mathbf{a}}_i$ represents the predicted attribute and $\mathbf{a}_i$ the actual attribute. 
A combined quality metric aggregates individual attribute MSEs:
$
    \text{MSE}_{\text{combined}} = \sum_{\text{attr}} w_{\text{attr}} \cdot \text{MSE}_{\text{attr}},
$
with equal weights $w_{\text{attr}} = 0.25$ for each of the four attributes (scaling, rotation, opacity, and color) in our implementation. 

Clusters exhibiting low polynomial regression error ($\text{MSE}_{\text{combined}} < \tau_{\text{max}}$) demonstrate strong correlations between positions and attributes, indicating geometric structure characteristic of edge-aligned or boundary-defining properties.  
Conversely, high MSE suggests scattered, patch-like arrangements where attributes vary irregularly and resist compact parametric representations.

\textit{Iterative Cluster Splitting}.
For clusters failing to meet the quality threshold $\tau_{\text{max}}$, we perform adaptive splitting to isolate coherent sub-regions. 
The splitting process leverages polynomial regression residual patterns:
\begin{enumerate}
    \item \textit{Residual feature construction}. For cluster $\mathcal{C}$ with $n$ Gaussians, we compute absolute residuals for each attribute:
    $
        \mathbf{R}_{\text{attr}} = |{\mathbf{a}}_{\text{pred}} - {\mathbf{a}}_{\text{actual}}|,
    $ 
    where the ${\mathbf{a}}_{\text{pred}}$ and ${\mathbf{a}}_{\text{actual}}$ are normalized.
    We concatenate residuals across all attributes to form an $n \times d_{\text{residual}}$ feature matrix $\mathbf{R}$. 
    
    \item \textit{Spatial-residual feature fusion}. To maintain spatial locality while grouping Gaussians with similar modeling errors, we construct a combined feature representation:
    $
        \mathbf{F} = [\beta \cdot \mathbf{R} \;||\; (1-\beta) \cdot {\mathbf{X}}],
    $
    where ${\mathbf{X}}$ contains positions, and $\beta$ weight the residual and spatial components. 
    
    \item \textit{K-means sub-clustering}. We apply K-means clustering on $\mathbf{F}$ to partition $\mathcal{C}$ into $k$ sub-clusters, where $k \in [2, \min(4, \lceil \text{MSE}_{\text{combined}}/\tau_{\text{max}} \rceil + 1)]$ is dynamically determined based on the severity of modeling failure.
    
    \item \textit{Recursive refinement}. Each sub-cluster is re-evaluated with polynomial regression. Sub-clusters meeting the quality threshold are accepted as SketchGS candidates, while those exceeding $\tau_{\text{max}}$ undergo further splitting.
\end{enumerate}

This recursive refinement continues until clusters either achieve satisfactory polynomial regression performance, reach the minimum cluster size threshold $S_{\min}$, or exceed the maximum iteration limit $T_{\max}$.
The adaptive splitting process effectively balances modeling fidelity with compression efficiency, ensuring that only clusters with strong geometric coherence are classified as SketchGS.

\textbf{Post-processing and Outlier Removal}.
Even within validated SketchGS clusters, we observe that the polynomial model for scaling attribute
tends to introduce a bias, which can result in poorly modeled long and thin Gaussians. We introduce a post-processing phase to address a critical issue caused by the polynomial regression for scaling.
To mitigate this, we decode the SketchGS candidates and use the Interquartile Range (IQR) method to identify outlier Gaussians with extreme scaling values that deviate from the expected distribution.
This post-processing ensures that only visually consistent Gaussians are retained for the later encoding, improving the accuracy of the SketchGS representation and ensuring better alignment with the underlying scene structure.
Gaussians that are filtered out during this process are not discarded but instead reclassified as PatchGS, ensuring that no geometric information is lost while maintaining the most appropriate representation for each Gaussian based on its characteristics.

\textbf{\gm{Gaussians Categorization}}.
Through this hierarchical pipeline, Gaussians are naturally split into two categories:
\begin{itemize}
    \item \textit{SketchGS}. Clusters that satisfy multi-criteria DBSCAN coherence, exhibit low polynomial regression error, and meet size constraints.
    These clusters typically concentrate along object boundaries, edges, and geometric discontinuities where Gaussians align in structured configurations. 
    Compared to our previous line-based approach~\cite{shi2025sketch}, this clustering-based method can identify a broader range of structured patterns,
    including curved features and complex boundaries that may not align perfectly with straight line segments. 
    Specifically, up to \SI{88.7}{\percent} and \SI{65.9}{\percent} Gaussians are classified as SketchGS for indoor scenes and for outdoor scenes, respectively, which are $1.2 \times$ and $707.8 \times$ more compared to our previous work.
    \item \textit{PatchGS}. 
    All remaining Gaussians that either failed to cluster under the multi-criteria DBSCAN constraints, belonged to clusters with persistently high polynomial regression error despite splitting, or were excluded due to size constraints of DBSCAN.
    These Gaussians typically \gm{represent} smooth surface regions with less geometric structure and are sparsely distributed in the 3DGS representation. 
\end{itemize}

In \cref{tab:sketch_patch_stats}, we show the average model size and relative ratio of raw SketchGS and PatchGS, which are classified by our clustering method, across different datasets. Note that the raw SketchGS and PatchGS are not further encoded and optimized, which will be detailed in \cref{subsec:encode_optimize}.

\begin{table}[hbpt]
        \caption{The average model size (MB) of raw SketchGS and raw PatchGS across different datasets. The relative ratio of each component is indicated.
        } 
        \label{tab:sketch_patch_stats}
        \vspace{0.5em}
        \centering
            \begin{tabular}{l|c|c|c}
                \toprule
                \textbf{Model} & \textbf{Deep Blending} & \textbf{Tanks\&Temples} & \textbf{Mip-NeRF360} \\
                \midrule
                $\bullet$ Vanilla 3DGS  & $703.77$ (\SI{100}{\percent}) & $436.50$ (\SI{100}{\percent}) & $738.74$ (\SI{100}{\percent}) \\
                \hspace{2mm}$\circ$ Raw PatchGS & $142.14$ (\SI{20.2}{\percent}) & $245.07$ (\SI{56.1}{\percent}) & $290.14$ (\SI{39.3}{\percent}) \\
                \hspace{2mm}$\circ$ Raw SketchGS & $579.63$ (\SI{79.8}{\percent}) & $191.43$ (\SI{43.9}{\percent}) & $448.60$ (\SI{60.7}{\percent}) \\
                \bottomrule
            \end{tabular}
\end{table}

\subsection{Gaussian Encoding and Optimization} \label{subsec:encode_optimize}

Following the hierarchical clustering and categorization process described in \cref{subsec:scene_abstract}, we obtain two distinct Gaussian groups with fundamentally different characteristics: SketchGS exhibiting strong correlations between their 3D positions and Gaussian attributes along boundary-defining features, and PatchGS distributed sparsely across broader surface regions capturing low-frequency appearance variations.

To efficiently compress these categorized Gaussians, we employ a compression pipeline that tailors encoding strategies to the structural properties of each category.
For SketchGS, we leverage PR models to compactly encode their attributes by exploiting the coherent relationships between positions and attributes within clusters.
For PatchGS, we apply a combination of pruning-based optimization to reduce redundancy, followed by vector quantization to compress the remaining Gaussian attributes.
Finally, we apply cascaded bitstream compression and serialization to further exploit redundancies.
This compression framework builds upon our previous work~\cite{shi2025sketch}, which established the effectiveness of polynomial regression for SketchGS and pruning-based optimization for PatchGS.
These core strategies remain unchanged in this extension, as they have proven robust for their respective Gaussian categories.
The key innovations introduced in this extended version are twofold.

First, we incorporate vector quantization~\cite{papantonakis2024reducing} for PatchGS attributes, motivated by the observation that these Gaussians represent low-frequency appearance variations in smooth regions where aggressive quantization introduces minimal perceptual artifacts.
Second, we introduce a cascaded bitstream compression and serialization pipeline that systematically compresses polynomial coefficients, SketchGS positions, and quantized attributes through specialized techniques (half-precision conversion, Google's Draco geometry compression, and entropy coding), followed by final gzip compression to capture inter-component redundancies.

\textbf{Sketch Gaussian Encoding}.
After identifying SketchGS through our clustering-based categorization, we encode their attributes efficiently while preserving visual quality. 
The coherent relationships between Gaussian positions and their attributes within each cluster, strengthened by the robust multi-criteria DBSCAN and adaptive PR-based refinement detailed in \cref{subsec:scene_abstract}, enable compact representation through polynomial modeling.
As shown in \cref{tab:sketch_stats}, this approach achieves approximately $20\times$ compression for SketchGS across different datasets, with the PR models themselves occupying only \SI{6}{\percent} to \SI{10}{\percent} of the encoded representation while Gaussian positions constitute the remaining over \SI{90}{\percent}.

\begin{table}[hbpt]
        \caption{The average size (MB) and relative ratios of each component of PR-encoded SketchGS.
        } 
        \label{tab:sketch_stats}
        \vspace{0.5em}
        \centering
            \begin{tabular}{l|c|c|c}
                \toprule
                \textbf{Model} & \textbf{Deep Blending} & \textbf{Tanks\&Temples} & \textbf{Mip-NeRF360} \\
                \midrule
                    \hspace{0mm}$\bullet$ Raw SketchGS & $579.63$ & $191.43$ & $448.60$ \\
                \midrule
                    \hspace{0mm}$\bullet$ PR-Encoded SketchGS  & $31.37$ (\SI{100}{\percent}) & $10.42$ (\SI{100}{\percent}) & $29.80$ (\SI{100}{\percent}) \\
                        \hspace{2mm}$\circ$ Gaussian Positions & $29.32$ (\SI{93.5}{\percent}) & $9.45$ (\SI{90.7}{\percent}) & $26.77$ (\SI{89.8}{\percent}) \\
                        \hspace{2mm}$\circ$ PR Models & $2.05$ (\SI{6.5}{\percent}) & $0.97$ (\SI{9.3}{\percent}) & $3.03$ (\SI{10.2}{\percent}) \\
                \bottomrule
            \end{tabular}
\end{table}

The effectiveness of PR for SketchGS stems from the predictable relationships between 3D positions and Gaussian attributes within coherent clusters.
For each cluster $\mathcal{C}$, we fit PR models that predict Gaussian attributes (scaling, rotation, opacity, and color) from their 3D positions.
This parametric encoding transforms the storage requirement from per-Gaussian attribute values to a small set of polynomial coefficients shared across all Gaussians in the cluster, achieving substantial compression while maintaining high reconstruction accuracy.
By encoding SketchGS with polynomial models, we efficiently represent the high-frequency boundary features of the scene, which typically concentrate significant storage in the original 3DGS model, with minimal memory footprint.
This compact representation not only reduces storage requirements but also enables adaptive processing, making it suitable for scalable 3D scene reconstruction and rendering applications.

Importantly, our clustering-based approach to identifying SketchGS allows polynomial encoding to be applied more broadly than in our previous line-based method.
The line-based approach was limited to detecting Gaussians near explicit line segments, often missing coherent structures in curved regions, organic objects, and natural scenes where linear features are sparse or absent.
In contrast, our clustering method automatically discovers geometrically coherent Gaussian groups regardless of whether they align with straight edges or curves, capturing a larger fraction of compressible Gaussians and thus achieving better overall compression ratios.
Our experiments (\cref{sec:experiments}), particularly \cref{tab:gaussian_stat}, further demonstrate the storage efficiency of SketchGS encoding across diverse scene types.

\textbf{Patch Gaussian Pruning and Optimization}.
The core idea behind PatchGS optimization is to reduce the number of Gaussians in smoother and broader regions of the 3D scene, where low-frequency variations are typically captured. 
While SketchGS efficiently encodes high-frequency features through parametric models, PatchGS handles the volumetric representation of smoother regions where a sparser distribution is sufficient. 
To this end, we leverage retraining, which allows for selective pruning of Patch Gaussians while ensuring that visual quality is maintained. The pruning and retraining methodology for PatchGS remains consistent with our previous work~\cite{shi2025sketch}, preserving a proven optimization pipeline.

The process starts with the encoding and decoding of SketchGS. 
Once the SketchGS is (compactly) encoded by PR models, we decode it and fix the decoded version for pruning and retraining, and the decoded SketchGS remains unchanged during the subsequent process.
Note that the decoded SketchGS is used during retraining, rather than the original SketchGS, allowing the PatchGS to compensate for errors introduced by the encoding-decoding process of the SketchGS.
The retraining focuses on optimizing the PatchGS, which are first pruned by randomly and uniformly removing some of them to improve the compactness of the model, and then retrained to improve the visual quality of the results.
This retraining process ensures that the PatchGS is optimized to align with the visual structure of the scene \wrt the given SketchGS.

Through pruning, retraining, and quantization, we achieve a more compact 3DGS representation where each Gaussian type serves its designated role: SketchGS efficiently captures high-frequency, boundary-defining features, while optimized PatchGS provides comprehensive coverage of smoother regions.
This dual optimization strategy significantly reduces the total number of Gaussians required to represent the scene while maintaining high fidelity. 
The result is a hybrid representation that efficiently allocates storage resources according to the geometric characteristics of different scene regions, \ie compact parametric models for high-frequency features and optimized sparse distribution for volumetric regions. 

\textbf{Vector Quantization}.
Unlike SketchGS, which captures sharp discontinuities and high-frequency boundary features requiring precise attribute values to maintain edge sharpness and structural definition, PatchGS models gradual appearance changes across broader regions where fine-grained attribute precision has minimal perceptual impact. Therefore, for additional storage efficiency, we apply vector quantization to the optimized PatchGS following Papantonakis \etal~\cite{papantonakis2024reducing}. 

This compression scheme employs K-means clustering to create codebooks for various Gaussian attributes, including opacity, scaling, quaternion rotation (real and imaginary parts), base color coefficients, and spherical harmonics color components. 
Instead of storing exact values, the indices are maintained to the nearest values in fixed-size codebooks. 
For vector attributes, separate indices are used for each component while sharing a single codebook. 
Based on the empirical experiments from Papantonakis \etal~\cite{papantonakis2024reducing}, 256-entry codebooks with 1-byte indices are used, offering optimal compression while preserving visual quality. 
Note that Gaussian positions are not compressed with codebooks because it leads to significant quality degradation, according to their pilot experiments.
Instead, 16-bit half-float quantization is applied to Gaussian positions and codebook entries to further reduce storage requirements while maintaining visual fidelity. 

\textbf{Bitstream Compression and Serialization}.
After encoding the SketchGS via polynomial modeling and optimizing the PatchGS through pruning, retraining, and vector quantization, our method produces three primary components: i) polynomial regression model coefficients, ii) SketchGS positions, and iii) quantized PatchGS attributes and positions. 
While these components are already significantly compressed compared to the original 3DGS representation, their serialized forms exhibit additional redundancy and structure that can be further exploited through specialized compression techniques.
To maximize storage efficiency, we apply a cascaded compression pipeline.

\textit{Polynomial Model Compression}.
The polynomial regression models that encode Sketch Gaussian attributes consist of floating-point coefficient arrays. 
To compress these coefficients efficiently, we first convert all full-precision floating-point coefficients to half-precision (16-bit) floats, reducing the raw storage by $2\times$ while maintaining sufficient numerical precision for accurate attribute prediction during decoding. 
Empirical analysis shows that for polynomial coefficients with typical magnitudes of \num[output-exponent-marker = e]{1e-4} (measured as mean absolute coefficient value), the absolute conversion error remains below \num[output-exponent-marker = e]{1e-8}, corresponding to a relative error of less than \SI{0.01}{\percent}.
We then apply zlib compression to the serialized half-float coefficients. 
This two-stage compression achieves an overall compression ratio of approximately $2.14\times$ for polynomial model data.

\textit{Sketch Gaussian Position Compression}.
While SketchGS positions represent only spatial coordinates without attribute information (which is encoded in the polynomial models), they constitute a substantial portion of the encoded SketchGS representation.
As shown in \cref{tab:sketch_stats}, position data typically occupy over \SI{90}{\percent} of the PR-encoded SketchGS storage, making their efficient compression critical for achieving higher compression ratios.
These 3D positions $\mathbf{\mu}_i \in \mathbb{R}^3$ exhibit strong spatial coherence within clusters, as they correspond to structured geometric features such as edges and boundaries.
To exploit this spatial coherence, we employ Google Draco~\cite{draco}, a specialized geometry compression library designed for 3D mesh and point cloud data.
Draco achieves high compression ratios through a combination of quantization, prediction-based encoding, and entropy coding. 
In our experiments, Draco achieves the compression ratio of approximately $22.3\times$ for SketchGS positions, drastically reducing the storage footprint while maintaining sufficient positional accuracy (with mean error smaller than \SI{0.04}{\percent} of the scene diameter). 

\textit{Final Bitstream Packing}.
After applying specialized compression to the polynomial models and Sketch Gaussian positions, we serialize all three components (compressed polynomial coefficients, Draco-compressed Sketch positions, and quantized Patch Gaussians) into a unified bitstream with a structured header that specifies component offsets and sizes, which is essential for practical streaming scenarios.
Despite the component-level compression already applied, the concatenated bitstream often exhibits residual redundancy at the inter-component level, such as repeated header patterns, alignment padding, and correlations between components.
To capture these remaining redundancies, we apply gzip compression to the entire packed bitstream as a final post-processing step, which provides an additional \SI{5}{\percent} reduction in storage.

Through this comprehensive cascaded compression pipeline, combining component-specific techniques (polynomial precision reduction, Draco geometry compression, vector quantization) with unified bitstream entropy coding, we achieve substantial high compression ratios while maintaining visual fidelity across diverse scene types.

\section{Experimental results}
\label{sec:experiments}

\subsection{Experimental Setup}

\begin{figure*}[t]
\centering
    \subfloat[Playroom.]{%
        \includegraphics[width=0.23\textwidth]{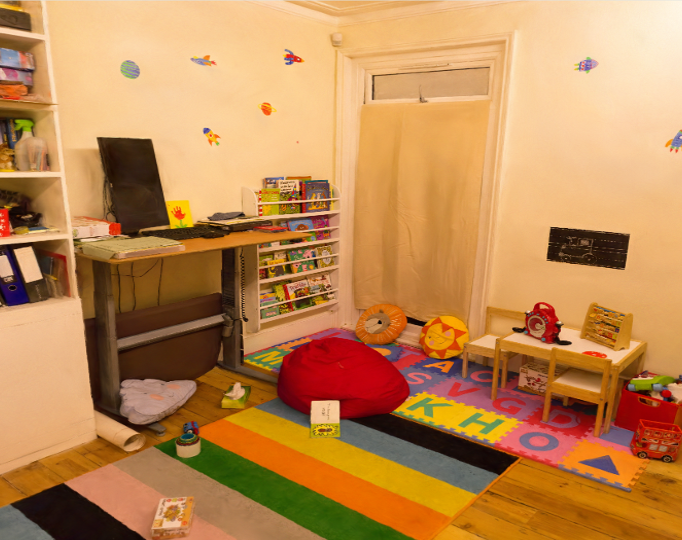}%
        \label{fig:thumb_playroom}%
        }
    \hspace{0.1em}
    \subfloat[Drjohnson.]{%
        \includegraphics[width=0.23\textwidth]{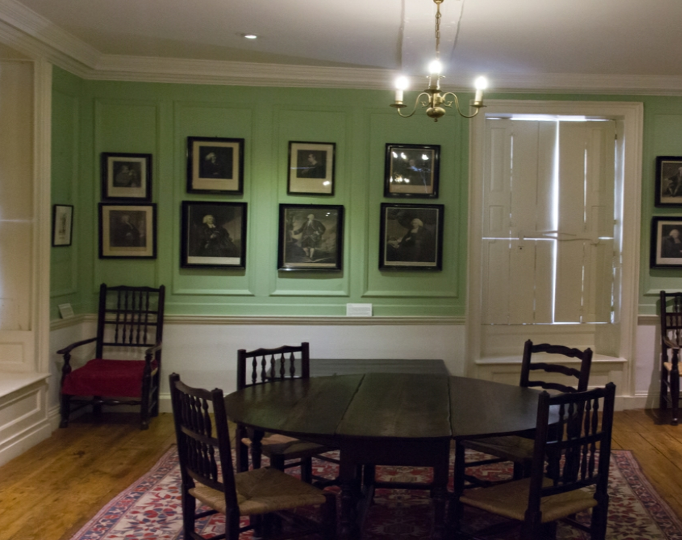}%
        \label{fig:thumb_drjohnson}%
        }
    \hspace{0.1em}
    \subfloat[Train.]{%
        \includegraphics[width=0.23\textwidth]{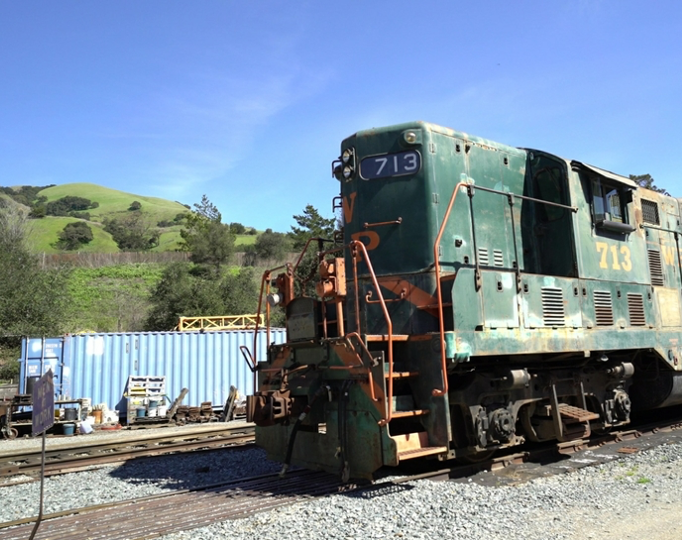}%
        \label{fig:thumb_train}%
        }
    \hspace{0.1em}
    \subfloat[Truck.]{%
        \includegraphics[width=0.23\textwidth]{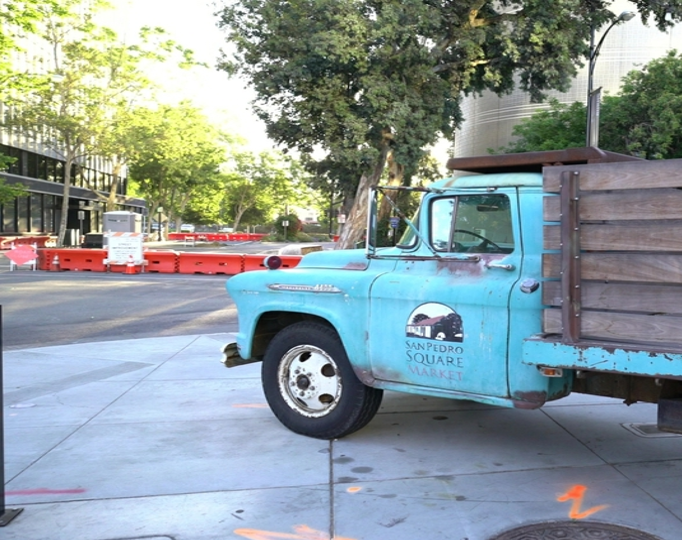}%
        \label{fig:thumb_truck}%
        } \\
    \subfloat[Kitchen.]{%
        \includegraphics[width=0.23\textwidth]{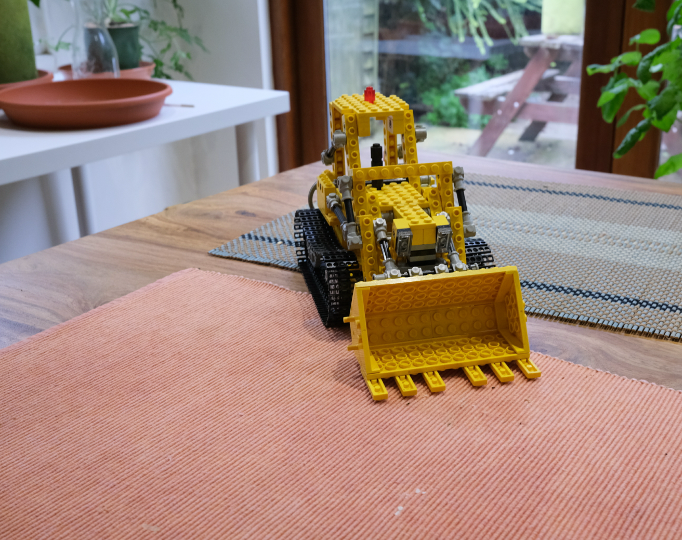}%
        \label{fig:thumb_kitchen}%
        }
    \hspace{0.1em}
    \subfloat[Room.]{%
        \includegraphics[width=0.23\textwidth]{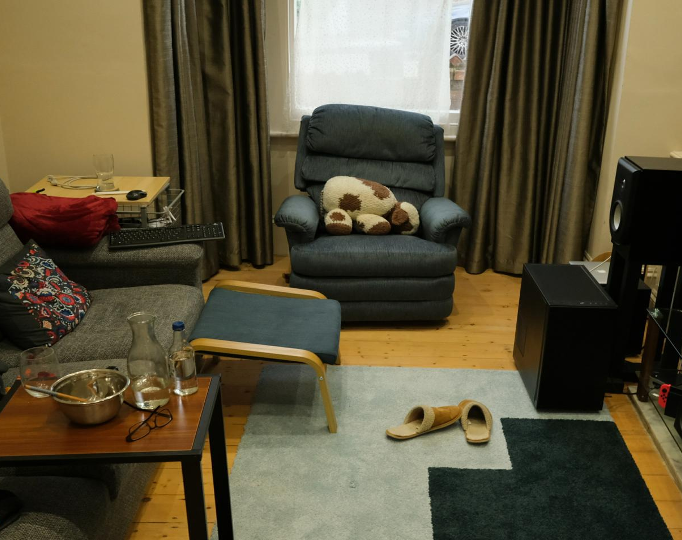}%
        \label{fig:thumb_room}%
        }
    \hspace{0.1em}
    \subfloat[Garden.]{%
        \includegraphics[width=0.23\textwidth]{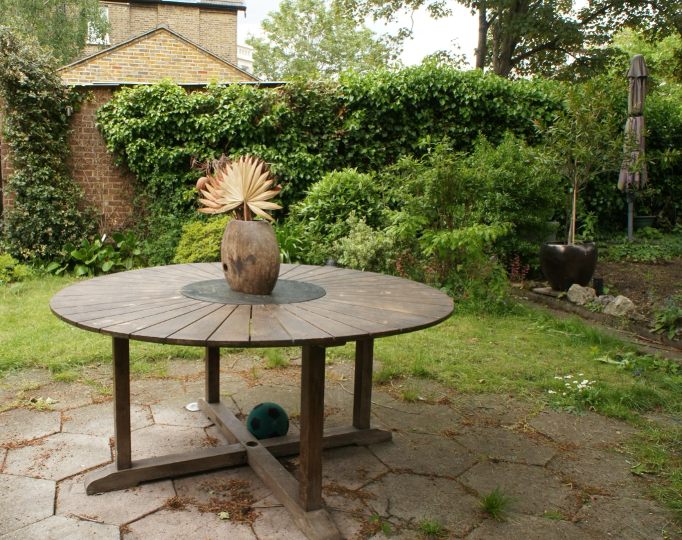}%
        \label{fig:thumb_garden}%
        }
    \caption{The seven scenes used in our evaluation: Playroom, Drjohnson, Train, Truck, Kitchen, Room, and Garden. These scenes encompass diverse man-made environments, ranging from interior spaces with furniture and architectural details to exterior objects.}
    \label{fig:thumbnails}
\end{figure*}

\textbf{Dataset and Metrics}.
We conducted a comprehensive evaluation of our method across seven representative scenes from three distinct datasets: \textit{Playroom} and \textit{Drjohnson} from the Deep Blending dataset~\cite{hedman2018deep}, \textit{Room}, \textit{Kitchen}, \textit{Garden}, and \textit{Train} from the Mip-NeRF360 dataset~\cite{barron2022mip}, and \textit{Truck} from the Tanks\&Temples dataset~\cite{knapitsch2017tanks}. 
As illustrated in \cref{fig:thumbnails}, these scenes were strategically selected to encompass a diverse spectrum of geometric complexities and structural characteristics. 
This comprehensive selection enables systematic assessment of our method's adaptability across varying scene morphologies, including bounded indoor environments with well-defined structural elements (Playroom, Drjohnson, Room, Kitchen), expansive unbounded outdoor scenes with irregular structures (Truck, Train, Garden).

We quantitatively assessed the visual fidelity of our 3D Gaussian representations using three complementary metrics: peak signal-to-noise ratio (PSNR), structural similarity index measure (SSIM)~\cite{wang2004image}, and learned perceptual image patch similarity (LPIPS)~\cite{zhang2018unreasonable}. We use the same train/test split settings from the original 3DGS work~\cite{kerbl20233D} for consistent and meaningful comparisons. The metrics are computed based on the testing set.

\textbf{Implementation}.
Our implementation is based on the official 3D Gaussian Splatting codebase~\cite{gaussiancode}. 
We initially trained 3DGS models for each scene using the default hyperparameters on an NVIDIA H100 GPU. 
Subsequently, we applied our proposed categorization method to distinguish between Sketch and Patch Gaussians based on their structural characteristics and spatial coherence properties, as detailed in \cref{sec:method}.
Following the categorization, polynomial regression (PR) is utilized to model the spatial distribution of Sketch Gaussians, and the trained PR models and SketchGS positions are further compressed with quantization and Google's Draco, respectively. Patch Gaussians undergo pruning, retraining, and quantization to achieve further compact representation. Finally, all the encoded files are packed into bitstream with gzip compression.
The PR model's degree was determined through a comprehensive grid search, ensuring optimal representation. 
We maintained consistent hyperparameters across all training stages.

The identification and modeling of SketchGS were implemented using Python's multiprocessing capabilities, running on dual AMD Epyc 9334 CPUs with 3.9 GHz and 32 cores per CPU. 
For scenes with millions of Gaussians, our experiments demonstrate that through K-D tree and parallel processing optimization of the clustering and refinement stages, we achieve at least two orders of magnitude faster speed, where the gain is mainly from clustering. For refining and polynomial modeling, processing \qtyrange[range-units=single]{100}{200} clusters per scene requires only \qtyrange[range-units=single]{1}{2}{\minute}.
The subsequent PatchGS retraining takes a fraction of the original 3DGS training time, typically requiring only \SI{70}{\percent} to \SI{80}{\percent} of the initial training time with the same iterations due to the reduced Gaussian population after pruning. 
Consequently, our method achieves comparable or even faster end-to-end processing times compared to standard 3DGS training while producing significantly more storage-efficient representations. In \cref{subsec:results}, we analyze the time efficiency of our algorithm.

\textbf{Comparison Methods}.
We implemented three comparative approaches to validate our method:
\begin{itemize}
    \item \textit{Sketch\&Patch (S\&P)}. This represents our previous work~\cite{shi2025sketch}, which employs Line3D++~\cite{hofer2017efficient} for extracting 3D line segments as a geometric prior to identify SketchGS. 
    While this line-based method demonstrated promising results for scenes with strong linear features characteristic of man-made environments, it fundamentally relies on explicit straight-line detection and cannot generalize well to scenes with curved boundaries, organic structures, or natural environments. 
    By comparing against this baseline, we demonstrate the improvements achieved through our new clustering-based categorization that automatically adapts to diverse geometric structures without imposing handcrafted geometric assumptions.
    \item \textit{Prune\&Retrain}. 
    This method applies a uniform pruning and retraining strategy across all Gaussians, without distinguishing between Sketch and Patch categories. 
    All Gaussians are treated homogeneously, which represents a naive compression baseline that does not leverage structural scene properties or geometric coherence patterns.
    This method serves as an ablation study to demonstrate the significance of recognizing and processing Gaussians differently based on their structural roles.
    \item \textit{Sketch}. The method focuses exclusively on the encoding of SketchGS, deliberately omitting the pruning and retraining of PatchGS. Note that the SketchGS and PatchGS are retrieved by our clustering and refinement method.
    To quantify the storage efficiency and visual quality trade-offs inherent in compact boundary-defining feature representation, we progressively reduce the number of SketchGS and measure the visual quality and model size. This method is the only one not to retrain the model.
\end{itemize}

\begin{figure*}[t]%
\centering
    \subfloat[][Mip-NeRF360.]{%
        \includegraphics[width=0.324\textwidth]{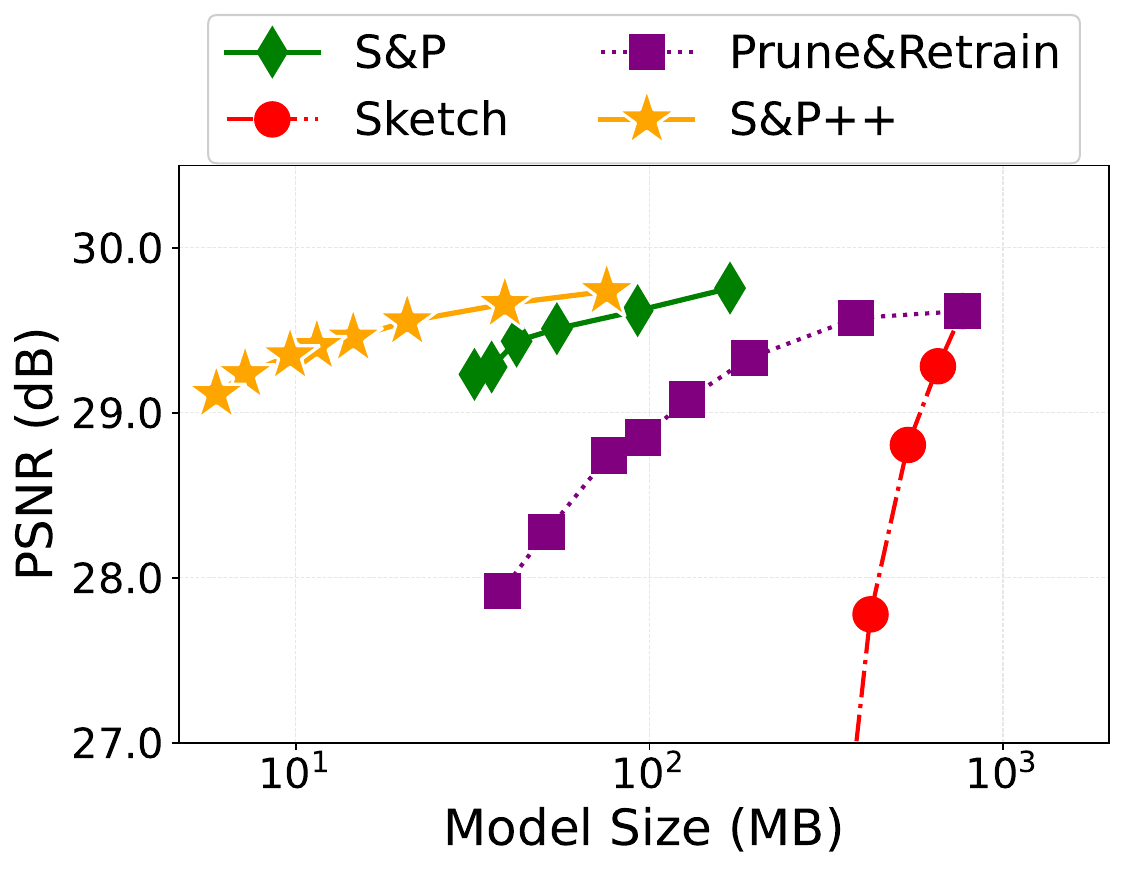}
        }
    \subfloat[][Tanks\&Temples.]{%
        \includegraphics[width=0.324\textwidth]{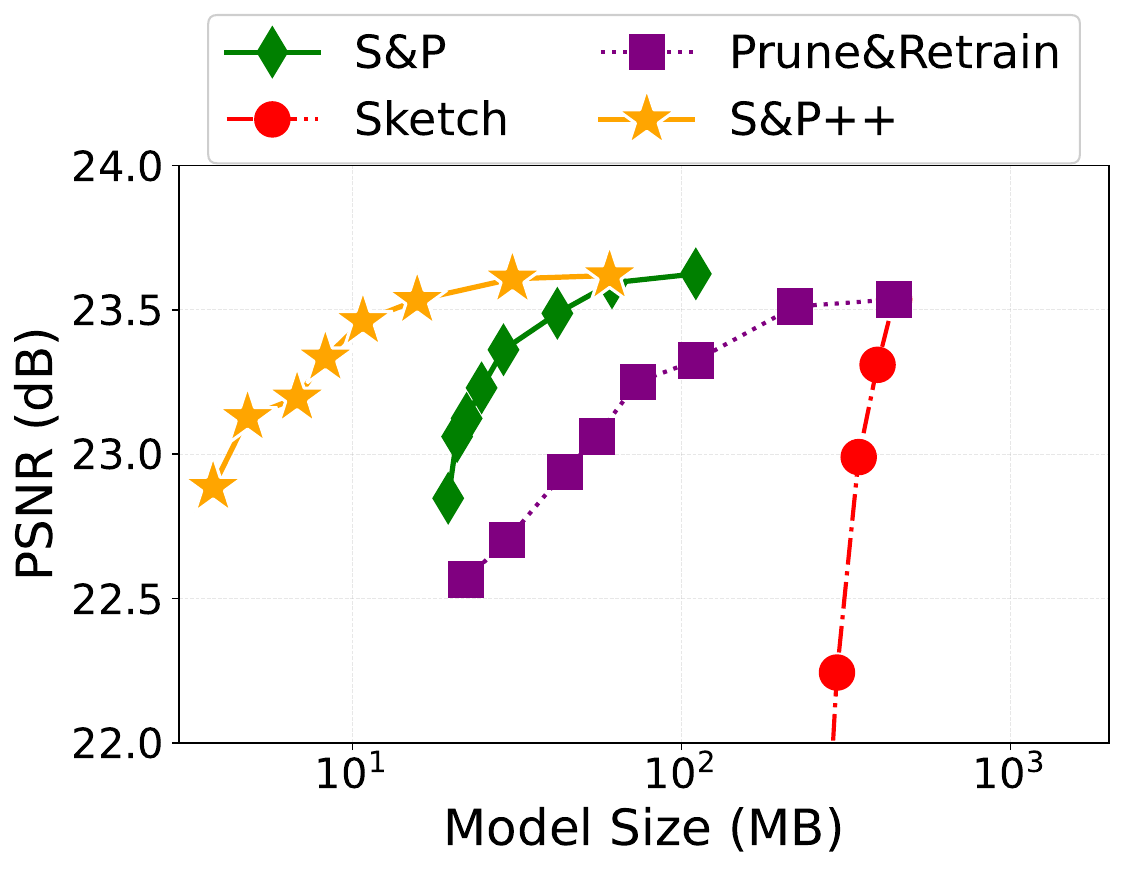}
        }
    \subfloat[][DeepBlending.]{%
        \includegraphics[width=0.324\textwidth]{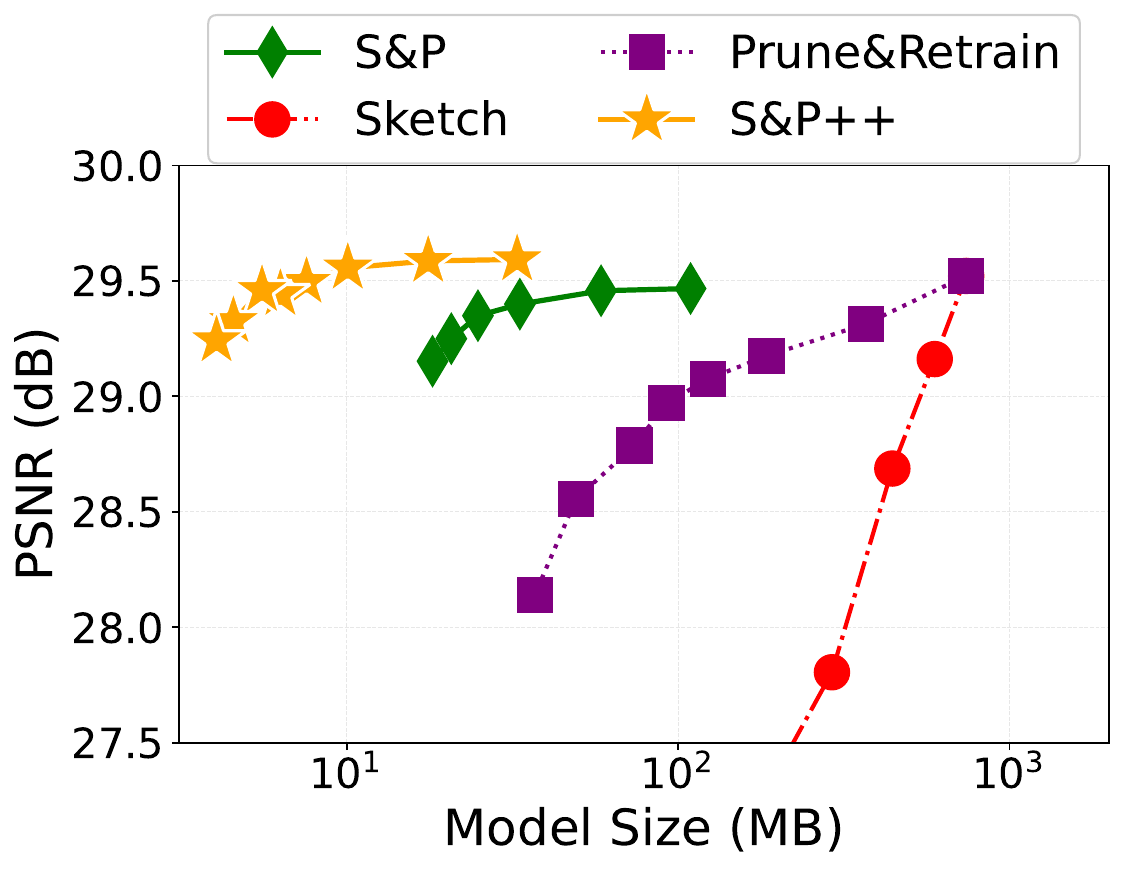}
        }
    \caption[]{R-D curves for PSNR: (a) Mip-NeRF360, (b) Tanks\&Temples, and (c) DeepBlending. Note the model size is log-scaled.}%
    \label{fig:psnr_rd_curve}
    \vspace{0.cm}
    \centering
    \subfloat[][Mip-NeRF360.]{%
        \includegraphics[width=0.324\textwidth]{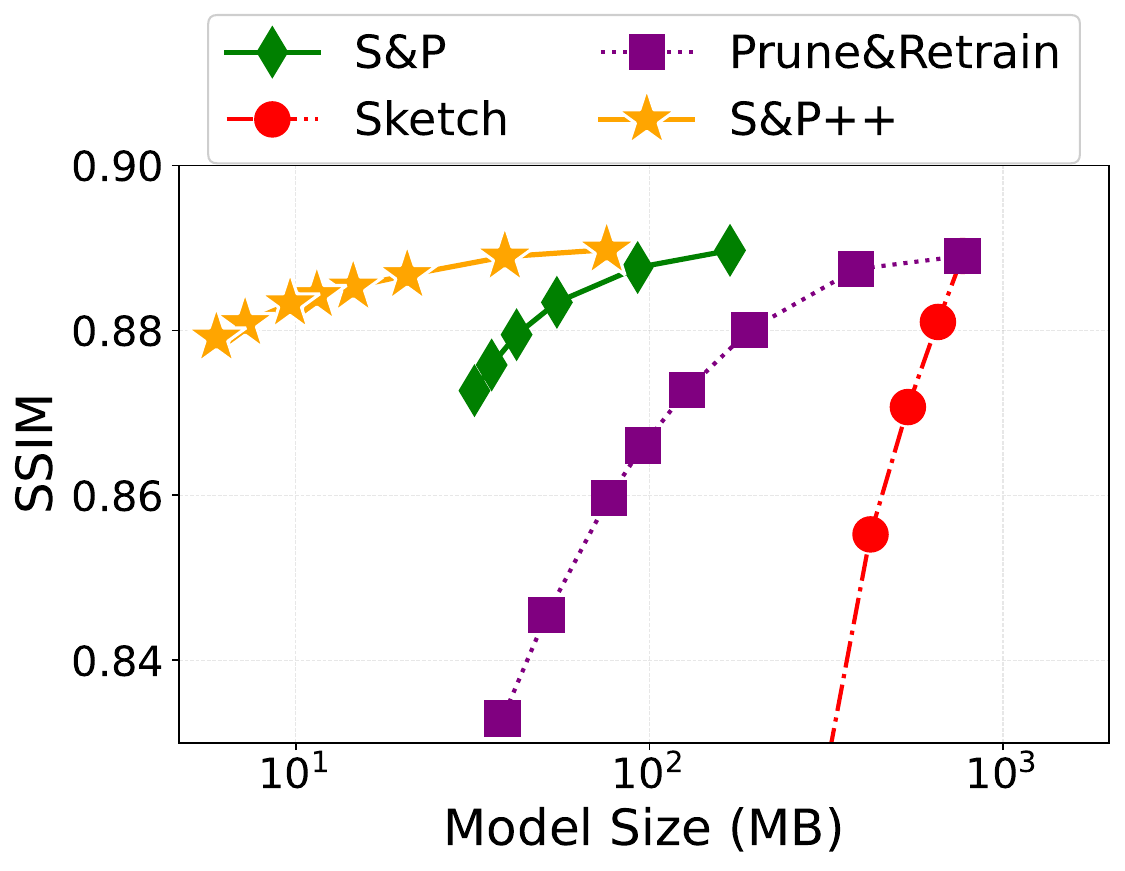}
        }
    \subfloat[][Tanks\&Temples.]{%
        \includegraphics[width=0.324\textwidth]{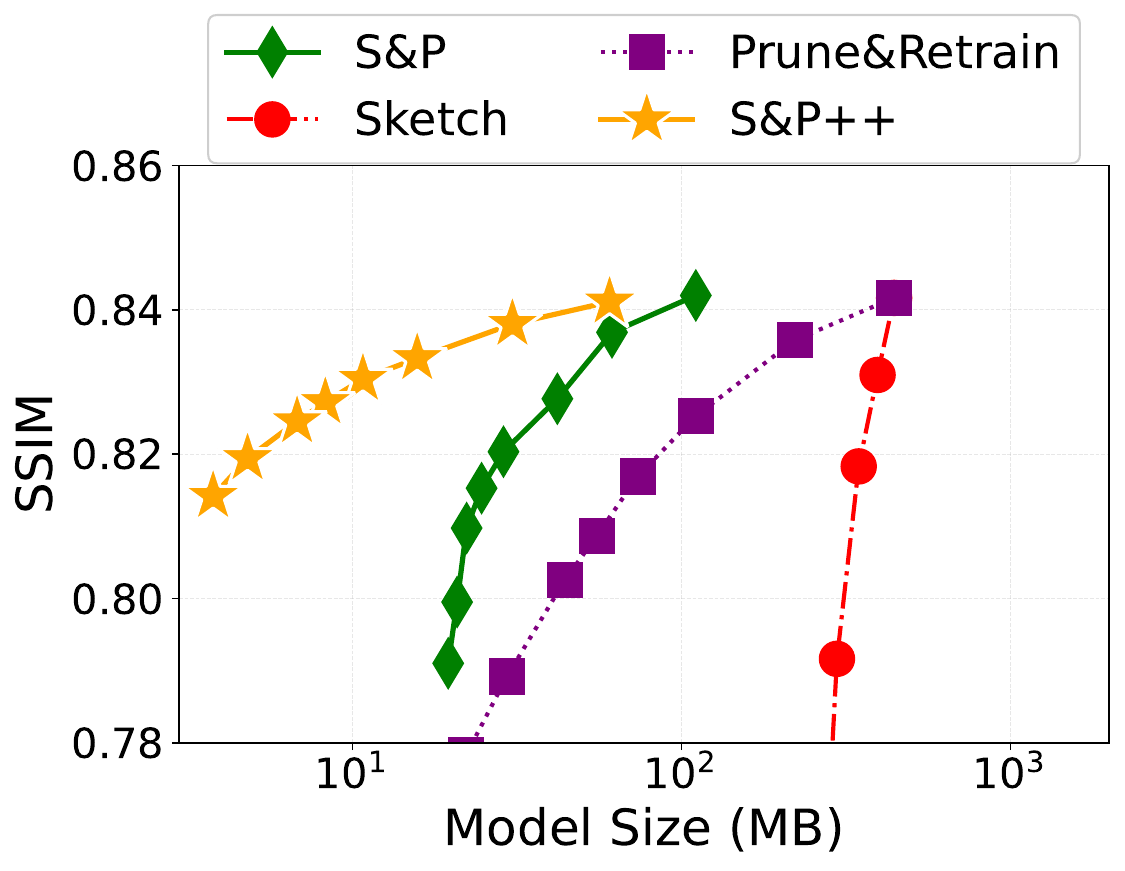}
        }
    \subfloat[][DeepBlending.]{%
        \includegraphics[width=0.324\textwidth]{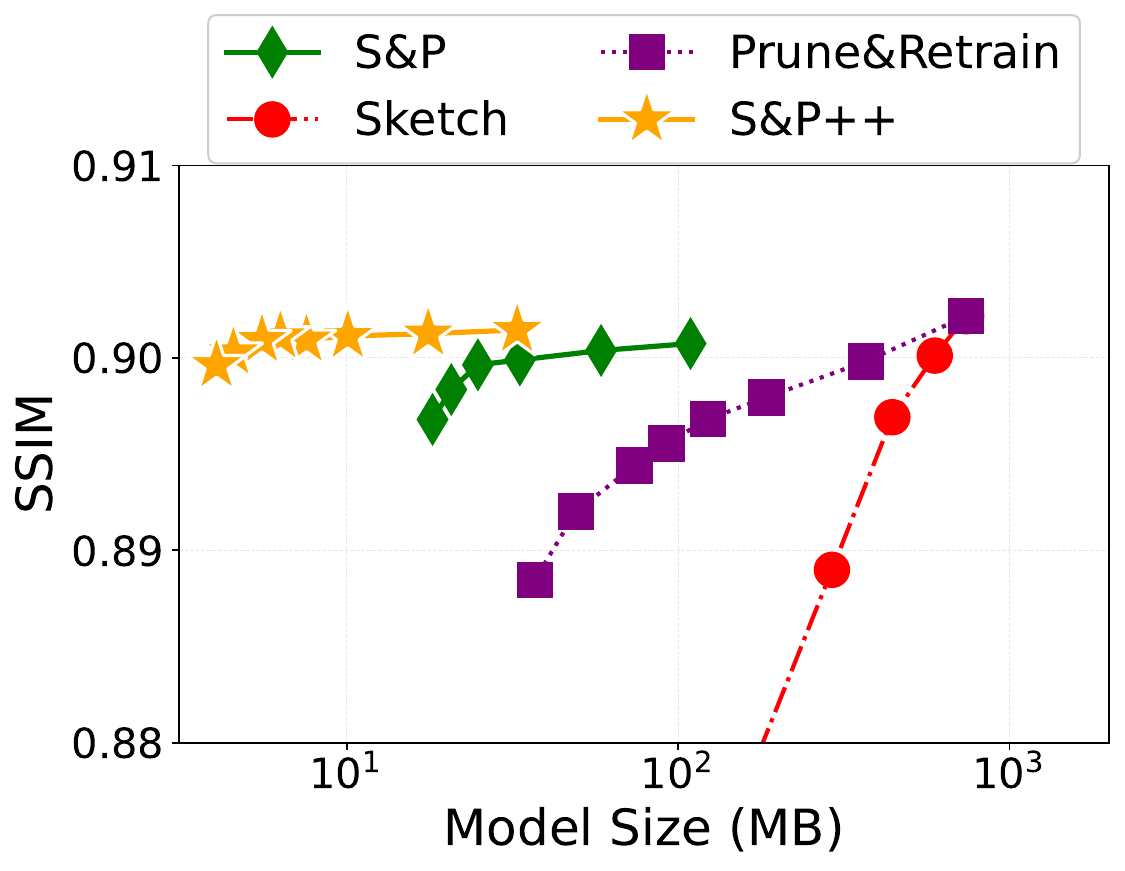}
        }
    \caption[]{R-D curves for SSIM: (a) Mip-NeRF360, (b) Tanks\&Temples, and (c) DeepBlending. Note the model size is log-scaled.}%
    \label{fig:ssim_rd_curve}
    \vspace{0.cm}
    \centering
    \subfloat[][Mip-NeRF360.]{%
        \includegraphics[width=0.324\textwidth]{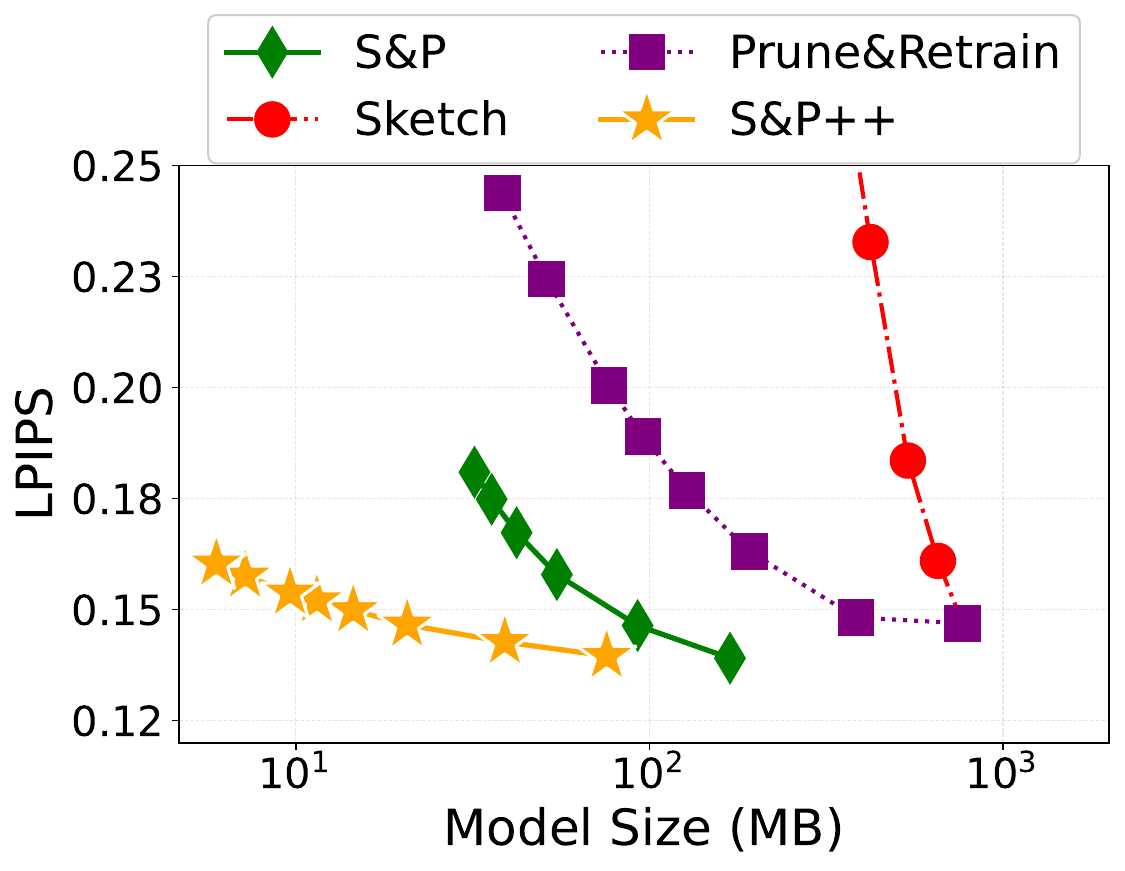}
        }
    \subfloat[][Tanks\&Temples.]{%
        \includegraphics[width=0.324\textwidth]{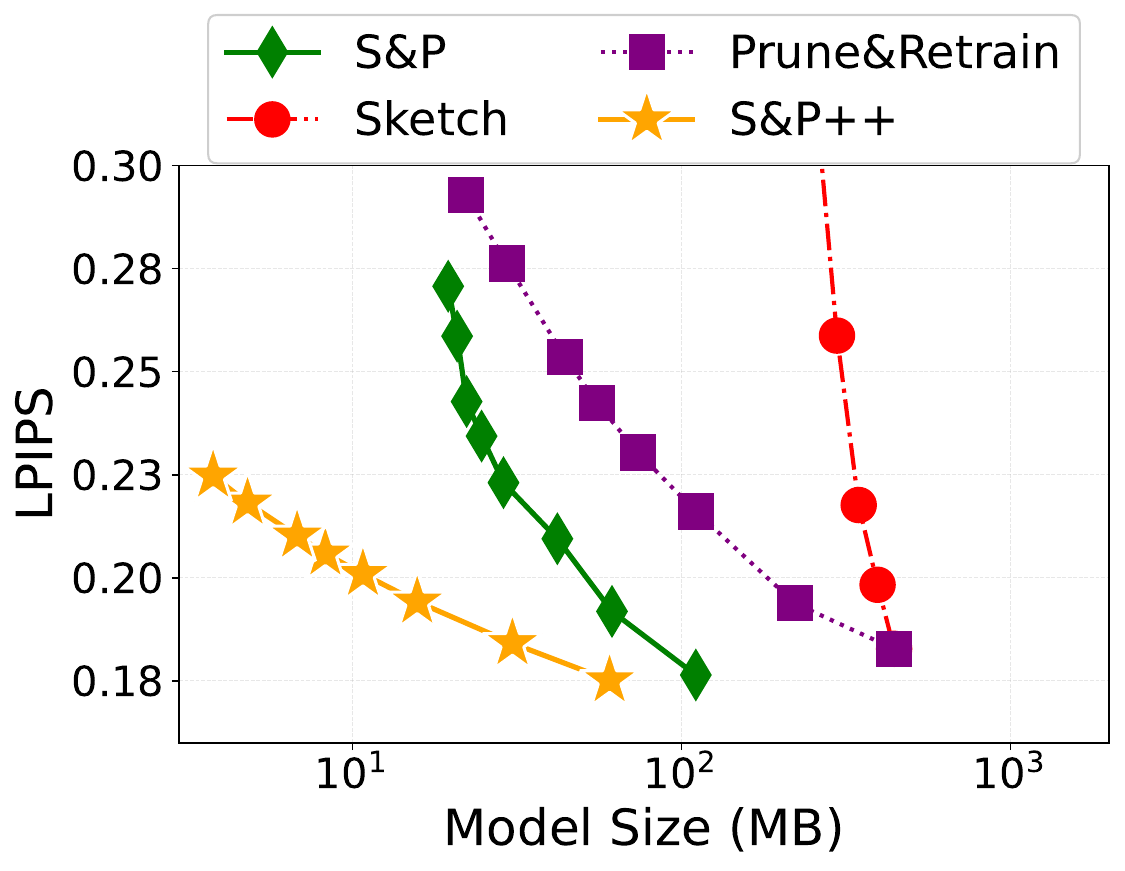}
        }
    \subfloat[][DeepBlending.]{%
        \includegraphics[width=0.324\textwidth]{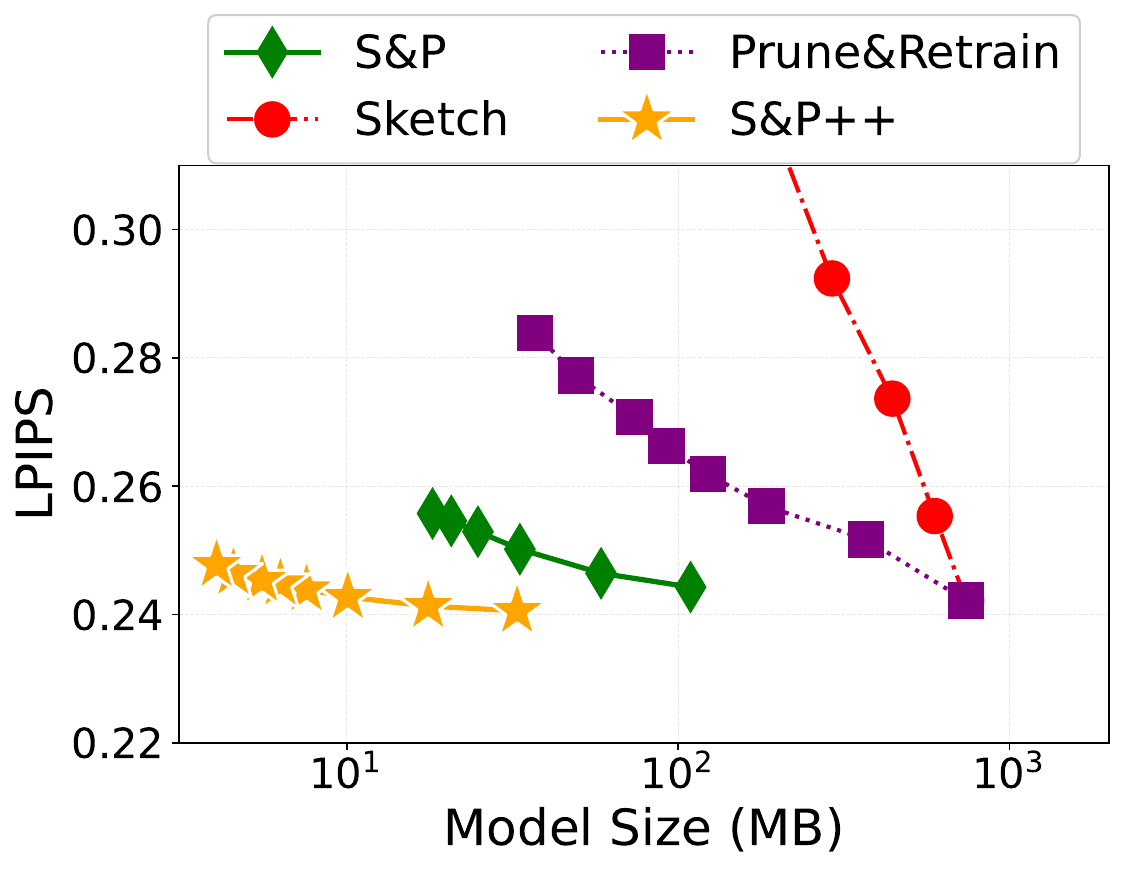}
        }
    \caption[]{R-D curves for LPIPS: (a) Mip-NeRF360, (b) Tanks\&Temples, and (c) DeepBlending. Note the model size is log-scaled.}%
    \label{fig:lpips_rd_curve}
\end{figure*}

\subsection{Results Analysis} \label{subsec:results}

\cref{fig:psnr_rd_curve,fig:ssim_rd_curve,fig:lpips_rd_curve} plot the visual quality versus the model size across the three benchmark datasets: Mip-NeRF360, Tanks\&Temples, and DeepBlending.
To generate rate-distortion (R-D) curves, we varied parameters for each method.
For the \textit{Prune\&Retrain} baseline, we created multiple models by reducing the number of Gaussians by factors of $\{1, 2, 4, 6, 8, 10, 15, 20\}$ through uniform pruning, followed by retraining.
For the \textit{Sketch} method, we reduce the number of SketchGS by randomly selecting $\{0\%, 25\%, 50\%, 75\%, 100\%\}$ of the Sketch Gaussians and adding these to the full set of Patch Gaussians.
For our previous method (denoted as \textit{S\&P}), we utilized all line segments from Line3D++~\cite{hofer2017efficient} for SketchGS identification and controlled the bitrate by reducing PatchGS with downsample factors of $\{1, 2, 4, 6, 8, 10, 15, 20\}$ through uniform pruning and retraining.
For our proposed method (denoted as \textit{S\&P++}), we control the bitrate in the same manner while employing the improved DBSCAN-based clustering for SketchGS identification and compression techniques.

We make the following observations from these results.

\textbf{Compression Efficiency.}
Compared to the original 3DGS models (the rightmost points), our \textit{S\&P++} method achieves substantial compression ratios while maintaining comparable or even superior visual quality.
For instance, on DeepBlending, \textit{S\&P++} compresses the model from \SI{703.77}{\mega\byte} to \SI{4.03}{\mega\byte} (a 175$\times$ compression ratio, \SI{0.5}{\percent} of the original size) with only a \SI{0.28}{\decibel} drop in PSNR (from 29.52 to 29.24) and negligible degradation in SSIM (from 0.902 to 0.898).
Remarkably, on Mip-NeRF360, \textit{S\&P++} at \SI{75.69}{\mega\byte} actually surpasses the original \SI{767}{\mega\byte} model, achieving \SI{29.74}{\decibel} PSNR compared to \SI{29.62}{\decibel}. This improvement can be explained by the fact that raw 3DGS models contain redundant Gaussians that may introduce rendering artifacts; better pruning and densification strategies can improve reconstruction quality at lower sizes~\cite{shi2025sketch,zhang2025gaussianspa,kim2024color,li2025geogaussian}.

\textbf{Improvement over S\&P.}
Our proposed \textit{S\&P++} method consistently achieves smaller model sizes at the same quality while maintaining better visual quality at the same model size, compared to our previous \textit{S\&P} method.
For example, at equivalent PSNR levels, \textit{S\&P++} requires only \SI{25}{\percent} to \SI{38}{\percent} of the storage needed by \textit{S\&P}. Specifically, compared to \textit{S\&P}, \textit{S\&P++} achieves {2.7$\times$} smaller on Mip-NeRF360 (\SI{20.6}{\mega\byte} vs.\ \SI{54.7}{\mega\byte} at \SI{29.5}{\decibel} PSNR), {3.3$\times$} smaller on DeepBlending (\SI{10.1}{\mega\byte} vs.\ \SI{33.3}{\mega\byte} at \SI{29.4}{\decibel} PSNR), and {3.9$\times$} smaller on Tanks\&Temples (\SI{15.7}{\mega\byte} vs.\ \SI{61.5}{\mega\byte} at \SI{23.5}{\decibel} PSNR).
This improvement stems from our DBSCAN-based clustering approach, which more accurately identifies and refines SketchGS.
Unlike \textit{S\&P}'s reliance on detected line segments with Line3D++~\cite{hofer2017efficient}, which may miss important boundary features or include biased detections, our multi-criteria clustering directly operates on Gaussian attributes to capture true high-frequency and boundary-defining features, resulting in more efficient SketchGS encoding and better PatchGS allocation.

\textbf{Robustness Compared to Prune\&Retrain.}
The \textit{Prune\&Retrain} baseline exhibits rapid quality degradation as the model size decreases, whereas \textit{S\&P++} maintains stable performance across a wide range of compression ratios.
On Mip-NeRF360, \textit{Prune\&Retrain} drops \SI{1.70}{\decibel} in PSNR when compressed from \SI{767}{\mega\byte} to \SI{38}{\mega\byte}, while \textit{S\&P++} drops only \SI{0.63}{\decibel} over a similar relative compression (from \SI{76}{\mega\byte} to \SI{6}{\mega\byte}).
At comparable model sizes around \SI{38}{\mega\byte}, \textit{S\&P++} outperforms \textit{Prune\&Retrain} by \textbf{\SI{1.74}{\decibel}} in PSNR, \textbf{\SI{6.7}{\percent}} in SSIM, and \textbf{\SI{41.4}{\percent}} in LPIPS.
This substantial performance gap arises because uniform global pruning indiscriminately removes Gaussians regardless of their importance, inevitably eliminating critical high-frequency features essential for sharp edges and fine details.
In contrast, our method's categorization preserves SketchGS (high-frequency boundary-defining features) intact while selectively pruning only from PatchGS (low-frequency volumetric regions), where redundancy is higher and removal has less perceptual impact.
We further analyze this effect in \cref{subsec:ablation_sketch_patch}.

\textbf{Necessity of PatchGS Retraining.}
The \textit{Sketch} method, which simply combines SketchGS with the original unpruned PatchGS without retraining, consistently yields the worst performance across all datasets and metrics.
This poor performance indicates that PatchGS retraining is essential. After SketchGS are identified and refined, the remaining PatchGS must be retrained to compensate for representation errors and adapt to the modified Gaussian distribution.
Without this retraining step, PatchGS cannot effectively fill the gaps left by SketchGS extraction, leading to visible artifacts and quality degradation. As a learning-based representation, retraining 3DGS is a common and widely used strategy to maintain the quality and size trade-off~\cite{3DGSzip2025,shi20253d}.

\begin{figure*}[t]%
    \centering
    \includegraphics[width=\textwidth]{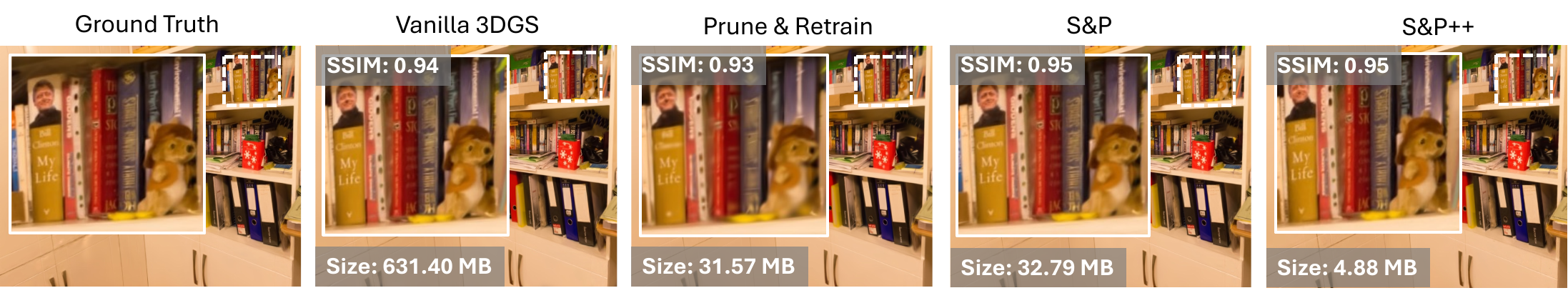}
    \caption[]{Rendered images of 3DGS on Playroom generated from four methods. }%
    \label{fig:vis_playroom}
    %
    %
    \vspace{0.3cm}
    \centering
    \includegraphics[width=\textwidth]{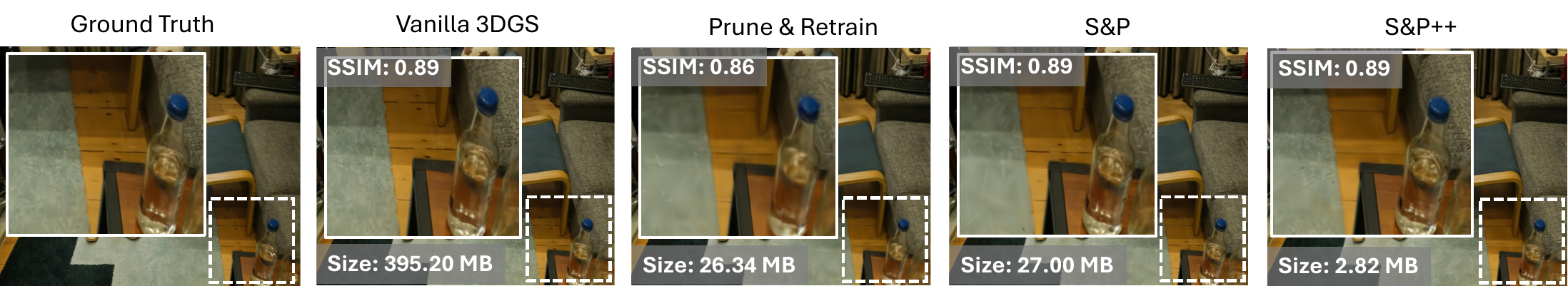}
    \caption[]{Rendered images of 3DGS on Room generated from four methods.}%
    \label{fig:vis_room}
    \vspace{0.3cm}
    \centering
    \includegraphics[width=\textwidth]{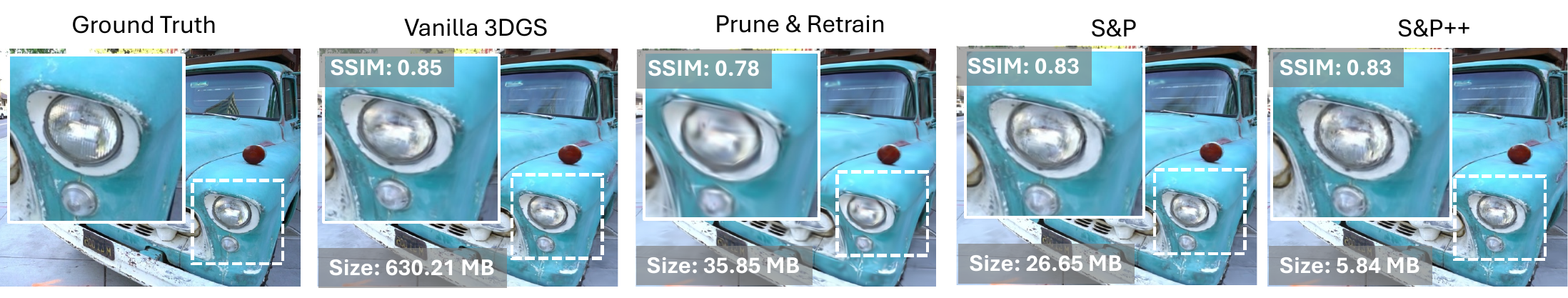}
    \caption[]{Rendered images of 3DGS on Truck generated from four methods.}%
    \label{fig:vis_truck}
\end{figure*}

\textbf{Qualitative Comparison}.
To illustrate the performance of the proposed method, we also present the visual comparison with the other methods in \cref{fig:vis_playroom,fig:vis_room,fig:vis_truck}, for qualitative analysis. 
For fair comparison, we selected models representing each method's maximum achievable compression while maintaining acceptable visual quality, corresponding to the leftmost points in the R-D curves. 

It can be observed that both \textit{S\&P} and \textit{S\&P++} excel in preserving geometric details and structural integrity, owing to their explicit categorization of Gaussians into SketchGS and PatchGS with separate compression strategies.
On Playroom, both methods achieve SSIM of 0.95, which surpass even the original Vanilla 3DGS (0.94), while \textit{Prune\&Retrain} drops to 0.93.
The zoomed regions reveal that our methods effectively preserve sharp edges and boundary features, such as the clearly readable book titles on the bookshelf and the fine fur texture of the stuffed animal.
On Truck, the performance gap becomes more pronounced. \textit{S\&P} and \textit{S\&P++} maintain SSIM of 0.83, whereas \textit{Prune\&Retrain} degrades to 0.78, exhibiting visible blurring around the headlight contours and loss of metallic surface details.
This advantage stems from the categorization-based approach: by identifying and preserving SketchGS (high-frequency boundary features) separately from PatchGS (low-frequency volumetric regions), our methods retain critical geometric structures that uniform global pruning inevitably destroys.

Notably, \textit{S\&P++} achieves visual quality comparable to \textit{S\&P} while requiring substantially less storage.
At equivalent SSIM levels, \textit{S\&P++} is {6.7$\times$} smaller on Playroom (\SI{4.88}{\mega\byte} vs.\ \SI{32.79}{\mega\byte}), {9.6$\times$} smaller on Room (\SI{2.82}{\mega\byte} vs.\ \SI{27.00}{\mega\byte}), and {4.6$\times$} smaller on Truck (\SI{5.84}{\mega\byte} vs.\ \SI{26.65}{\mega\byte}).
This significant storage reduction, without sacrificing perceptual quality, demonstrates the effectiveness of our improved clustering-based SketchGS identification, which more precisely captures boundary-defining features and enables more aggressive yet visually lossless compression of PatchGS.

\subsection{Ablation Effect of Sketch and Patch} \label{subsec:ablation_sketch_patch}

\begin{figure*}[t!]
    \centering
    \subfloat[SSIM and SketchGS ratio vs. model size.]{
        \begin{minipage}[b]{0.37\textwidth}
            \centering
            \includegraphics[width=\textwidth]{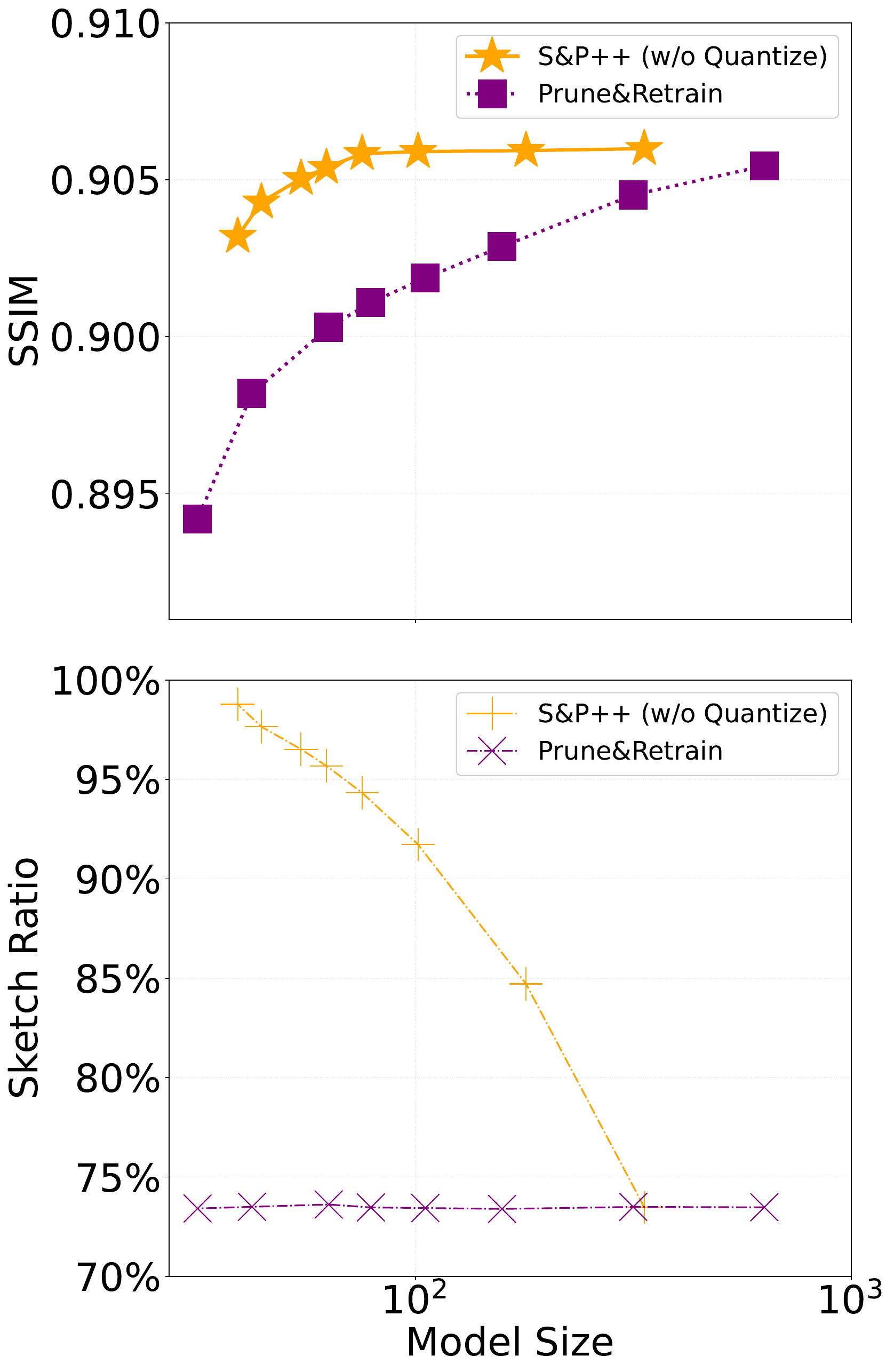}
        \end{minipage}
        \label{subfig:sketch_ratio_plot}
    }
    \subfloat[Prune\&Retrain, \SI{39}{\mega\byte}.]{%
        \begin{minipage}[b]{0.29\textwidth}
            \centering
            \includegraphics[width=\textwidth]{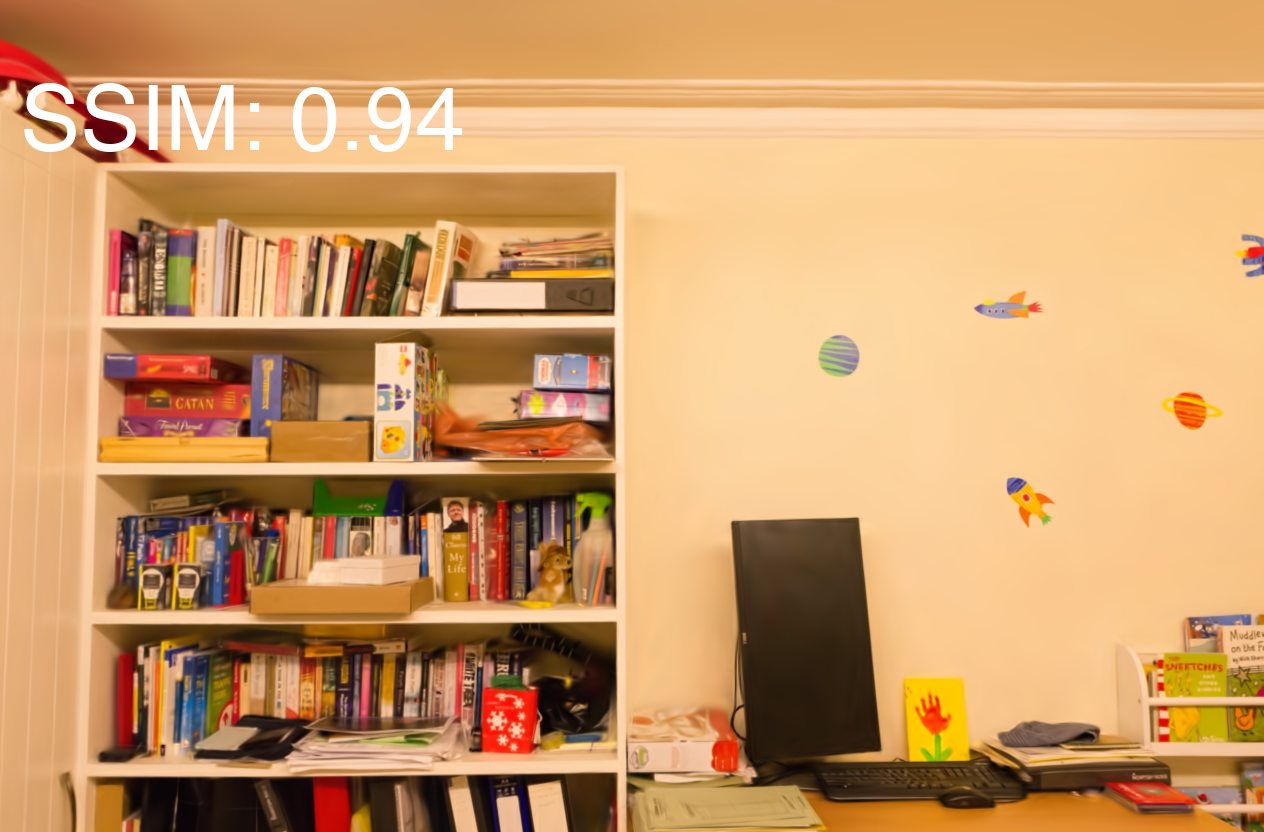}\\
            \includegraphics[width=\textwidth]{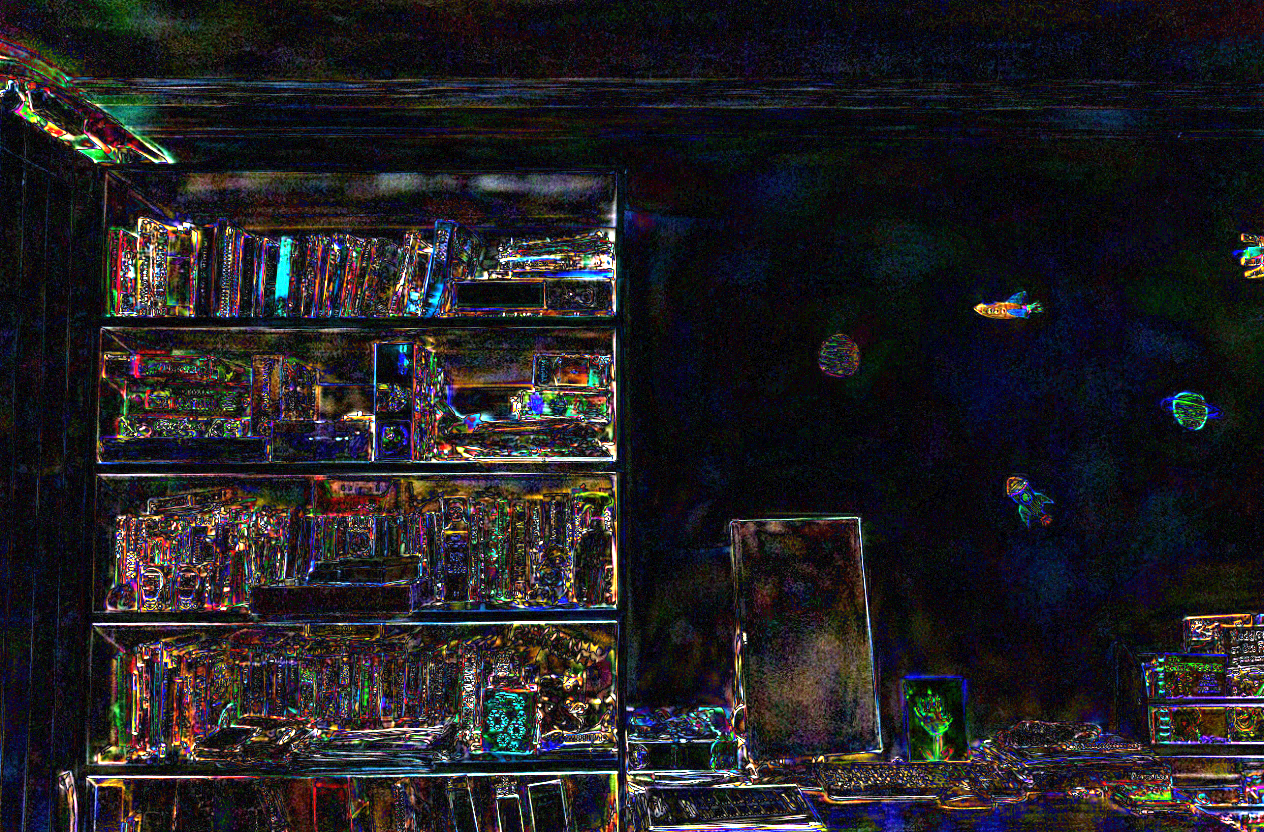}\\
            \includegraphics[width=\textwidth]{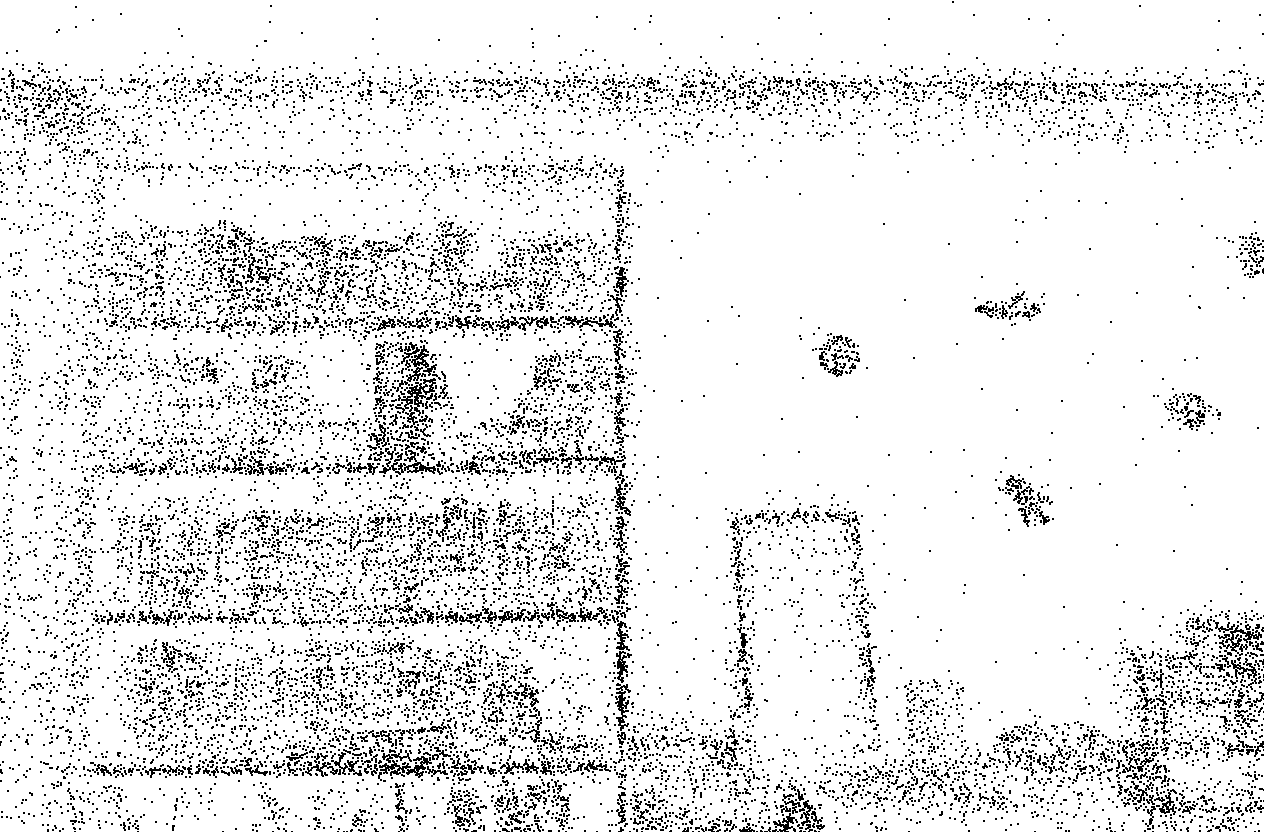}
        \end{minipage}
        \label{subfig:prune&retrain}
    }
    \hfill
    \subfloat[S\&P++ (w/o Quantize), \SI{31}{\mega\byte}.]{%
        \begin{minipage}[b]{0.29\textwidth}
            \centering
            \includegraphics[width=\textwidth]{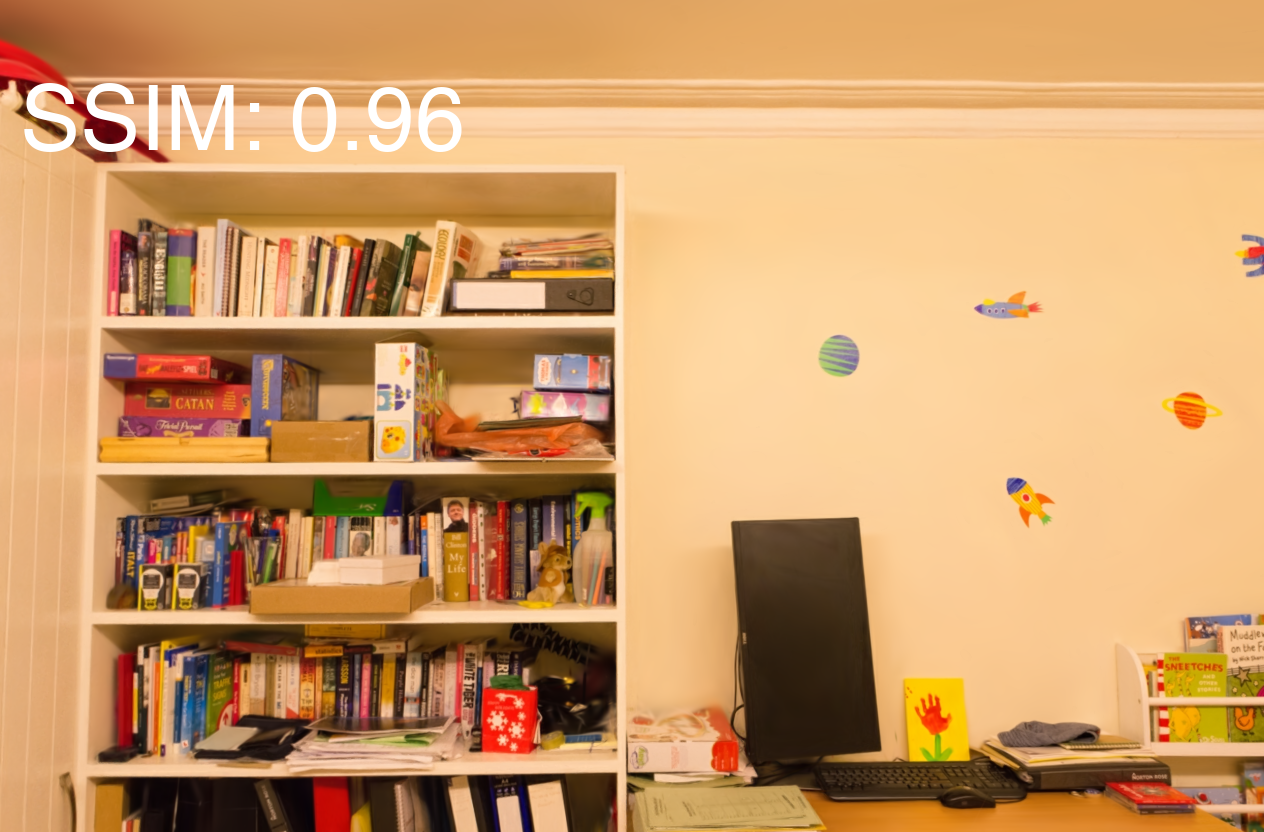}\\
            \includegraphics[width=\textwidth]{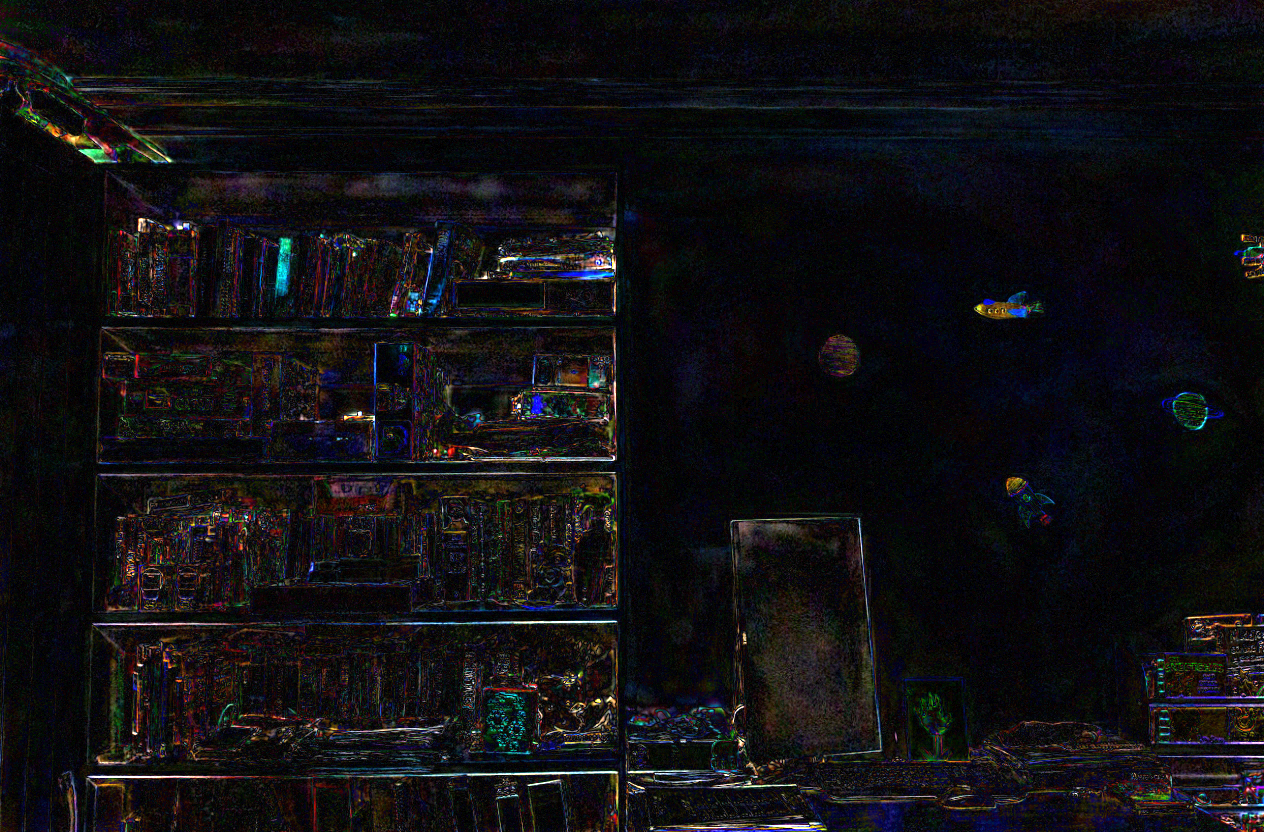}\\
            \includegraphics[width=\textwidth]{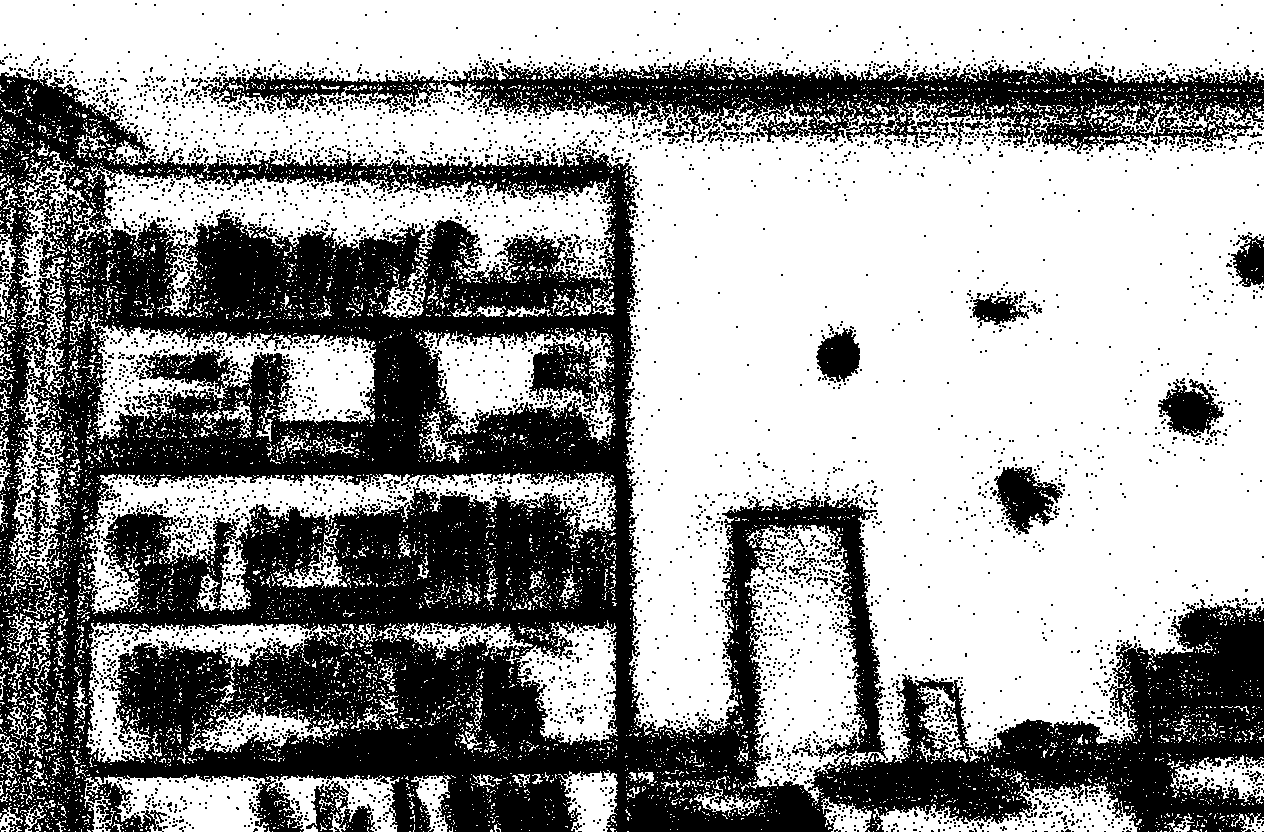}
        \end{minipage}
        \label{subfig:s&p++}
    }
    \hfill
    
    \caption{Ablation study of Sketch-Patch categorization on the \textit{Playroom} scene. 
    (a) SSIM and Sketch ratio versus model size.
    (b-c) Visual comparison showing rendered image (top), error map showing absolute RGB difference from ground truth (middle), and Gaussian centers (bottom). 
    Our method consistently outperforms uniform pruning across all compression levels. The quality advantage primarily stems from high-frequency boundary regions where our selective preservation of SketchGS maintains structural definition, while uniform pruning degrades these perceptually critical features.
    }
    \label{fig:ablation_sketchgs}
\end{figure*}

As shown in \cref{fig:psnr_rd_curve,fig:ssim_rd_curve,fig:lpips_rd_curve}, the superior performance of our method \textit{S\&P++} over the \textit{Prune\&Retrain} approach underscores the importance of differential treatment of Gaussian categories. 
With the same model size, our method consistently achieves better visual quality. 
This is because uniform pruning and retraining, while effective for general optimization, fail to capitalize on the distinct characteristics of boundary and non-boundary regions. 

To better illustrate this distinction, we show the visualization of the rendering performance versus the ratio of SketchGS for the scene \textit{Playroom} constructed by methods \textit{S\&P++ (w/o Quantize)} and \textit{Prune\&Retrain} in \cref{subfig:sketch_ratio_plot}. 
For fair comparison, we evaluate \textit{S\&P++} without vector quantization and bitstream compression to maintain comparable model sizes with \textit{Prune\&Retrain}.
Otherwise, with these additional compression techniques applied, the R-D curve of our complete \textit{S\&P++} method would shift significantly leftward, making direct visual comparison with \textit{Prune\&Retrain} difficult to interpret due to non-overlapping model size ranges.
This controlled comparison isolates the impact of our clustering-based categorization from the additional compression enhancements.

The Sketch ratio, defined as the number of Sketch Gaussians over the total number of Gaussians, provides valuable insight into the effectiveness of our method. 
As can be observed in \cref{subfig:sketch_ratio_plot} (bottom), in our \textit{S\&P++ (w/o Quantize)} method, the Sketch ratio increases as the model size decreases, ranging from approximately \SI{73}{\percent} at the largest model size to \SI{98}{\percent} at the most compressed state. 
This trend occurs because we fix the SketchGS while progressively pruning and optimizing the PatchGS, ensuring the preservation of critical boundary features even under extreme pruning.
In contrast, the \textit{Prune\&Retrain} method maintains a constant Sketch ratio of approximately \SI{73}{\percent} across all model sizes, as it uniformly prunes and optimizes all Gaussians without distinguishing their structural roles. 
This uniform treatment leads to suboptimal resource allocation, particularly evident in their SSIM values shown in \cref{subfig:sketch_ratio_plot} (top). 
Even when our method has the highest SketchGS ratio (\SI{98}{\percent}) at the smallest model size, it achieves better visual quality compared to \textit{Prune\&Retrain}. 
This superior performance demonstrates that preserving SketchGS while selectively optimizing PatchGS is more effective than uniform pruning, especially under stringent storage constraints.

We further visualize the qualitative differences between the 3DGS models with comparable size, which are en/decoded from two methods in \cref{subfig:prune&retrain} and \cref{subfig:s&p++}. 
The upper row shows the rendered results, where despite using \SI{31}{\mega\byte} storage, our method achieves noticeably higher quality with 0.96 SSIM, compared with \textit{Prune\&Retrain} at \SI{39}{\mega\byte} with 0.94 SSIM.
The middle row presents the error maps (absolute difference between rendered and ground truth RGB images).
Our method exhibits substantially darker error maps, particularly along object boundaries and structural edges, confirming superior reconstruction accuracy.
The bottom row visualizes the spatial distribution of Gaussian centers.
As can be observed, the quality difference primarily stems from the SketchGS regions, \ie, the high-frequency boundary features.
Our method's denser SketchGS coverage along structural boundaries corresponds to sharper, better-defined features in the rendering, while \textit{Prune\&Retrain}'s uniform pruning has removed critical boundary-defining Gaussians, leading to increased artifacts in these perceptually important regions.

In summary, the consistent superiority of our method validates the significance of distinguishing between SketchGS and PatchGS through clustering-based categorization, enabling targeted optimization strategies for each category.

\begin{table}[h]
        \caption{Impact of each component of the proposed method on model size at equivalent visual quality. We report storage requirements (in MB) across three datasets, and indicate relative compression ratios with respect to the baseline.} 
        \label{tab:gaussian_stat}
        \centering
        \vspace{0.5em}
            \begin{tabular}{l|c|c|c}
                \toprule
                \textbf{Model} & \textbf{Deep Blending} & \textbf{Tanks\&Temples} & \textbf{Mip-NeRF360} \\
                \midrule
                $\bullet$ Vanilla 3DGS  & $703.77$ ($1.0 \times$) & $436.50$ ($1.0 \times$) & $738.74$ ($1.0 \times$) \\
                $\bullet$ S\&P++ GS & $4.16$ ($169.2 \times$) & $3.85$ ($113.4 \times$) & $5.99$ ($123.3 \times$) \\
                    \hspace{2mm}+ w/ Bitstream Packing  & $4.03$ ($174.6 \times$) & $3.76$ ($116.1 \times$) & $5.90$ ($125.2 \times$) \\
                \midrule
                \hspace{0mm}$\circ$ Sketch Gaussians & $579.63$ ($1.0 \times$) & $191.43$ ($1.0 \times$) & $448.60$ ($1.0 \times$) \\
                    \hspace{2mm}+ w/ PR Encoding & $31.37$ ($18.5 \times$) & $10.42$ ($18.4 \times$) & $29.80$ ($15.1 \times$) \\
                    \hspace{2mm}+ w/ Draco Compress.  & $3.75$ ($154.6 \times$) & $1.31$ ($146.1 \times$) & $3.84$ ($116.8 \times$) \\
                    \hspace{2mm}+ w/ PR Quantization & $2.63$ ($220.4 \times$) & $0.79$ ($241.9 \times$) & $2.24$ ($200.3 \times$) \\
                \midrule
                \hspace{0mm}$\circ$ Patch Gaussians & $124.14$ ($1.0 \times$) & $245.07$ ($1.0 \times$) & $290.14$ ($1.0 \times$) \\
                    \hspace{2mm}+ w/ Prune \& Retrain & $7.10$ ($20.0 \times$) & $12.25$ ($20.0 \times$) & $14.51$ ($20.0 \times$) \\
                    \hspace{2mm}+ w/ Quantization & $1.53$ ($92.2 \times$) & $3.06$ ($80.1 \times$) & $3.75$ ($77.37 \times$) \\
                \bottomrule
            \end{tabular}
\end{table}

\subsection{Model Size Breakdown}

We present \cref{tab:gaussian_stat}, which provides a detailed breakdown of our hybrid representation's storage requirements, comparing SketchGS and PatchGS components with the baseline Vanilla 3DGS at equivalent visual quality levels in terms of LPIPS ($0.15$ on Mip-NeRF360, $0.23$ on Tanks\&Temples, and $0.24$ on DeepBlending).
Our \textit{S\&P++} achieves compression ratios of $174.6\times$, $116.1\times$, and $125.2\times$ for Deep Blending, Tanks\&Temples, and Mip-NeRF360 datasets respectively, reducing vanilla 3DGS models to just $4.03$ MB, $3.76$ MB, and $5.90$ MB while maintaining equivalent visual quality.

The results reveal that our parametric encoding of SketchGS achieves remarkable efficiency, requiring no more than \SI{2.63}{\mega\byte} of storage after full compression (PR encoding, Draco compression, and coefficient quantization) across all test scenes. 
The compact representation of SketchGS is particularly noteworthy, as these boundary-defining elements typically require dense sampling in traditional approaches to maintain edge and contour fidelity. 
Our method's ability to encode these features efficiently while preserving their visual importance validates the effectiveness of our PR-based approach combined with specialized geometry compression. 
Furthermore, this efficiency in SketchGS representation allows for a more optimal allocation of storage resources to PatchGS, contributing to the overall balance between storage efficiency and visual quality.

Moreover, we can observe that pruning and optimization, combined with vector quantization, contribute significantly to compressing the PatchGS, resulting in an average of $77\times$ to $92\times$ reduction in size compared to the original PatchGS representation.

In summary, this breakdown demonstrates that our method's superior compression performance stems from the synergy of specialized techniques for each Gaussian category.
SketchGS compression achieves $200\times$ to $242\times$ ratios through PR encoding, Draco compression, and quantization, while PatchGS compression achieves $77\times$ to $92\times$ ratios through pruning and vector quantization.
The asymmetric treatment enabled by our clustering-based categorization is essential for achieving the overall $116\times$ to $175\times$ compression ratios while maintaining visual quality.

\subsection{Time Efficiency}

\begin{figure}[t]
    \centering
    \begin{minipage}[c][8cm][c]{0.44\textwidth}
        \centering
        \includegraphics[width=\textwidth]{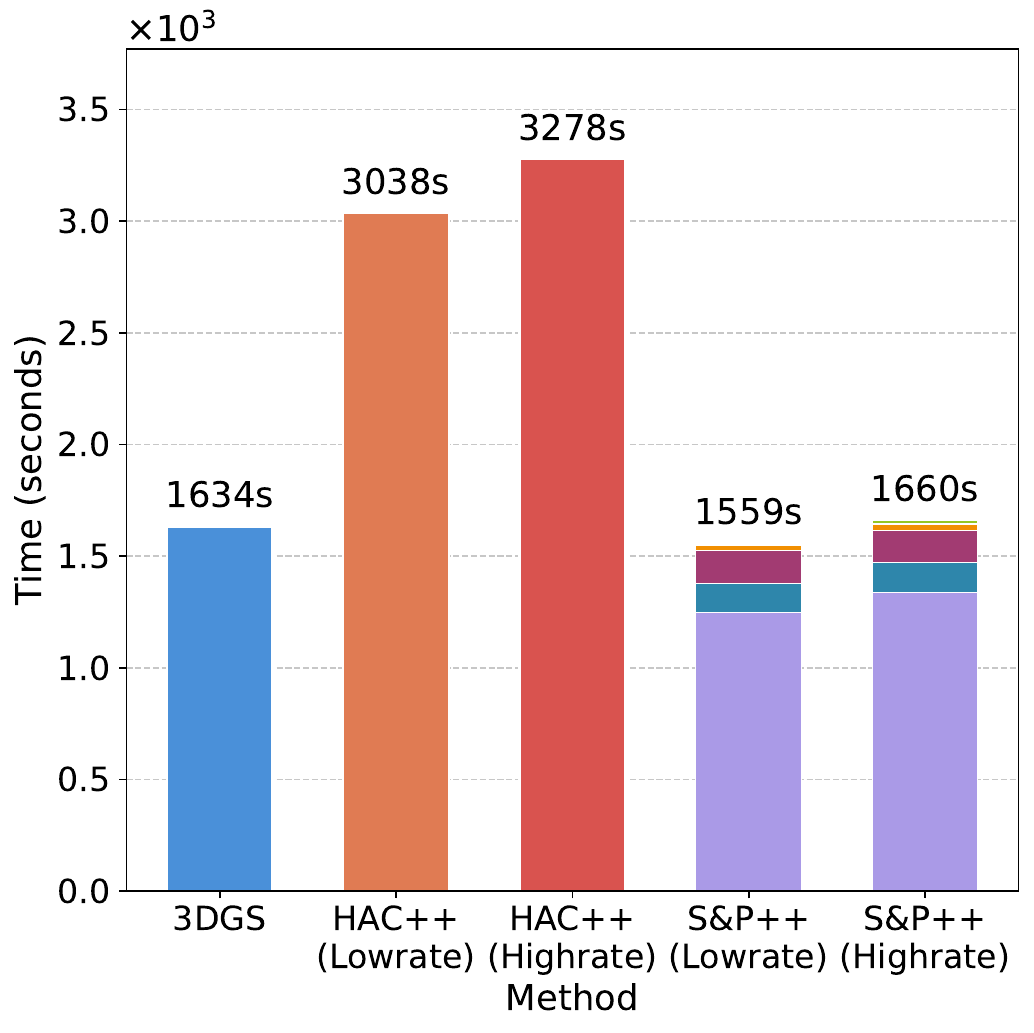}
        \caption{Training time across methods on Mip-NeRF360.  \textit{S\&P++} (stacked bars shows component breakdown) achieves comparable or faster training times than 3DGS while providing $116\times$-$175\times$ compression, and is $1.9\times$ faster than \textit{HAC++}~\cite{Chen2025}.}
        \label{fig:time_comparison}
    \end{minipage}%
    \hfill
    \begin{minipage}[c][8cm][c]{0.48\textwidth}
        \centering
        \captionof{table}{Training time breakdown (seconds) for \textit{S\&P++} on Mip-NeRF360. Lowrate configuration uses downsample factor $20$ for PatchGS pruning; highrate configuration skips PatchGS pruning. PatchGS retraining dominates computational cost (\SI{80}{\percent}), while categorization and encoding require \SI{20}{\percent} of time.}
        \small
        \begin{tabular}{cccc}
        \toprule
        Method & Component & Lowrate & Highrate \\
        \midrule
        3DGS & -- & \multicolumn{2}{c}{1633.66} \\
        HAC++ & -- & 3038.19 & 3278.08 \\
        \midrule
        \multirow{8}{*}{S\&P++} 
         & Clustering & \multicolumn{2}{c}{132.90} \\
         & Poly. Regression & \multicolumn{2}{c}{144.14} \\
         & Post-processing & \multicolumn{2}{c}{26.65} \\
         & Downsampling & 1.79 & 2.80 \\
         & Retraining & 1246.15 & 1338.12 \\
         & Quantization & 5.05 & 13.35 \\
         & Bitstream & 1.80 & 1.80 \\
        \cmidrule{2-4}
         & \textbf{Total} & \textbf{1558.48} & \textbf{1659.76} \\
        \bottomrule
        \end{tabular}
        \label{tab:time_breakdown}
    \end{minipage}
\end{figure}

Beyond compression efficiency and visual quality, computational time is also crucial for practical deployment. 
We compare the end-to-end compression time of our method against the training time of vanilla 3DGS, and compression time of \textit{HAC++}~\cite{Chen2025} which is reported as the state-of-the-art (SOTA) in 3DGS compression~\cite{3DGSzip2025}.
All methods are evaluated on the Mip-NeRF360 dataset, with each experiment repeated three times and averaged to ensure reliable measurements.
For \textit{HAC++}, we follow the original settings from the paper, which used two compression configurations to produce lowrate and highrate versions.
For our \textit{S\&P++} method, we also report results for two compression configurations: lowrate with aggressive PatchGS pruning (downsample factor $20$) and highrate without PatchGS pruning.
\cref{fig:time_comparison} presents the average training time comparison, while \cref{tab:time_breakdown} provides a detailed breakdown of the computational pipeline of our method.

As shown in \cref{fig:time_comparison}, vanilla 3DGS requires \SI{1634}{\second} for training on average, while \textit{HAC++} exhibits significantly higher computational costs of \SI{3038}{\second} (lowrate) and \SI{3278}{\second} (highrate), representing $1.9\times$ to $2.0\times$ overhead compared to vanilla 3DGS.
This substantial overhead stems from the complex hierarchical anchor-based decomposition and iterative refinement process.
In contrast, our \textit{S\&P++} method demonstrates superior computational efficiency, completing in \SI{1559}{\second} (lowrate) and \SI{1660}{\second} (highrate).
Remarkably, our lowrate configuration achieves \SI{4.6}{\percent} speedup over vanilla 3DGS while simultaneously providing $113\times$-$176\times$ compression, and our highrate configuration remains comparable to vanilla 3DGS with only \SI{4}{\second} overhead.
Compared to \textit{HAC++}, our method is approximately $1.9\times$ faster for both compression levels.

The computational efficiency of our method stems from its algorithmic design.
As shown in \cref{tab:time_breakdown}, the categorization and modeling stages require approximately \SI{304}{\second} total or \SI{19}{\percent} of training time. This includes clustering (\SI{133}{\second}), polynomial regression (\SI{144}{\second}), and post-processing (\SI{27}{\second}).
The downsampling and retraining stage dominates at \SI{80}{\percent} of total time, but operates on a dramatically reduced Gaussian population in the lowrate configuration after aggressive PatchGS pruning (downsample factor $20$, removing $\sim$ \SI{95}{\percent} of PatchGS), while the highrate configuration skips pruning entirely and only performs retraining on the original PatchGS population.
The reduced population in lowrate enables faster per-iteration optimization, allowing retraining to complete in \SI{1246}{\second} compared to \SI{1634}{\second} of vanilla 3DGS, while the highrate configuration requires \SI{1338}{\second} due to maintaining the full PatchGS set.
The remaining stages (quantization, and bitstream compression) impose minimal overhead of less than \SI{1}{\percent}.
Unlike the iterative hierarchical optimization of \textit{HAC++}, which requires multiple scene passes, our one-time clustering-based categorization followed by targeted optimization avoids iterative refinement overhead while achieving superior compression ratios.

In summary, our \textit{S\&P++} method achieves high compression ratios ($113\times$-$176\times$), superior visual quality, and training efficiency comparable to or better than vanilla 3DGS, while offering approximately $2\times$ speedup over \textit{HAC++}.
The lowrate configuration is ideal for storage-constrained scenarios requiring both training efficiency and compact representation, while the highrate configuration maintains training time parity with vanilla 3DGS while delivering substantial compression.
This combination makes our approach highly practical for real-world deployment where both storage and computational efficiency are critical.

\subsection{Comparison with State-of-the-Art Methods}

\begin{figure*}[t]%
    \centering
    \centering
    \subfloat[][Mip-NeRF360.]{%
        \includegraphics[width=0.324\textwidth]{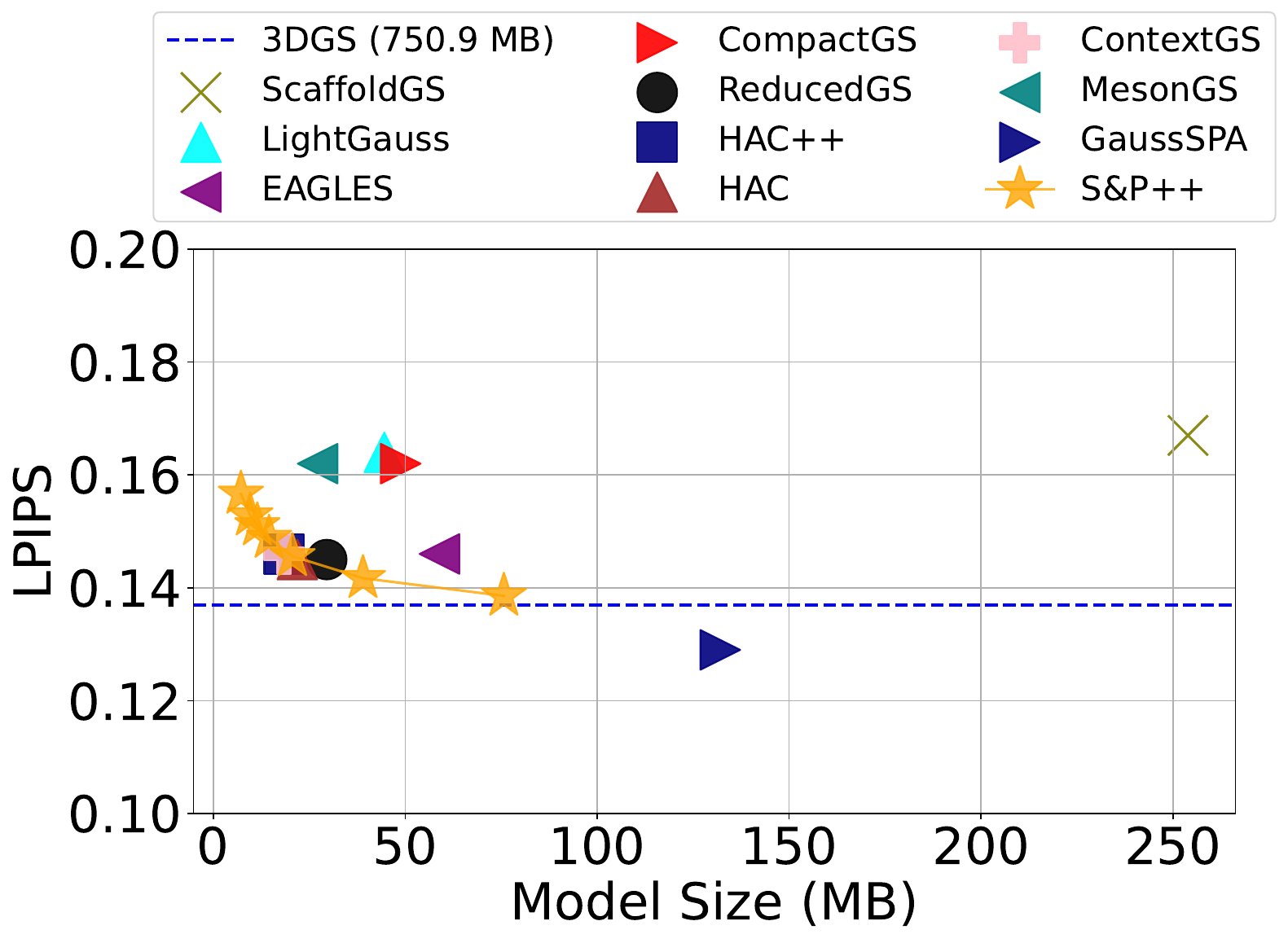}
        \label{subfig:sota_comp_mipnerf360}
        }
    \subfloat[][Tanks\&Temples.]{%
        \includegraphics[width=0.324\textwidth]{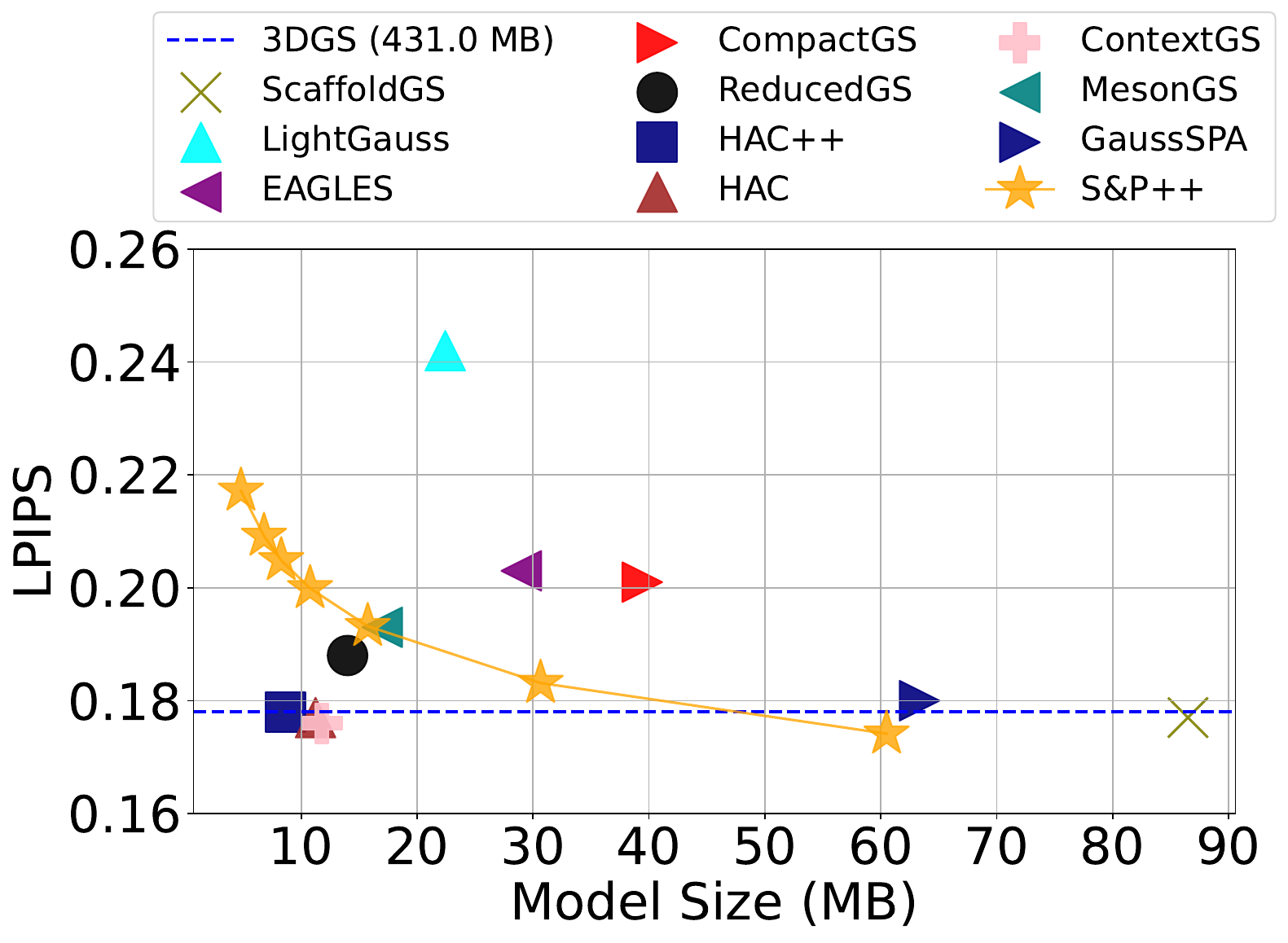}
        \label{subfig:sota_comp_tanks&temple}
        }
    \subfloat[][DeepBlending.]{%
        \includegraphics[width=0.324\textwidth]{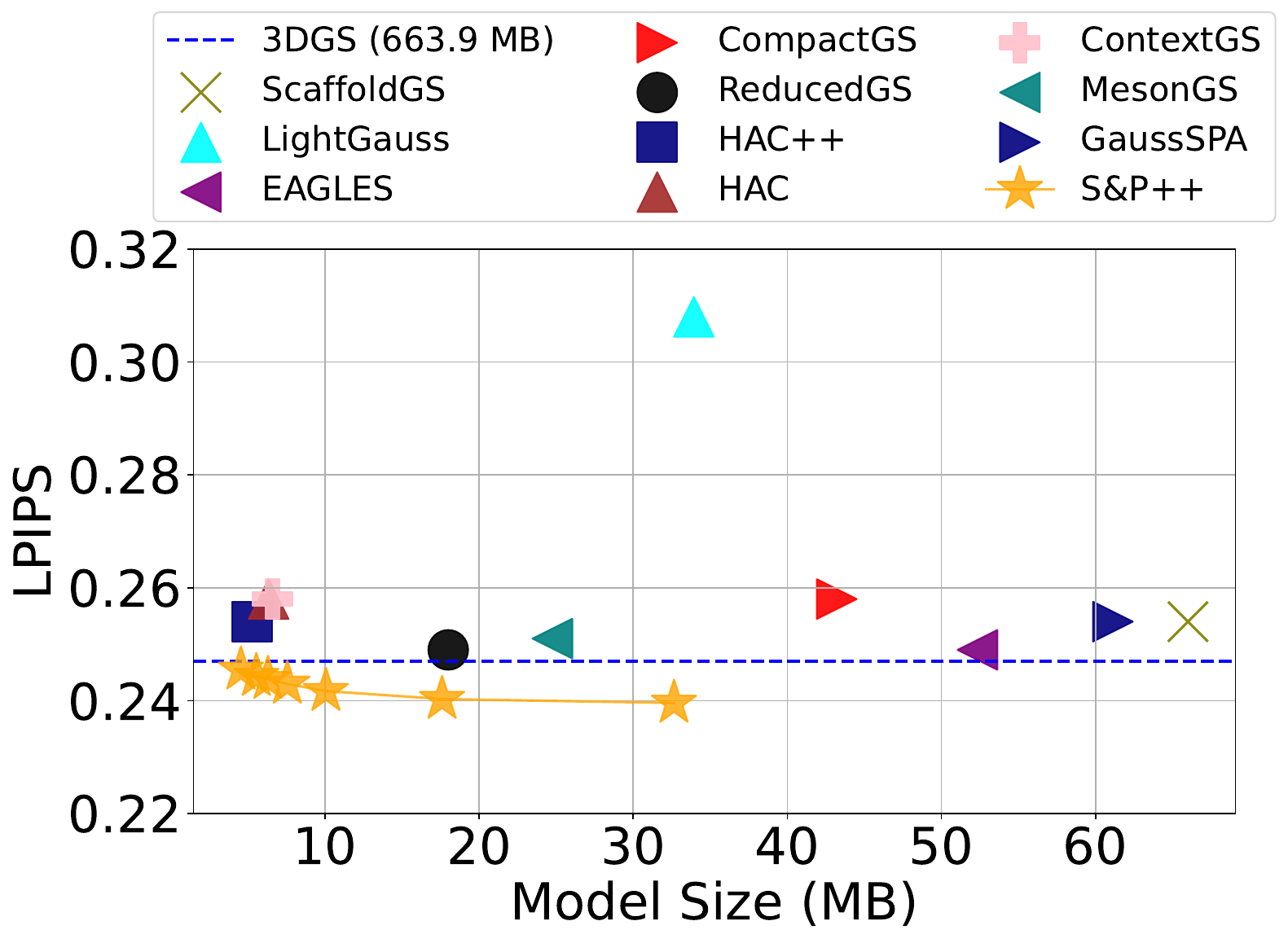}
        \label{subfig:sota_comp_db}
        }
    \caption[]{Comparison with the SOTAs on LPIPS: (a) Mip-NeRF360, (b) Tanks\&Temples, and (c) DeepBlending.}%
    \label{fig:sota_comp}
\end{figure*}

Our \textit{S\&P++} method is fundamentally designed to construct a compact representation with semantic structure—distinguishing features from volumetric content to enable applications such as adaptive streaming. Nevertheless, we evaluate its performance against SOTA 3DGS compression methods to demonstrate its competitiveness in rate-distortion trade-offs.
We compare against ten recent compression approaches: \textit{ScaffoldGS}~\cite{lu2024scaffold}, \textit{LightGaussian}~\cite{fan2023lightgaussian}, \textit{EAGLES}~\cite{girish2024eagles}, \textit{CompactGS}~\cite{lee2024compact}, \textit{ReducedGS}~\cite{papantonakis2024reducing}, \textit{ContextGS}~\cite{globersons2024contextgs}, \textit{MesonGS}~\cite{xie2024mesongs}, \textit{GaussSpa}~\cite{zhang2025gaussianspa}, \textit{HAC}~\cite{chen2024hac}, and \textit{HAC++}~\cite{Chen2025}.
\cref{fig:sota_comp} presents rate-distortion curves showing LPIPS versus model size across three datasets: Mip-NeRF360, Tanks\&Temples, and DeepBlending.

As shown, our method achieves highly competitive performance across all three datasets, consistently forming part of the rate-distortion frontier alongside the best-performing methods.
On the Mip-NeRF360 dataset (\cref{subfig:sota_comp_mipnerf360}) and Tanks\&Temples dataset (\cref{subfig:sota_comp_tanks&temple}), which includes both bounded indoor scenes and unbounded outdoor environments, our method achieves strong performance which is competitive to other SOTAs, achieving LPIPS values around $0.138$ to $0.217$ at the range from \SI{4}{\mega\byte} to \SI{60}{\mega\byte}.
Notably, our method demonstrates particularly strong performance on the DeepBlending dataset (\cref{subfig:sota_comp_db}), where it achieves the best quality-size trade-off among all compared methods. 
Specifically, at model sizes ranging from approximately \SI{4}{\mega\byte} to \SI{32}{\mega\byte}, our method achieves LPIPS values of $0.239$ to $0.245$, outperforming all competing approaches including the recent \textit{HAC++} which requires \SI{6}{\mega\byte} to reach $0.258$ LPIPS.
This superior performance on DeepBlending, which consists of bounded indoor scenes (\textit{Playroom} and \textit{Drjohnson}), validates our hypothesis that clustering-based categorization effectively identifies and compactly encodes high-frequency boundary features characteristic of man-made environments.

The exceptional performance on DeepBlending and competitive performance on Tanks\&Temples and Mip-NeRF360 correlate with scene structure characteristics.
DeepBlending primarily contains man-made environments with well-defined geometric structures, such as bookshelves, walls, building facades, where our clustering-based categorization excels at identifying coherent boundary features, yielding high SketchGS ratios (typically \SI{80}{\percent} as shown in \cref{tab:sketch_patch_stats}) that enable substantial compression through polynomial regression encoding.
In contrast, Mip-NeRF360 and Tanks\&Temples include unbounded outdoor scenes with organic structures and natural environments, where geometric coherence is less pronounced and volumetric content dominates, resulting in lower proportions of compressible structured features.
This scene-dependent behavior is expected and reflects a fundamental characteristic: methods specializing in structured feature compression excel on man-made environments, while more general pruning-based approaches may perform more uniformly across scene types at the cost of not exploiting structural coherence.

\subsection{Adaptive Streaming Capability} \label{subsec:streaming}

Beyond compression efficiency, the semantic separation of SketchGS and PatchGS provides a natural framework for adaptive streaming, which cannot be achieved by existing methods such as \textit{HAC++}~\cite{Chen2025}, which lack semantically meaningful decomposition. While a comprehensive evaluation of streaming performance is beyond the scope of this work, we present a preliminary exploration to illustrate the potential of our representation.

The asymmetry between the compact high-frequency boundary-defining representation (SketchGS) and the low-frequency volumetric content (PatchGS) motivates a layered streaming approach.
Conceptually, such a protocol could work as follows. The client first receives layer $L_0$ containing the compact SketchGS representation, establishing the geometric skeleton of the scene with minimal data transfer.
Subsequently, PatchGS could be transmitted in incremental layers ($L_1$ and so on), with Gaussians potentially ordered by opacity and volumetric coverage to prioritize visually significant content~\cite{shi2025lapisgs,fan2023lightgaussian,papantonakis2024reducing}. 
At any point during transmission, the renderer would combine all received layers to produce the current frame.

\begin{figure*}[t!]%
\centering
    \subfloat[][$L_0$ (1.48 MB).]{%
        \includegraphics[width=0.2\textwidth]{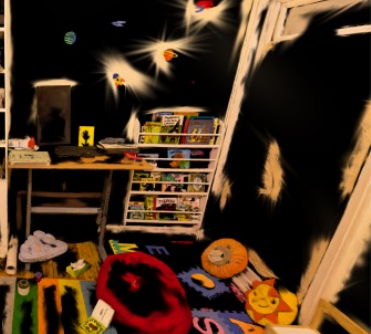}
        }
    \subfloat[][$L_1$ (5.47 MB).]{%
        \includegraphics[width=0.2\textwidth]{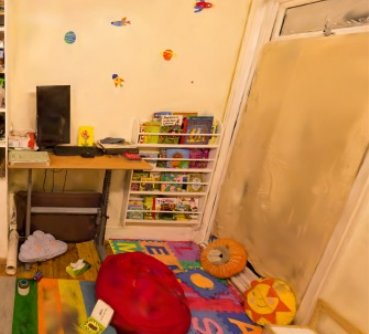}
        }
    \subfloat[][$L_2$ (9.46 MB).]{%
        \includegraphics[width=0.2\textwidth]{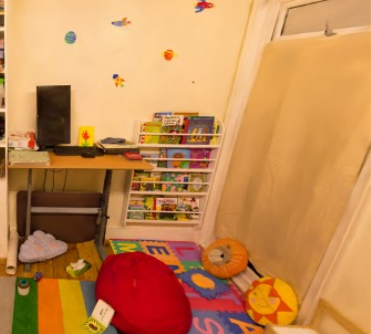}
        }
    \subfloat[][$L_3$ (13.46 MB).]{%
        \includegraphics[width=0.2\textwidth]{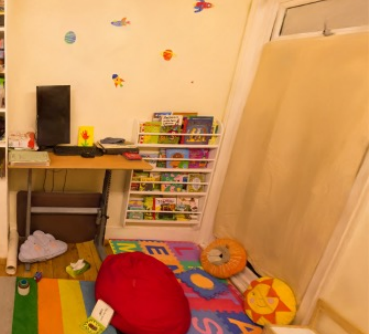}
        }
    \subfloat[][$L_4$ (17.45 MB).]{%
        \includegraphics[width=0.2\textwidth]{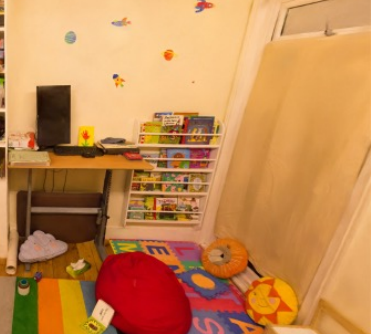}
        }
    \caption{Illustration of progressive rendering during adaptive streaming on the Playroom scene. From $L_0$ to $L_4$: SketchGS only (High-frequency boundary regions), then progressive addition of \SI{25}{\percent}, \SI{50}{\percent}, \SI{75}{\percent}, and \SI{100}{\percent} PatchGS.}
    \label{fig:streaming}
\end{figure*}

\cref{fig:streaming} illustrates the progressive rendering at each layer. We apply our method to the Playroom scene and produce a compressed model achieving comparable quality to vanilla 3DGS (PSNR: \SI{29.99}{\decibel}, SSIM: 0.906, LPIPS: 0.242). The resulting model is \SI{17.45}{\mega\byte}, with SketchGS requiring only \SI{1.48}{\mega\byte} (\SI{8.5}{\percent}) while PatchGS accounts for \SI{15.97}{\mega\byte} (\SI{91.5}{\percent}).
We construct four layers for our Sketch-Patch GS model, where $L_0$ is the SketchGS, $L_1$ includes the top \SI{25}{\percent} of PatchGS, $L_2$ the top \SI{50}{\percent}, $L_3$ the top \SI{75}{\percent}, and $L_4$ the complete set.
As shown, with $L_0$ alone, the renderer produces a structural skeleton capturing object boundaries and edges, which is not yet a complete rendering but providing a geometric foundation.
Upon receiving $L_1$ (\SI{5.47}{\mega\byte}), the scene becomes viewable as PatchGS begins filling volumetric regions.
Quality improves through subsequent layers $L_2$ (\SI{9.46}{\mega\byte}) and $L_3$ (\SI{13.46}{\mega\byte}), reaching full quality with $L_4$.

This decomposition is similar to the base-layer and enhancement-layer paradigm adopted in scalable video coding standards. Such a capability could benefit mobile and web-based 3DGS viewers under varying network conditions, VR/AR applications requiring responsive initial loading, and bandwidth-constrained streaming scenarios.
We leave a rigorous evaluation of streaming performance, including comparisons with alternative streaming strategies, to future work.

\section{Discussion and Conclusion}
\label{sec:conclude}

Previously, we proposed \textit{Sketch\&Patch}~\cite{shi2025sketch}, which introduced a novel perspective on hybrid 3D Gaussian representation by recognizing and leveraging the distinct roles of different Gaussian types in man-made environments. In this work, we propose a generalized framework \textit{Sketch\&Patch++} that successfully operates on arbitrary scenes with significant compression performance, overcoming the limitation of our earlier line-based approach to man-made environments.
By employing multi-criteria density-based clustering directly on Gaussian attributes, \textit{Sketch\&Patch++} identifies SketchGS and PatchGS based on geometry and appearance coherence, eliminating dependency on external geometric primitives such as Line3D++. Furthermore, Advanced compression and optimization techniques are tailored to SketchGS and PatchGS based on their unique roles in 3DGS representation.
Experimental results across diverse datasets, including both man-made and natural scenes, demonstrate that \textit{Sketch\&Patch++} achieves up to 175$\times$ (\SI{0.5}{\percent} of the original model size) compression while maintaining comparable visual quality, reaching comparable or superior performance to state-of-the-art compression methods.
Compared to \textit{Sketch\&Patch}, the generalized method achieves 2.7$\times$ to 3.9$\times$ smaller model sizes at equivalent quality levels.
Moreover, the semantic separation of SketchGS and PatchGS provides a unique capability not offered by existing methods: the potential for adaptive streaming, where the compact SketchGS layer can serve as an initial preview followed by progressive PatchGS refinement.

While our generalized method addresses the scene-type limitation and achieves superior compression performance, several challenges remain for future investigation.
First, our current implementation focuses on static scenes, and extending this approach to dynamic scenarios presents interesting challenges. Dynamic scenes would require temporal coherence in both Sketch and Patch Gaussian representations, as well as efficient updating mechanisms to handle moving objects and changing geometries. Developing incremental clustering strategies that can adapt to scene changes without full recomputation would be particularly valuable for real-time applications.
Second, while we demonstrated the potential for adaptive streaming through our preliminary exploration in \cref{subsec:streaming}, a comprehensive evaluation remains as future work.
Rigorous evaluation of this capability, including comparisons with alternative streaming strategies, latency measurements, and user studies under realistic network conditions, would be valuable for validating the practical benefits of our representation for bandwidth-adaptive applications.
Third, incorporating semantic understanding into the categorization process presents a promising direction.
Currently, our method relies on geometric and appearance features for clustering, but integrating semantic information could enable more intelligent Gaussian allocation based on object importance and viewer attention patterns.
For instance, Gaussians representing salient objects or regions of interest could be prioritized differently than background elements, potentially improving perceptual quality at aggressive compression ratios.

\bibliographystyle{plain}
\bibliography{manuscript}

\end{document}